\documentclass[twoside,11pt]{article}
\usepackage{jair, theapa, rawfonts}

\usepackage{amsmath, amssymb, amsfonts, mathrsfs}
\usepackage{mathtools}
\usepackage{times, helvet, courier}
\usepackage{graphicx, epsfig, epstopdf}
\usepackage{wrapfig, caption, multirow}
\usepackage[ruled,linesnumbered,noresetcount,vlined]{algorithm2e}
\usepackage{geometry}
\usepackage{rotating}
\usepackage{url}
\usepackage{color,soul} 
\usepackage{stmaryrd}
\usepackage{xargs}                      
\usepackage[pdftex,dvipsnames]{xcolor}  
\usepackage{picinpar}
\usepackage{floatrow}
\usepackage{tikz}
\def\checkmark{\tikz\fill[scale=0.4](0,.35) -- (.25,0) -- (1,.7) -- (.25,.15) -- cycle;} 
\newfloatcommand{capbtabbox}{table}[][\FBwidth]

\def\includeMemo{false}
\newcommand{\memo}[1]{
  \ifthenelse {\boolean{\includeMemo}}{\medskip\noindent\fbox{\begin{minipage}[b]{\dimexpr\linewidth-1em}#1\end{minipage}}\medskip\newline}
}

\let\chapter\undefined

\newboolean{comments}
\definecolor{darkgreen}{rgb}{0,0.6,0}
\newcommand{\kibitz}[2]{
\ifthenelse{\boolean{comments}}{{\color{#1}#2}}{}}

\setboolean{comments}{true}

\newtheorem{definition}{\bf Definition}

\newcommand{\benumerate}{\begin{list}{$\bullet$}{\topsep=0pt \parsep=0pt \itemsep=1pt \labelwidth=1.5em \labelsep=0.5em \leftmargin=20pt}}
\newcommand{\eenumerate}{\end{list}}
\newcommand{\bitemize}{\begin{list}{$\bullet$}{\topsep=0pt \parsep=0pt \itemsep=1pt \leftmargin=10pt}}
\newcommand{\eitemize}{\end{list}}

\newcommand{\iOne}{\textit{(i)}\xspace}
\newcommand{\iTwo}{\textit{(ii)}\xspace}
\newcommand{\iThree}{\textit{(iii)}\xspace}

\def\emptycell{\multicolumn{1}{c}{}}
\newcommand{\tuple}[1]{\langle #1 \rangle}

\newcommand{\size}[1]{|#1|}

\def\is{\!=\!}
\def\st{\: | \:}

\newcommand{\argmax}{\operatornamewithlimits{argmax}}
\newcommand{\argmin}{\operatornamewithlimits{argmin}}
\newcommand{\argminmax}{\operatornamewithlimits{arg \min\!/\!\max}}

\DeclareMathOperator*{\bigtimes}{\vartimes}
\newcommand{\scope}[1]{\setf{x}^{#1}}
\newcommand{\listf}[1]{{\mathcal{#1}}}
\newcommand{\setf}[1]{{\bf{#1}}}
\newcommand{\varf}[1]{{\it{#1}}}

\newcommand{\algref}[2]{\medskip\noindent\textbf{#1} \cite{#2}.\hspace{4pt}}
\newcommand{\domainref}[1]{\medskip\noindent\textbf{#1}.\hspace{4pt}}

\newcommand{\nth}[1]{$#1^\text{th}$}

\newcommand{\rev}[1]{#1}

\allowdisplaybreaks

\jairheading{1}{1993}{1-15}{6/91}{9/91}
\ShortHeadings{DCOP: Model and Applications Survey}
{Fioretto, Pontelli, \& Yeoh}
\firstpageno{1}

\begin{document}

\title{Distributed Constraint Optimization Problems and Applications: \\ A Survey}
\author{\name Ferdinando Fioretto \email fioretto@umich.edu \\
       \addr Department of Industrial and Operations Engineering \\
       University of Michigan \\
       Ann Arbor, MI  48109, USA
       \AND
       \name Enrico Pontelli \email epontell@cs.nmsu.edu \\
       \addr Department of Computer Science \\
       New Mexico State University \\
       Las Cruces, NM 88003, USA
       \AND
       \name William Yeoh \email wyeoh@wustl.edu \\
       \addr Department of Computer Science and Engineering \\
       Washington University in St.~Louis \\
       St. Louis, MO 63130, USA
}

\maketitle

\begin{abstract}
The field of \emph{multi-agent system (MAS)} is an active area of research within artificial intelligence, with an increasingly important impact in industrial and other real-world applications. In a MAS, autonomous agents interact to pursue personal interests and/or to achieve common objectives. \emph{Distributed Constraint Optimization Problems (DCOPs)} have emerged as a prominent agent model to govern the agents' autonomous behavior, where both algorithms and communication models are driven by the structure of the specific problem. During the last decade, several extensions to the DCOP model have been proposed to enable support of MAS in complex, real-time, and uncertain environments.

This survey provides an overview of the DCOP model, offering a classification of its multiple extensions and addressing both resolution methods and applications that find a natural mapping within each class of DCOPs. The proposed classification suggests several future perspectives for DCOP extensions, and identifies challenges in the design of  efficient resolution algorithms, possibly through the adaptation of strategies from different areas. 
\end{abstract}


\section{Introduction}
\label{sec:introduction}

An \emph{agent} can be defined as an entity (or computer program) that behaves autonomously within an arbitrary system in the pursuit of some goals \rev{\cite{wooldridge:09}}. 
A \emph{multi-agent system (MAS)} is a system where multiple agents interact in the pursuit of such goals. Within a MAS, agents may interact with each other \emph{directly,} via communication acts, or \emph{indirectly,} by acting on the shared environment. In addition, agents may decide to \emph{cooperate,} to achieve a common goal, or to \emph{compete,} to serve their own interests at the expense of other agents. In particular, agents may form cooperative \emph{teams}, which can in turn compete against other teams of agents. 
Multi-agent systems play an important role in 
\emph{distributed artificial intelligence,}  thanks to their ability to model a wide variety of real-world scenarios, where information and control are decentralized and distributed among a set of agents.


Figure~\ref{fig:dcop_mas} illustrates a MAS example. It represents a sensor network where a group of agents, equipped with sensors, seeks to determine the position of some targets. 
Agents may interact with each other and move away from the current position.
The figure depicts the targets as star-shaped objects. The dotted lines define an interaction graph and the directional arrows illustrate agents' movements. 
In addition, various events that obstruct the sensors of an agent may dynamically occur. For instance, the presence of an obstacle along the agent's sensing range may be detected after the agent's movement.

Within a MAS, an agent is: 
\bitemize
\item \emph{Autonomous}, as it operates without the direct intervention of humans or other entities and has full control over its own actions and internal state (e.g.,~in the example, an agent can decide to sense, to move, etc.);
\item \emph{Interactant}, in the sense that it interacts with other agents in order to achieve its objectives (e.g.,~in the example, agents may exchange information concerning results of sensing activities);
\item \emph{Reactive}, as it responds to changes that occur in the environment and/or to the requests from other agents
(e.g.,~in the example, agents may react with a move action to the sudden appearance of obstacles).
\item \emph{Proactive}, because of its goal-driven behavior, which allows the agent to take initiatives beyond the reactions in response to its environment.
\eitemize

\begin{figwindow}[0,r,%
{\includegraphics[width=.35\textwidth]{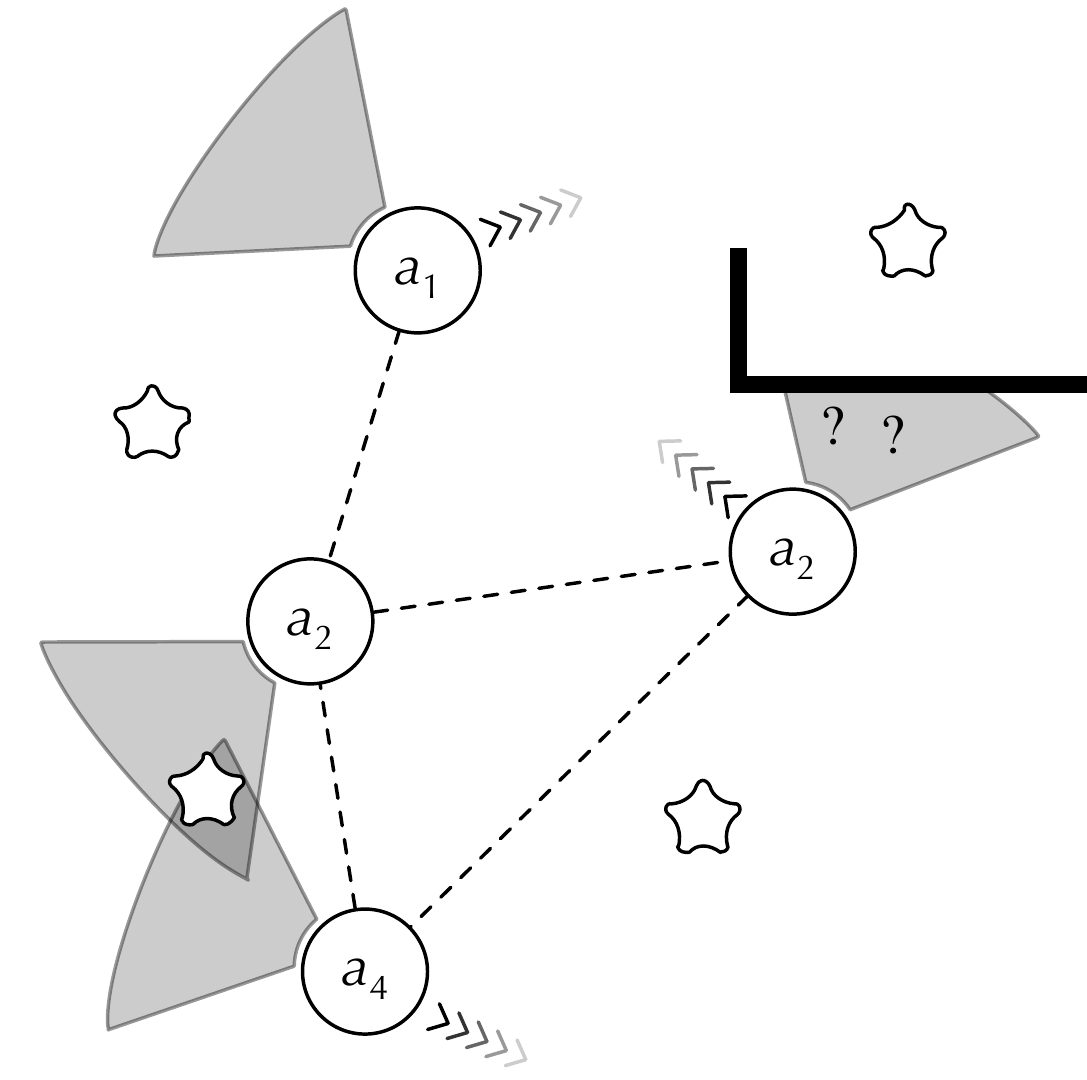}},%
{Illustration of a multi-agent system: Sensors (agents) seek to determine the position of the targets.\label{fig:dcop_mas}}]
Agent architectures are the fundamental mechanisms underlying the autonomous agent components, supporting their behavior in real-world, dynamic, and uncertain environments. 
Agent architectures based on \emph{decision theory}, \emph{game theory}, and \emph{constraint programming} have successfully been developed and are popular in the \emph{Autonomous Agents and Multi-Agent Systems (AAMAS)} community.

\hspace*{1.5em}\emph{Decision theory} \cite{raiffa:1968} assumes that the agent's actions and the environment are inherently uncertain and models such uncertainty explicitly. Agents acting in complex and dynamic environments are required to deal with various sources of uncertainty. The \emph{Decentralized Partially Observable Markov Decision Processes (Dec-POMDPs)} framework \cite{bernstein:02} is one of the most general multi-agent frameworks,  focused on team coordination in  presence of uncertainty about agents' actions and observations. The ability to capture a wide range of complex scenarios makes Dec-POMDPs of central interest within MAS research. However, the result of this generality is a high complexity for generating  optimal solutions. Dec-POMDPs are non-deterministic exponential (NEXP) complete \cite{bernstein:02}, even for two-agent problems, and scalability remains a critical challenge \cite{amato:13}. 
\end{figwindow}

\emph{Game theory} \cite{binmore:92} studies interactions between self-interested agents, aiming at maximizing the welfare of the participants. Some of the most compelling applications of game theory to MAS have been in the area of \emph{auctions} and \emph{negotiations} \cite{kraus:97,noriega:99,parsons:02}. These approaches model the trading process by which agents can reach agreements on matters of common interest, using market oriented  and cooperative mechanisms, such as reaching \emph{Nash equilibria.} Typical resolution approaches aim at deriving a set of equilibrium strategies for each agent, such that, when these strategies are employed, no agent can profit by unilaterally deviating from their strategies. A limitation of game theoretical-based approaches is the lack of an agent's ability to reason upon a global objective, as the underlying model relies on the interactions of self-interested agents.

\emph{Constraint programming} \cite{rossi:06} aims at solving decision-making problems formulated as optimization problems of some real-world objective. Constraint programs use the notion of constraints -- i.e.,~relations among entities of the problems (\emph{variables}) -- in both problem modeling and problem solving.  Constraint programming relies on  inference techniques that prevent the exploration of those parts of the solution search space whose assignments to variables are inconsistent with the constraints and/or dominated with respect to the objective function. \emph{Distributed Constraint Optimization Problems (DCOPs)} \cite{modi:05,petcu:05,gershman:09,yeoh:12} are problems where agents need to coordinate their value assignments, in a decentralized manner, to optimize their objective functions. DCOPs  focus on attaining a global optimum given the interaction graph of a collection of agents. This approach can be effectively used to model a wide range of problems. Problem solving and communication strategies are directly linked in DCOPs. This feature makes the algorithmic components of a DCOP suitable for exploiting the structure of the interaction graph of the agents to generate efficient solutions. 

The absence of a framework to model dynamic problems and uncertainty makes DCOPs unsuitable at solving certain classes of multi-agent problems, such as those characterized by action uncertainty and dynamic environments. However, since its original introduction, the DCOP model has undergone a process of continuous evolution to capture diverse characteristics of agent behavior and the environment in which they operate. Researchers have proposed a number of DCOP frameworks that differ from each other in terms of expressiveness and classes of problem they can target, extending the DCOP model to handle both dynamic and uncertain environments. However, current research has not explored how the different
DCOP frameworks relate to each other within the general MAS context, which is critical to understand:
\iOne What resolution methods could be borrowed from other MAS paradigms, and 
\iTwo What applications can be most effectively modeled  within each framework.
\rev{While there are important existing surveys for Distributed Constraint Satisfaction \cite{yokoo:00} and Distributed Constraint Optimization \cite{meisels:08}, this survey aims to comprehensively analyze and categorize the different DCOP frameworks proposed by the MAS community. We do so by presenting an extensive review of the DCOP model and its extensions, the different resolution methods, as well as a number of applications modeled within each particular DCOP extension. This analysis also provides opportunities to identify open challenges and discuss future directions in the general DCOP research area.}

\begin{table}[!tb]
\renewcommand{\arraystretch}{0.95}
\caption{Commonly Used Symbols and Notations\label{legend}}
	\small 
	\begin{tabular*}{0.95\textwidth}{c l @{\extracolsep{12pt}} l  @{\extracolsep{0.5cm}} l @{\extracolsep{12pt}} l}
    \hline
    \multicolumn{4}{c}{\textbf{List of key symbols}} \\
    \hline\hline
	   $a_i$ 			& Agent							& $\pi(\cdot)$		  & Projection operator \\ 
  	 $x_i$  			& Decision variable				& $p_i(\cdot)$			& Probability function \\
  	 $r_i$  			& Random variable		    	& $L_{a_i}$ 			& $a_i$'s local variables \\  
  	 $D_i$ 			& Domain of $x_i$ 			  	& $N_{a_i}$ 		  & $a_i$'s neighbors \\	
  	 $\Omega_i$ 		& Event space of $r_i$ 			& $C_{a_i}$ 			& $a_i$'s children \\	
   	 $f_i$ 			& \rev{Cost} function  				& $PC_i$			 	& $a_i$'s pseudo-children  \\
		 $\scope{i}$ 	& Scope of $f_i$  		 	& $P_{a_i}$ 			& $a_i$'s parent \\
   	 $m$ 				& Number of agents   			& $PP_{a_i}$ 		 & $a_i$'s pseudo-parents \\
   	 $n$ 				& Number of variables     		& $\alpha(f_i)$		  & Agents whose variables are in $\scope{i}$ \\  
   	 $q$ 				& Number of random variables   	& $E_{C}$				& Set of edges of the constraint graph \\ 
   	 $k$ 				& Number of \rev{cost} functions  	& $E_{T}$				& Tree edges of the pseudo-tree\\
	 $d$  	  			& Size of the largest domain	& $E_{F}$ & Set of edges of the factor graph \\
   	 $\setf{F}_g$		& Global objective function 	& $w^*$ 	   & Induced width of the pseudo-tree \\	
   	 $\vec{\setf{F}}$	& Vector of objective functions & $l$ 	   & Size of the largest neighborhood  \\	   
	 $F_i$ 			& Objective function in $\vec{\setf{F}}$	& $z$ 	   & Size of the largest local variable set \\
	 $\vec{\setf{F}}^\circ$ & Utopia point       					& $s$ 	   & Maximal sample size \\
	 $\bot$			& Infeasible value		   					& $p$ 	   & Size of the Pareto set \\  
	 $\sigma$			& Complete assignment			   				& $b$ 	   & Size of the largest bin \\
	 $\sigma_{V}$		& Partial assignment for the variables in $V \subseteq \setf{X}$	  & $\ell$   & \rev{Number of iterations of the algorithm} \\
     $\Sigma$			& State space              					& &  \\
    \hline
  	\end{tabular*}
\end{table}

This survey paper is organized as follows. 
The next section provides an overview on two relevant constraint satisfaction models and their generalization to the distributed cases. 
Section~\ref{sec:dcop_classification} introduces DCOPs, overviews the representation and coordination models adopted during the resolution of DCOPs, and it proposes a classification of the different variants of DCOPs based on the characteristics of the agents and the environment.
Section~\ref{sec:classical_dcop} presents the classical DCOP model as well as two notable extensions: One characterized by asymmetric cost functions and another by multi-objective optimization. 
Section~\ref{sec:dynamic_dcop} presents a DCOP model where the environment changes over time.  
Section~\ref{sec:probabilistic_dcop} discusses  DCOP models in which agents act under uncertainty and may have partial knowledge of the environment in which they act. 
Section~\ref{sec:noncooperative_dcop} discusses DCOP models in which agents are non-cooperative. 
For each of these models, the paper introduces their formal definitions, discusses related concepts, and describes several resolution algorithms. 
A summary of the various classes of problems discussed in this survey is given in 
Table \ref{tab:dcop_complexity}. 
Section~\ref{sec:applications} describes a number of applications that have been proposed in the DCOP literature. 
Section~\ref{sec:analysis} provides a critical review on the DCOP variants surveyed and focuses on their applicability in various settings. 
Additionally, it describes some potential future directions for research. 
Finally, Section~\ref{concl} provides concluding remarks. 
To facilitate the reading of this survey, Table~\ref{legend} summarizes the most commonly used symbols and notations.


\section{Overview of (Distributed) Constraint Satisfaction and Optimization}
\label{sec:dcop_overview}

\begin{figure}[t]
	\includegraphics[width=0.70\textwidth]{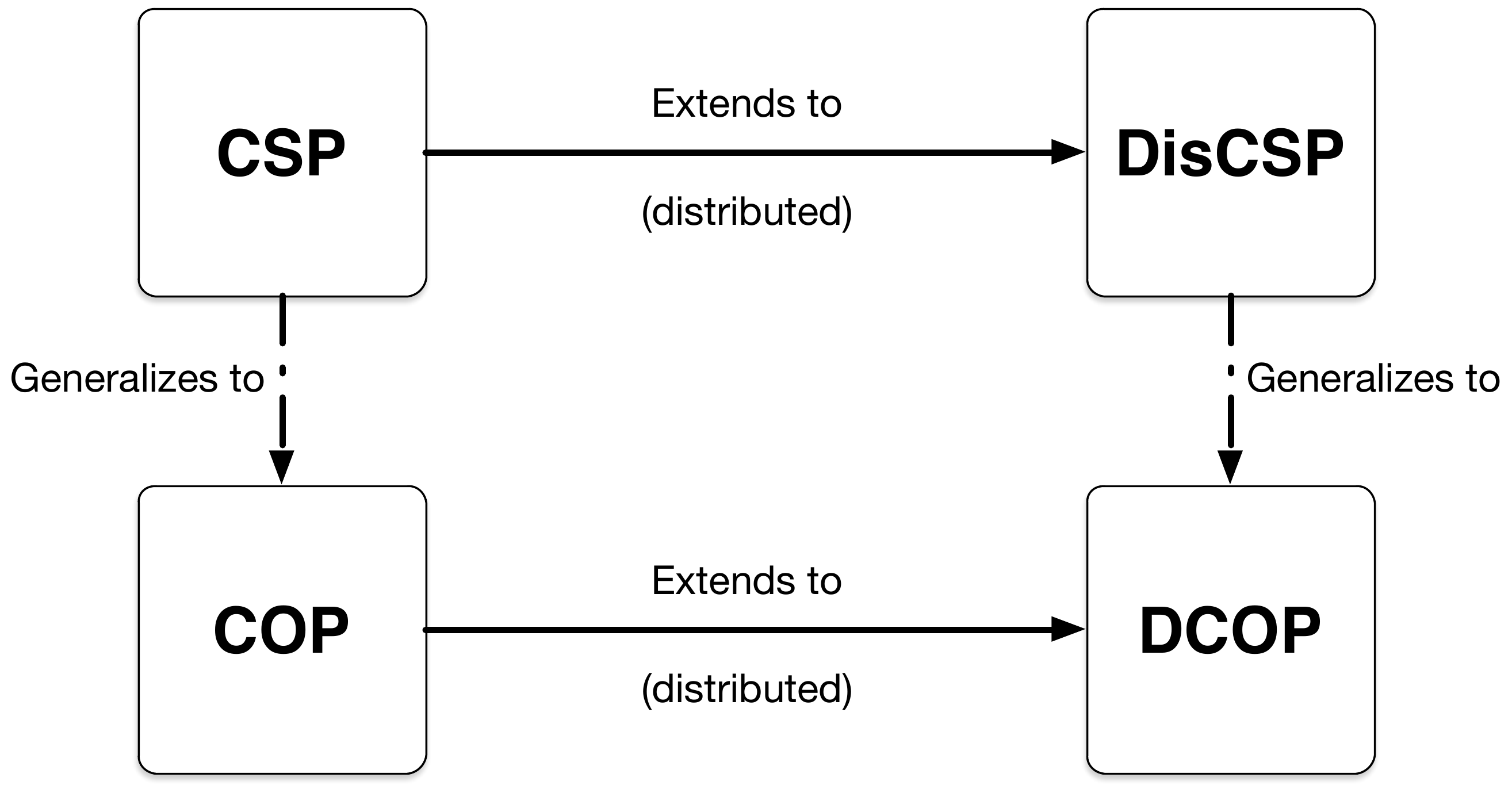}
	\caption{DCOP Problems as a Generalization and Extension of Constraint Satisfaction Problems \label{fig:csps}}
\end{figure}

This section provides an overview of several constraint satisfaction models, which form the foundation of DCOPs. Figure~\ref{fig:csps} illustrates the relations among these models.

\subsection{Constraint Satisfaction Problems}

\emph{Constraint Satisfaction Problems (CSPs)}~\rev{\cite{golomb:65,mackworth:85,apt:03,rossi:06}} are decision problems that involve the assignment of values to variables, under a set of specified constraints on how variable values should be related to each other. A number of problems can be formulated as CSPs, including resource allocation, vehicle routing, circuit diagnosis, scheduling, and bioinformatics. 
Over the years, CSPs have become the paradigm of choice to address difficult combinatorial problems, drawing and integrating insights from diverse domains, including artificial intelligence and operations research \cite{rossi:06}.

A CSP is a tuple $\langle \setf{X}, \setf{D}, \setf{C} \rangle$, where: 
\bitemize
\item $\setf{X} \is \{x_1, \ldots, x_{n}\}$ is a finite set of variables. %
\item $\setf{D} \is \{D_1, \ldots, D_{n}\}$ is a set of finite domains for the variables in $\setf{X}$, with $D_i$ being the set of possible values for the variable $x_i$. 
\item $\setf{C}$ is a finite set of constraints over subsets of $\setf{X}$, where a constraint $c_i$, defined on the \rev{$k$ variables $x_{i_1}, \ldots, x_{i_k}$, is a relation $c_i \subseteq \bigtimes_{j=1}^k D_{i_j}$, where $\{i_1,\dots,i_k\} \subseteq \{1,\dots,n\}$. The set of variables $\scope{i} = \{ x_{i_1}, \ldots, x_{i_k} \}$ is referred to as the \emph{scope} of $c_i$.\footnote{The presence of a fixed ordering of variables is assumed.} 
$c_i$ is called a \emph{unary} constraint if $k=1$ and a \emph{binary} constraint if $k=2$. For all other values of $k$, the constraint is called a \emph{k}-ary constraint.\footnote{A constraint with $k=3$ is also called a \emph{ternary} constraint and a constraint with $k=n$ is also called a \emph{global} constraint.}}
\eitemize

A \rev{\emph{partial assignment}} is a value assignment for a proper subset of variables from $\setf{X}$ that is consistent with their respective domains, i.e.,~it is a partial function $\sigma: \setf{X} \rightarrow \bigcup_{i=1}^n D_i$ such that, for each $x_j\in \setf{X}$, if $\sigma(x_j)$ is defined, then $\sigma(x_j) \in D_j$. 
\rev{An assignment is \emph{complete}} if it assigns a value to each variable in $\setf{X}$. 
The notation $\sigma$ is used to denote a complete assignment, and, for a set of variables $\setf{V} = \{x_{i_1}, \ldots, x_{i_h}\} \subseteq \setf{X}$,  $\sigma_\setf{V} = \tuple{\sigma(x_{i_1}), \ldots, \sigma(x_{i_h}) }$ to denote the projection of the values in $\sigma$ associated to the variables in $\setf{V}$, where $i_1 < \cdots < i_h$. 
\rev{The goal in a CSP is to find a complete assignment $\sigma$ such that, for each $c_i \in \setf{C}$, $\sigma_{\scope{i}} \in c_i$, 
that is, a complete assignment that satisfies all the problem constraints. Such a complete assignment is called a \emph{solution} of the CSP.}

\subsection{Weighted Constraint Satisfaction Problems}

A solution of a CSP must satisfy all of its constraints. In many practical cases, however, it is desirable to consider complete assignments whose constraints can be violated according to a violation degree. The \emph{Weighted Constraint Satisfaction Problem (WCSP)} \cite{shapiro:81,larrosa:02} was introduced to capture this property. WCSPs are problems whose constraints are considered as preferences that specify the extent of satisfaction (or violation) of the associated constraint.

A WCSP is a tuple $\langle \setf{X}, \setf{D}, \setf{F} \rangle$, where $\setf{X}$ and $\setf{D}$ are the set of variables and their domains as defined in a CSP, and $\setf{F}$ is a set of \emph{weighted constraints}. 
A weighted constraint $f_i \in \setf{F}$ is a function $f_i : \bigtimes_{x_j \in \scope{i}} D_j \to \mathbb{R}^+ \cup \{\bot\}$, where $\scope{i} \subseteq \setf{X}$ is the scope of $f_i$ \rev{and $\bot$ is a special element used to denote that a given combination of values for the variables in $\scope{i}$ is not allowed, and it has the property that $a + \bot = \bot + a = \bot$, for all $a \in \mathbb{R}^+$. 
The \emph{cost} of an assignment $\sigma$ is the sum of the evaluation of the constraints involving all the variables in $\sigma$. 
A \emph{solution} is a complete assignment with cost different from $\bot$, and an optimal solution is a solution with minimal cost.}

Thus, a WCSP is a generalization of a CSP which, in turn, can be seen as a WCSP whose constraints use exclusively the costs $0$ and $\bot$. \rev{The terms WCSP and \emph{Constraint Optimization Problem (COP)} have been used interchangeably in the literature and the use of the latter term has been widely adopted in the recent years.}

\subsection{Distributed Constraint Satisfaction Problems}

When the elements of a CSP are distributed among a set of autonomous agents, the resulting model is referred to as a \emph{Distributed Constraint Satisfaction Problem (DisCSP)} \cite{yokoo:98,yokoo:01}.
A DisCSP is a tuple $\langle\setf{A}, \setf{X}, \setf{D}, \setf{C}, \alpha\rangle$, where $\setf{X}$, $\setf{D}$, and $\setf{C}$ are the set of variables, their domains, and the set of constraints, as defined in a CSP; 
$\setf{A} \is \{a_1, \ldots, a_{m}\}$ is a finite set of autonomous agents; and 
$\alpha: \setf{X} \to \setf{A}$ is a surjective function, from variables to agents, which assigns the control of each variable $x \in \setf{X}$ to an agent $\alpha(x)$. 
The goal in a DisCSP is to find a \rev{complete assignment} that satisfies all the constraints of the problem.

DisCSPs can be seen as an extension of CSPs to the multi-agent case, where agents communicate with each other to assign values to the variables they control so as to satisfy all the problem constraints. 
For a survey on the topic, the interested reader is referred to \cite{rossi:06} (Chapter 20).

\subsection{Distributed Constraint Optimization Problems}
Similar to the generalization of CSPs to \rev{COPs}, the \emph{Distributed Constraint Optimization Problem (DCOP)} model \cite{modi:05,petcu:05,gershman:09,yeoh:12} emerges as a generalization of the DisCSP model, where constraints specify a degree of preference over their violation, rather than a Boolean satisfaction metric.  
DCOPs can also be viewed as an extension of the COP framework to the multi-agent case, where agents control variables and constraints, and need to coordinate the value assignment for the variables they control so as to optimize a global objective function. 
The DCOP framework is formally introduced in the next section.

\section{DCOP Classification}
\label{sec:dcop_classification}

The DCOP model has undergone a process of continuous evolution to capture diverse characteristics of the agent behavior and the environment in which agents operate. 
This section proposes a classification of DCOP models from a multi-agent systems perspective.
It accounts for the different assumptions made about the behavior of the agents and their interactions with the environment. The classification is
based on the following elements (summarized in Table~\ref{tab:dcop_classification}):

\begin{table}[t]
	\renewcommand{\arraystretch}{1.00}
	{
	   \small \centering
	   \begin{tabular}{| l l | r r | }
	   	\multicolumn{2}{c}{\sc \textbf{Element}} & \multicolumn{2}{c}{\sc 
		\textbf{Characterization}} \\
			\cline{1-4}			
			\multirow{2}{*}{\sc Agent(s)} & {\sc Behavior} & Deterministic & Stochastic \\
			\multicolumn{1}{|c}{} & {\sc Knowledge} & Total & Partial \\
			\multicolumn{1}{|c}{} & {\sc Teamwork} & Cooperative & Competitive \\
			\cline{1-4}
			\multirow{2}{*}{\sc Environment} & {\sc Behavior} & Deterministic 
			& Stochastic \\
			\multicolumn{1}{|c}{} & {\sc Evolution} & Static & Dynamic \\			
			\cline{1-4}
	  \end{tabular}
	  }
	\caption{DCOP Classification Elements \label{tab:dcop_classification}}
\end{table}

\bitemize
\item {\bf Agent Behavior:} This parameter captures the stochastic nature of the \emph{effects} of  an action being executed.  These effects can be either \emph{deterministic} or \emph{stochastic}. 

\item {\bf Agent Knowledge:} This parameter captures the \emph{knowledge} of an agent about its own state and the environment. It can be \emph{total} or \emph{partial} (i.e.,~\emph{incomplete}). 

\item {\bf Agent Teamwork:} This parameter characterizes the approach undertaken by (teams of) agents to solve a distributed  problem. It can be either a \emph{cooperative} or a \emph{competitive} resolution approach. 
In the former class, all agents cooperate to achieve a common goal (i.e.,~they all optimize a global objective function). In the latter class, each agent (or team of agents) seeks to achieve its own individual goal (i.e.,~each agent optimizes its individual objective functions). 
	
\item {\bf Environment Behavior:} This parameter captures the exogenous properties of the environment. 
The response of the environment to the execution of an action can be either \emph{deterministic} or \emph{stochastic}.
	
\item {\bf Environment Evolution:} This parameter captures whether the DCOP does not change over time (\emph{static}) or it changes over time (\emph{dynamic}). 
	
\eitemize

\begin{figure}[!tb]
\hspace{-10pt}
\begin{floatrow}
	\capbtabbox{
	\renewcommand{\arraystretch}{1.8}
	\small \centering
	\resizebox{0.45\textwidth}{!}
	 {
	\begin{tabular}{ll | r | c | c |}
		\multicolumn{1}{c}{\multirow{5}{0em}{\begin{turn}{90}\bf Environment\end{turn}}} &
      	\multicolumn{1}{c}{\multirow{5}{*}{\begin{turn}{90}\bf Evolution\end{turn}}}
			& \emptycell & \multicolumn{2}{c}{\bf Environment Behavior} \\
		\cline{4-5}
		\emptycell & \emptycell &  	  & {\sc Deterministic}  & \sc{Stochastic} \\
		\cline{3-5}
		\emptycell & & {\sc Static}   & {Classical DCOP} & {Probabilistic DCOP}\\
		\cline{3-5}
		\emptycell & & {\sc Dynamic} & {Dynamic DCOP } & {\it ---} \\
		\cline{3-5}
	\end{tabular}
	}
	\vspace*{3.8em}
	}
	{\caption{DCOPs Models \label{tab:dcop_models}}}
	\ffigbox{
	\includegraphics[width=0.5\textwidth]{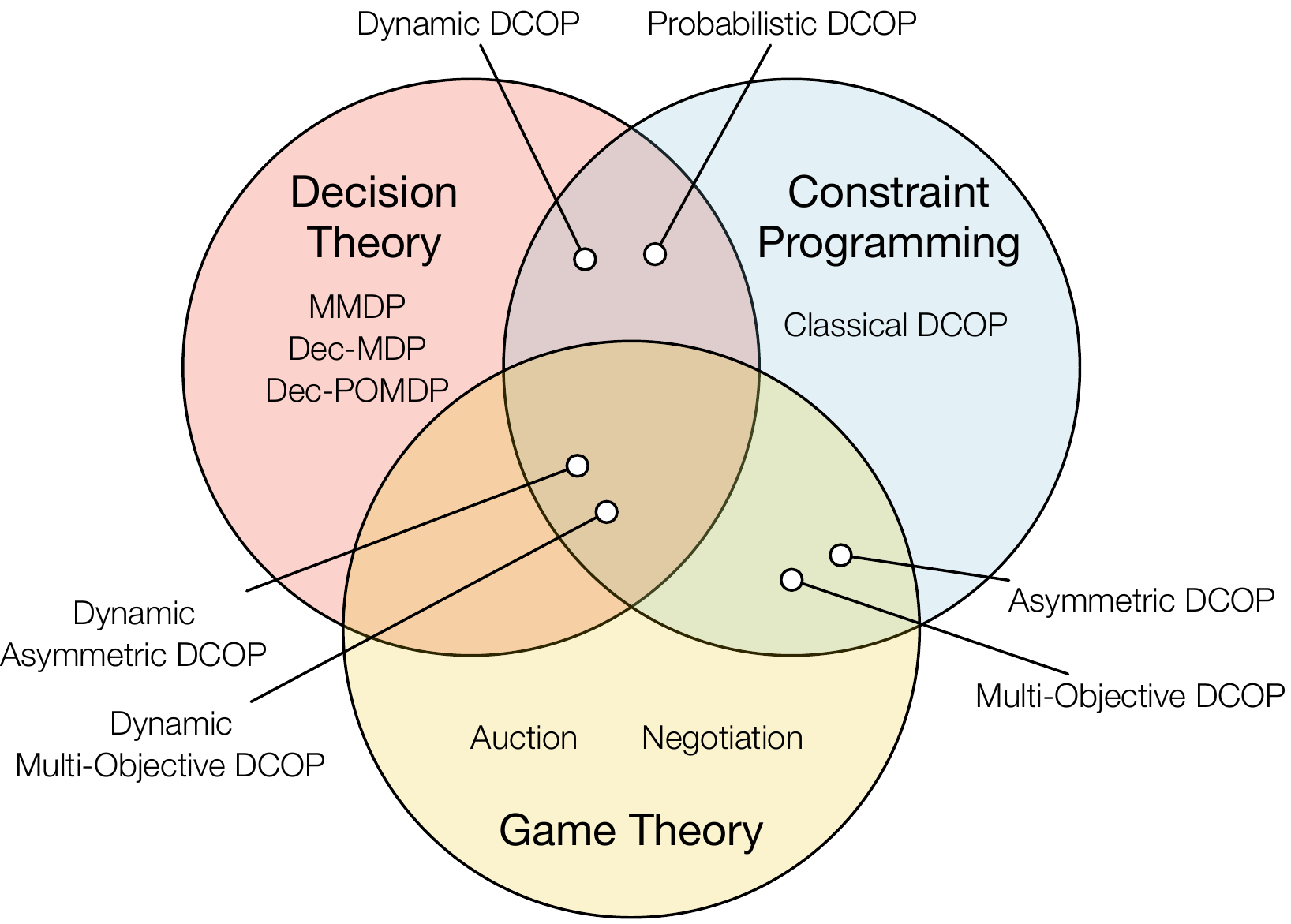}
	}
	{\caption{DCOPs Within a MAS Perspective
	 \label{fig:mas_circle}}}
\end{floatrow}
\end{figure}

Figure~\ref{fig:mas_circle} illustrates a categorization of the DCOP models proposed to date from a MAS perspective. This survey focuses on the DCOP models  proposed at the junction  of  constraint programming, game theory, and decision theory. 
The \emph{classical} DCOP model is directly inherited from constraint programming as it extends the WCSP model to a distributed setting. It is characterized by a static model, a deterministic environment and agent behavior, a total agent knowledge, and a cooperative agent teamwork. Game theoretical concepts explored in the context of auctions and negotiations have influenced the DCOP framework leading to the development of the Asymmetric DCOP and the Multi-Objective DCOP. 
The DCOP framework has also borrowed fundamental decision theoretical concepts related to modeling uncertain and dynamic environments, resulting in models like the Probabilistic DCOP and the Dynamic DCOP. Researchers from the DCOP community have also designed solutions that inherit from all of the three communities.

The next sections describe the different DCOP frameworks, starting with classical DCOPs before proceeding to its various extensions. The survey focuses on a categorization based on three dimensions: \emph{Agent knowledge}, \emph{environment behavior}, and \emph{environment evolution}. It assumes a \emph{deterministic agent behavior}, a \emph{fully cooperative agent teamwork}, and a \emph{total agent knowledge} (unless otherwise specified), as they are, by far, common assumptions adopted by the DCOP community. The DCOP models associated to this categorization are summarized in Table~\ref{tab:dcop_models}. The bottom-right entry of the table is left empty, indicating a promising model with dynamic and uncertain environments that, to the best of our knowledge, has not been explored yet. There has been only a modest amount of effort in modeling the different aspects of teamwork within the DCOP community. Section~\ref{sec:noncooperative_dcop} describes a formalism that has been adopted to model DCOPs with mixed cooperative and competitive agents. 

\section{Classical DCOP}
\label{sec:classical_dcop}

With respect to the proposed categorization, in the \emph{classical DCOP} model \cite{modi:05,petcu:05,gershman:09,yeoh:12} 
the agents are fully cooperative and have deterministic behavior and total knowledge. 
Additionally, the environment is static and deterministic. 
This section reviews the formal definitions of classical DCOPs, presents some relevant 
solving algorithms, and provides details of selected variants of classical DCOPs of 
particular interest.

\subsection{Definition}
\label{sec:classical_dcop:definition}

\label{def:dcop}
A classical DCOP is described by a tuple $P=\langle \setf{A, X, D}, \setf{F}, \alpha \rangle$, where:
\bitemize
\item $\setf{A} \is \{a_1, \ldots, a_{m}\}$ is a finite set of agents. %
\item $\setf{X} \is \{x_1, \ldots, x_{n}\}$ is a finite set of variables, with $n \geq m$. %
\item $\setf{D} \is \{D_1, \ldots, D_{n}\}$ is a set of finite domains for the variables in $\setf{X}$, with $D_i$ being the domain of variable $x_i$. %
\item $\setf{F} \is \{f_1, \ldots, f_{k}\}$ is a finite set of \emph{cost functions}, with $f_i : \bigtimes_{x_j \in \scope{i}} D_j \to 
\mathbb{R}^+ \cup \{\bot\}$, where similar to WCSPs, $\scope{i} \subseteq \setf{X}$ is the set of variables relevant to  $f_i$, referred to as the \emph{scope} of $f_i$. 
The \emph{arity} of a cost function is the number of variables in its scope. 
Each cost function $f_i$ represents a factor in a  \emph{global objective function} $\setf{F}_g(\setf{X}) \is \sum_{i=1}^k f_i(\scope{i})$. In the DCOP literature, the cost functions $f_i$ are also called \emph{constraints}, \emph{utility functions}, or \emph{reward functions}. %
\item $\alpha: \setf{X} \to \setf{A}$ is a total and onto function, from variables to agents, which assigns the control of each variable $x \in \setf{X}$ to an agent $\alpha(x)$. 
\eitemize

With a slight abuse of notation,  $\alpha(f_i)$ will be used to denote the set of agents whose variables are involved in the scope of $f_i$, i.e.,~$\alpha(f_i) = \{\alpha(x)\:|\: x\in \scope{i}\} $. 
\rev{A \emph{partial} assignment is a value assignment for a proper subset of variables of $\setf{X}$. An assignment is \emph{complete} if it assigns 
a value to each variable in $\setf{X}$.}
For a given \rev{complete assignment} $\sigma$, we say that a cost function $f_i $ is satisfied by $\sigma$ if $f_i(\sigma_{\scope{i}}) \neq \bot$.
\rev{A complete assignment is a \emph{solution} of a DCOP if it satisfies all its cost functions.} 
The goal in a DCOP is to find a solution that \rev{minimizes} the total \rev{problem cost expressed by its cost functions}:\footnote{\rev{Alternatively, one can define a maximization problem by substituting the $\argmin$ operator in Equation~\ref{eq:dcop_goal} with $\argmax$. Typically, if the objective functions are referred to as utility functions or reward functions, then the DCOP is a maximization problem. Conversely, if the objective functions are referred to as cost functions, then the DCOP is a minimization problem.}}
\begin{equation}
	\label{eq:dcop_goal}
	\sigma^*  \coloneqq \argmin_\setf{\sigma \in \Sigma} \setf{F}_g(\sigma) = \argmin_\setf{\sigma \in \Sigma} 
			         \sum_{f_i \in \setf{F}} f_i(\sigma_{\scope{i}}), 
\end{equation}
where $\Sigma$ is the \emph{state space}, defined as the set of all possible solutions. 

Given an agent $a_i$, $L_{a_i} \is \{x_j \in \setf{X} \:\: |\:\: \alpha(x_j) \is a_i\}$ denotes the set of variables controlled by agent $a_i$, or its \emph{local variables}, and $N_{a_i} \is \{ a_i' \in \setf{A} \st a_i \neq a_i',\ \exists f_j \in \setf{F},\ x_r, x_s \in \scope{j} ,\ \alpha(x_r) \is a_i \land \alpha(x_s) \is a_i' \}$ denotes the set of its \emph{neighboring agents}. 
A cost function $f_i$ is said to be \emph{hard} if $\forall \sigma \in \Sigma$ we have
that $f_i(\sigma_{\scope{i}}) \in \{0,\bot\}$. Otherwise, the cost function is said to be \emph{soft}.

Finding an optimal solution for a classical DCOP is known to be NP-hard \cite{modi:05}.

\subsection{DCOP: Representation and Coordination}
\label{sec:coord}

Representation in DCOPs plays a fundamental role, both from an agent coordination perspective and from an algorithmic perspective. 
This section discusses the most predominant representations adopted in various DCOP algorithms. 
It starts by describing some widely adopted assumptions regarding agent knowledge and coordination, which will 
apply  throughout this document, unless otherwise stated:
\benumerate
\item[\iOne] A variable and its domain are known exclusively to the agent controlling it and its neighboring agents. 
\item[\iTwo] Each agent knows the values of the cost function involving at least one of its local variables. No other agent has knowledge about such cost function. 
\item[\iThree] Each agent knows (and it may communicate with) exclusively its own neighboring agents. 
\eenumerate

\subsubsection{Constraint Graph}
Given a DCOP $P$,  $G_P = (\setf{X}, E_C)$ is the \emph{constraint graph} of $P$, where an undirected edge $\{x,y\} \in E_C$ exists if and only if there exists $f_j \in \setf{F}$ such that $\{x,y\} \subseteq \scope{j}$. A constraint graph is a standard way to visualize a DCOP instance. It underlines the agents' locality of interactions and therefore it is commonly adopted by DCOP resolution algorithms. 

Given an ordering $o$ on $\setf{X}$, a variable $x_i$ is said to have a higher \emph{priority} with respect to a variable $x_j$ if $x_i$ appears before $x_j$ in $o$. Given a constraint graph $G_P$ and an ordering $o$ on its nodes, the \emph{induced graph} $G_P^*$ on $o$ is the graph obtained by connecting nodes, processed in increasing order of priority, to all their higher-priority neighbors. For a given node, the number of higher-priority neighbors is referred to as its \emph{width}. The \emph{induced width} $w_o^*$ of $G_P$ is the maximum width over all the nodes of $G_P^*$ on ordering $o$.

\begin{figure}[!t]
		\includegraphics[width=0.95\textwidth]{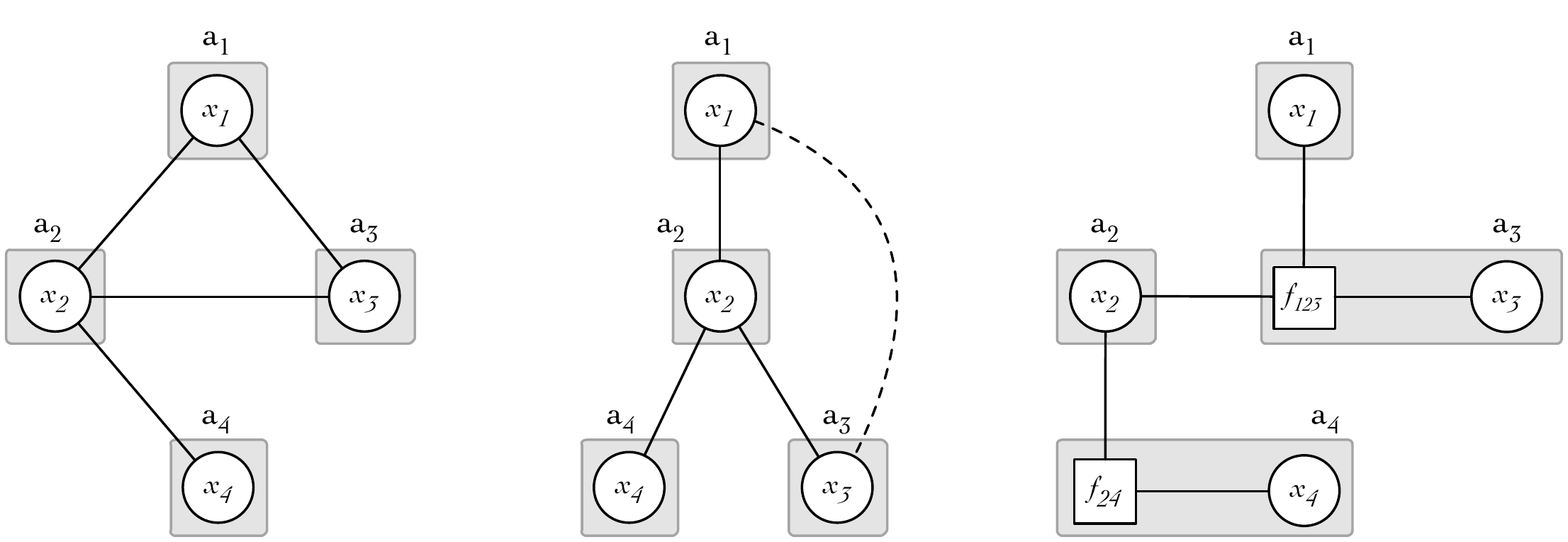}\\
	\noindent		\small
	\hspace{-0.3in} (a) Constraint Graph  
	\hspace{0.6in} (b) Pseudo-Tree 
  	\hspace{0.9in} (c) Factor Graph
  	\caption{DCOP Representations 
	\label{fig:dcop_representations}}
\end{figure}

Figure~\ref{fig:dcop_representations}(a) shows an example constraint graph of a DCOP with four agents $a_1$ through $a_4$, each controlling one variable with domain \{0,1\}. There are two cost functions: a $k$-ary cost function $f_{123}$ with scope $\scope{123} = \{x_1, x_2, x_3\}$ and represented by a clique among $x_1, x_2$, and $x_3$; and a binary cost function $f_{24}$ with scope $\scope{24} = \{x_2,x_4\}$.

\subsubsection{Pseudo-Tree}
A number of DCOP algorithms require a partial ordering among the agents. In particular, when such an order is 
derived from  a depth-first search (DFS) exploration, the resulting structure is known as a \emph{(DFS) pseudo-tree}. A {pseudo-tree} arrangement for a DCOP $P$ is a subgraph $T_P \is \langle \setf{X}, E_T \rangle$ of $G_P$ such that $T_P$ 
is a spanning tree of $G_P$ -- i.e.,~a connected subgraph of $G_P$ containing all the nodes and being a rooted tree -- with
the following additional condition: for each $x, y \in \setf{X}$, if $\{x,y\} \subseteq \scope{i}$ for some $f_i\in \setf{F}$, then $x,y$ appear in the same branch of $T_P$ (i.e.,~$x$ is an ancestor of $y$ in $T_P$ or vice versa).
Edges of $G_P$ that are \emph{in} (respectively \emph{out} of) $E_T$ are called \emph{tree edges} (respectively \emph{backedges}). The tree edges connect parent-child nodes, while backedges connect a node with its \emph{pseudo-parents} and its \emph{pseudo-children}. 
The \emph{separator} of an agent $a_i$ is the set containing all the ancestors of $a_i$ in the pseudo-tree (through tree edges or backedges) that are connected to $a_i$ or to one of its descendants.
The notation $C_{a_i}$, $PC_{a_i}$, $P_{a_i}$, and $PP_{a_i}$ will be used to indicate the set of children, pseudo-children, parent, and pseudo-parents of the agent $a_i$. 

Both constraint graph and pseudo-tree representations cannot deal explicitly with \rev{$k$-ary cost functions} (with $k>2$). A typical artifact to deal with such cost functions in a pseudo-tree representation is to introduce a virtual variable that monitors the value assignments for all the variables in the scope of the cost function, and generates the \rev{cost} values \cite{bowring:06} -- the role of the virtual variables can be delegated to one of the variables participating in the cost function \cite{pecora:06,matsui:08a}.

Figure~\ref{fig:dcop_representations}(b) shows one possible pseudo-tree of the example DCOP in Figure~\ref{fig:dcop_representations}(a), where $C_{a_1}  \is \{x_2\}$, $PC_{a_1} \is \{x_3\}$, $P_{a_4}  \is \{x_2\}$, and $PP_{a_3}  \is \{x_1\}$. The solid lines are tree edges and dotted lines are backedges.

\subsubsection{Factor Graph}
Another way to represent DCOPs is through a \emph{factor graph} \cite{kschischang:01}. A factor graph is a bipartite graph used to represent the factorization of a  function. In particular, given the global objective function $\setf{F}_g$, the corresponding factor graph $F_P \is \langle \setf{X}, \setf{F}, E_F \rangle$ is composed of variable nodes $x_i \in \setf{X}$, factor nodes $f_j \in \setf{F}$, and edges $E_F$ such that there is an undirected edge between
 factor node $f_j$ and variable node $x_i$ if $x_i \in \scope{j}$. 

Factor graphs can handle \rev{$k$-ary cost functions} explicitly. To do so, they use a similar  method as the one adopted within pseudo-trees with such cost functions: They delegate the control of a factor node to one of the agents controlling a variable in the scope of the cost function. From an algorithmic perspective, the algorithms designed over factor graphs can directly handle \rev{$k$-ary cost functions}, while algorithms designed over pseudo-trees require changes in the algorithm design so to delegate the control of the \rev{$k$-ary cost functions} to some particular entity.

Figure~\ref{fig:dcop_representations}(c) shows the factor graph of the example DCOP in Figure~\ref{fig:dcop_representations}(a), where each agent $a_i$ controls its variable $x_i$ and, in addition, $a_3$ controls the cost function $f_{123}$ and $a_4$ controls cost function $f_{24}$.

\subsection{Algorithms}
\label{sec:classical_dcop:algorithms}

The field of classical DCOPs is mature and a number of different resolution algorithms have been proposed. DCOP algorithms can be classified as being either \emph{complete} or \emph{incomplete}, based on whether they can guarantee the optimal solution or they trade optimality for shorter execution times, producing \rev{near-optimal} solutions. 
They can also be characterized based on their runtime characteristics, 
their memory requirements, and their communication requirements (e.g.,~the number and size of messages that they send and whether they communicate with their neighboring agents only or also to non-neighboring agents). Table~\ref{tab:algorithm_characteristics} tabulates the properties of a number of key DCOP algorithms that will be surveyed in Sections~\ref{sec:DCOP_complete} and~\ref{sec:DCOP_incomplete}. 
An algorithm is said \emph{anytime} if it can return a valid solution even if the DCOP agents are interrupted at any time before the algorithm terminates. Anytime algorithms are expected to seek for solutions of increasing quality as they keep running \cite{zivan:14}. 

All these algorithms were originally developed under the assumption that each agent controls exactly one variable. The description of their properties will follow the same assumption. These properties may change when generalizing the algorithms to allow for agents to control multiple variables, but they will depend on how the algorithms are generalized. Throughout this document, the following notation will be often adopted when discussing the complexity of the algorithms: 
\bitemize
\item $n = |\setf{X}|$ refers to the number of variables in the problem; in Table~\ref{tab:algorithm_characteristics}, $n$ also refers to the number of agents in the problem since each agent has exactly one variable;
\item $d = \max_{D_i \in \setf{D}} \size{D_i}$ refers to the size of the largest domain; 
\item $w^*$ refers to the induced width of the pseudo-tree;
\item $l = \max_{a_i \in \setf{A}} | N_{a_i}|$ refers to the largest number of neighboring agents; and
\item $\ell$ refers to the number of iterations in incomplete algorithms.
\eitemize

In addition, each of these classes can be categorized into several groups, depending on the degree of locality exploited by the algorithms,  the way local information is updated, and  the type of exploration process adopted. These different categories are described next.  

\begin{table}[t]
	\resizebox{1.0\linewidth}{!}
	  {
    \begin{tabular}{|c||c|c||c|c||c||c|c|c|}
    \hline
    \multirow{2}{*}{Algorithm} & \multicolumn{2}{c||}{Quality Characteristics} & \multicolumn{2}{c||}{Runtime Characteristics} & Memory & \multicolumn{3}{c|}{Communication Characteristics} \\
    		& Optimal? & Error Bound? & Complexity & Anytime? & per Agent & \# Messages & Message Size & Local Communication? \\
    \hline
    \hline
    SyncBB 
    & $\checkmark$ & $\checkmark$& $O(d^n)$ & $\checkmark$ & $O(n)$ & $O(d^n)$ & $O(n)$ & $\times$ \\
    AFB 
    & $\checkmark$ & $\checkmark$& $O(d^n)$ & $\checkmark$ & $O(n)$ & $O(d^n)$ & $O(n)$ & $\times$ \\
    ADOPT 
    & $\checkmark$ & $\checkmark$& $O(d^n)$ & $\times$ & $O(n\!+\!ld)$ & $O(d^n)$ & $O(n)$ & $\checkmark$ \\
    ConcFB 
    & $\checkmark$ & $\checkmark$& $O(d^n)$ & $\checkmark$ & $O(n)$ & $O(d^n)$ & $O(n)$ & $\times$ \\
    DPOP 
    & $\checkmark$ & $\checkmark$& $O(d^{w^*})$ & $\times$ & $O(d^{w^*})$ & $O(n)$ & $O(d^{w^*})$ & $\checkmark$ \\
    OptAPO 
    & $\checkmark$ & $\checkmark$& $O(d^n)$ & $\times$ & $O(ld)$ & $O(d^n)$ & $O(d\!+\!n)$ & $\times$ \\
    Max-Sum
    & $\times$ & $\times$& $O(\ell d^{\,l})$ & $\times$ & $O(d^{\,l})$ & $O(\ell nl)$ & $O(d)$ & $\checkmark$ \\
    Region Optimal 
    & $\times$ & $\checkmark$& $O(\ell d^{w^*})$ & $\checkmark$ & $O(d^{w^*})$ & $O(\ell n^2)$ & $O(d^{w^*})$ & $\times$ \\
    MGM 
    & $\times$ & $\times$& $O(\ell ld)$ & $\checkmark$ & $O(l)$ & $O(\ell nl)$ & $O(1)$ & $\checkmark$ \\
    DSA 
    & $\times$ & $\times$& $O(\ell ld)$ & $\checkmark$ & $O(l)$ & $O(\ell nl)$ & $O(1)$ & $\checkmark$ \\
    DUCT 
    & $\times$ & $\checkmark$& $O(\ell ld)$ & $\checkmark$ & $O(d^{w^*})$ & $O(\ell n)$ & $O(n)$ & $\checkmark$ \\
    D-Gibbs 
    & $\times$ & $\checkmark$& $O(\ell ld)$ & $\checkmark$ & $O(l)$ & $O(\ell nl)$ & $O(1)$ & $\checkmark$ \\
    \hline
    \end{tabular}
  }
\caption{\rev{Quality, Runtime, Memory, and Communication Characteristics of DCOP Algorithms}}
\label{tab:algorithm_characteristics}
\end{table}

\subsubsection {Partial Centralization}
In general, the DCOP solving process is decentralized, driving DCOP algorithms to follow the agent knowledge and communication restrictions described in Section~\ref{sec:coord}. However, some algorithms explore methods to centralize the decisions to be taken by a group of agents, by delegating them to one of the agents in the group. 
These algorithms explore the concept of \emph{partial centralization} \cite{hirayama:97,mailler:04,petcu:07b}, and thus they are classified as \emph{partially centralized} algorithms. 
Typically, partial centralization improves the algorithms' performance allowing agents to coordinate their local assignments more efficiently. However, such performance enhancement comes with a loss of information privacy, as the centralizing agent needs to be granted access to the local subproblem of other agents in the group \cite{greenstadt:07,mailler:04}. 
\rev{In contrast, \emph{fully} decentralized algorithms inherently reduce the amount of information privacy at cost of a larger communication effort}.

\subsubsection{Synchronicity}
DCOP algorithms can enhance their effectiveness by exploiting distributed and parallel processing. 
Based on the way the agents update their local information, DCOP algorithms are classified as \emph{synchronous} or \emph{asynchronous}.
Asynchronous algorithms allow agents to update the assignment for their variables based solely on their local view of the problem, and thus independently from the actual decisions of the other agents \cite{modi:05,farinelli:08,gershman:09}.  
In contrast, synchronous algorithms constrain the agents decisions to follow a particular order, typically enforced by the representation structure adopted \cite{mailler:04,petcu:05,pearce:07}. 

Synchronous algorithms tend to delay the actions of some agents guaranteeing that their local view of the problem is always consistent with that of the other agents.
In contrast, asynchronous algorithms tend to minimize the idle-time of the agents, which in turn can react quickly to each message being processed; however, they provide no guarantee on the consistency of the state of the local view of each agent. 
Such effect has been studied by \citeA{peri:13}, concluding that inconsistent agents' views may cause a negative impact on network load and algorithm performance, and that introducing some level of synchronization may be beneficial for some algorithms, enhancing their performance.

\subsubsection{Exploration Process}
The resolution process adopted by each algorithm can be classified in three categories \cite{yeoh:10b}:
\bitemize
\item \emph{Search-based algorithms} are based on the use of search techniques to explore the space of possible solutions. These algorithms are often derived from corresponding search techniques developed for centralized AI search problems, such as best-first search and depth-first search.
	
\item \emph{Inference-based algorithms} are derived from  dynamic programming and belief propagation techniques. These algorithms allow agents to exploit the structure of the constraint graph to aggregate \rev{costs} from their neighbors, effectively reducing the problem size at each step of the algorithm.
   
\item \emph{Sampling-based algorithms}  are incomplete approaches that sample the search space to approximate a function (typically, a probability distribution) as a product of statistical inference.
\eitemize

\begin{figure}[!t]
\centering
  \small
  \includegraphics[width=0.8\textwidth]{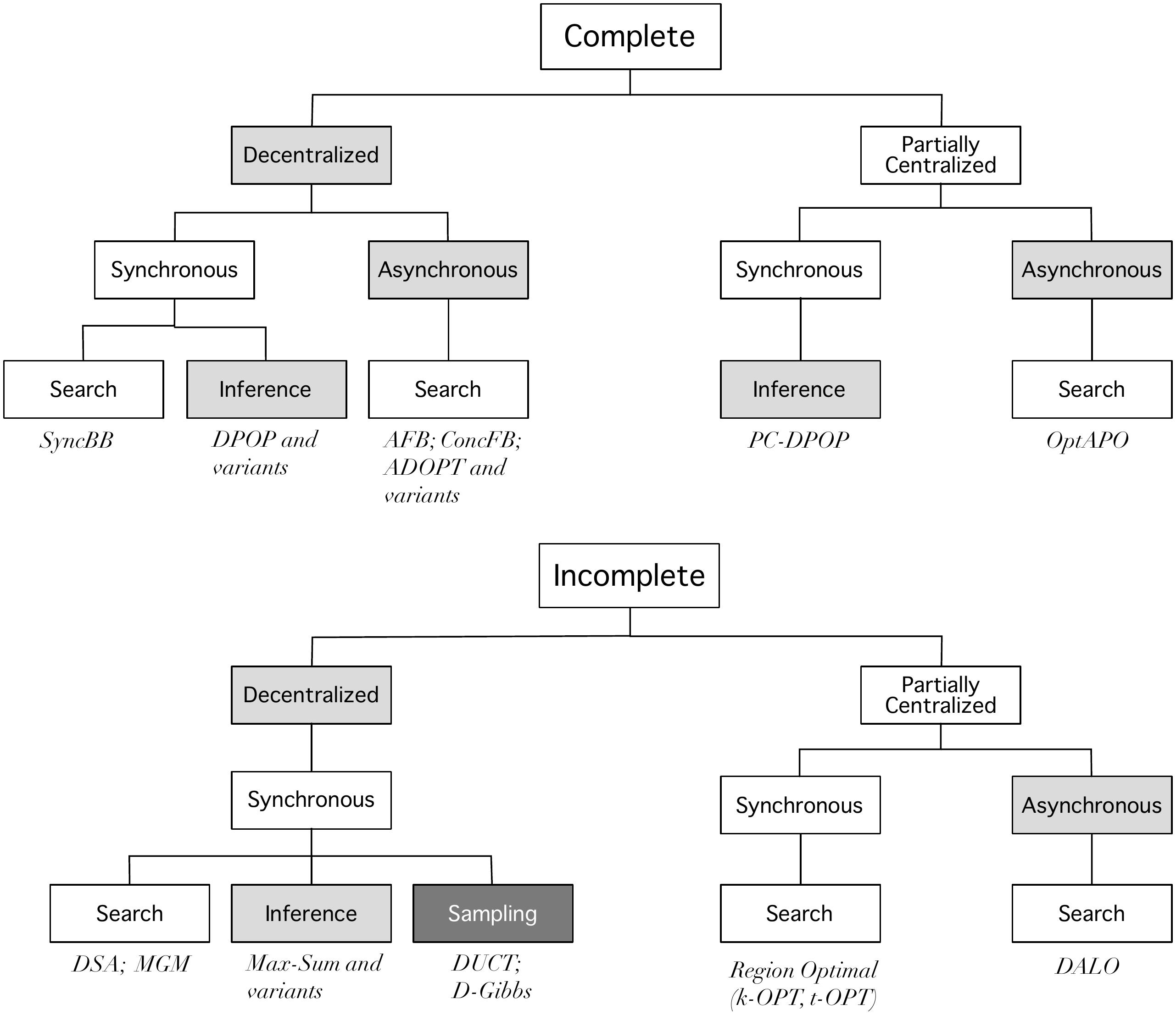}
 \caption{Classical DCOP Algorithm Taxonomy 
	\label{fig:dcop_taxonomy}}
\end{figure}

Figure~\ref{fig:dcop_taxonomy} illustrates a taxonomy of classical DCOP algorithms. The following subsections summarize some representative complete and incomplete algorithms from each of the classes introduced above. A detailed description of the DCOP algorithms is beyond the scope of this manuscript. The interested reader is referred to the original articles that introduce each algorithm.

\subsubsection{Complete Algorithms}
\label{sec:DCOP_complete}

Some of the algorithms described below were originally designed to solve the variant of DCOPs that maximizes rewards, while  others solve the variant that minimizes costs. However, the algorithms that maximize rewards can be easily adapted to minimize costs. For consistency, this survey describes the version of the algorithms that focus on  minimization of costs.  It also describes their quality, runtime, memory, and communication characteristics as summarized in Table~\ref{tab:algorithm_characteristics}.

\algref{SyncBB}{hirayama:97}
\emph{Synchronous Branch-and-Bound (\rev{SyncBB})} is a complete, synchronous, search-based algorithm that can be considered as a distributed version of a branch-and-bound algorithm. It uses a complete ordering of the agents to extend a \emph{Current Partial Assignment (CPA)} via a synchronous communication process. The CPA holds the assignments of all the variables controlled by all the visited agents, and, in addition, functions as a mechanism to propagate bound information. The algorithm prunes those parts of the search space whose solution quality is sub-optimal, by exploiting the bounds that are updated at each step of the algorithm. 

SyncBB agents perform $O(d^n)$ number of operations since the lowest priority agent needs to enumerate through all possible value combinations for all variables. While, by default, it is not an anytime algorithm, it can be easily extended to have an anytime property since it is a branch-and-bound algorithm. The memory requirement per SyncBB agent is $O(n)$ since the lowest priority agent stores the value assignment of all problem variables. In terms of communication requirement, SyncBB agents send $O(d^n)$ number of messages: The lowest priority agent enumerates through all possible value combinations for all variables and sends a message for each combination. The largest message, which contains the value assignment of all variables, is of size $O(n)$. 
Finally, the communication model of SyncBB depends on the given agent's complete ordering. Thus, agents may communicate with non-neighboring agents. 



\algref{AFB}{gershman:09}
\emph{Asynchronous Forward Bounding (AFB)} is a complete, asynchronous, search-based algorithm. 
It can be considered as an asynchronous version of SyncBB.  
In this algorithm, agents communicate their cost estimates, which in turn are used to compute bounds and prune the search space. 
In AFB, agents extend a CPA sequentially, provided that the \rev{lower bound on their costs does not exceed the global upper bound, that is, the cost of the best solution found so far}. 
Each agent performing an assignment (the ``assigning'' agent) triggers asynchronous checks of bounds, by sending \emph{forward} messages containing copies of the CPA to agents that have not yet assigned their variables. The unassigned agents that receive a CPA estimate the \rev{lower bound} of the CPA given their local view of the constraint graph and send their estimates back to the agent that originated the \emph{forward} message. This assigning agent will receive these estimates asynchronously and aggregate them into an updated \rev{lower bound. If the updated lower bound exceeds the current upper bound, the agent initiates a backtracking phase.}

The runtime, memory, and communication characteristics of AFB are identical to those of SyncBB for the same reasons. 
However, while both AFB and SyncBB agents communicate with non-neighboring agents, AFB agents broadcasts some of their messages while SyncBB agents do not.


\algref{ADOPT}{modi:05}
\emph{Asynchronous Distributed OPTimization (ADOPT)} is a complete, asynchronous, search-based algorithm. It can be considered as a distributed version of a memory-bounded best-first search algorithm. It makes use of a DFS pseudo-tree ordering of the agents. 
The algorithm relies on maintaining, in each agent, lower and upper bounds on the solution cost for the subtree rooted at its node in the DFS tree. Agents explore partial assignments in best-first order, that is, in \rev{increasing lower bound order}. They use \emph{COST} messages (propagated upwards in the DFS pseudo-tree) and \emph{THRESHOLD} and \emph{VALUE} messages (propagated downwards in the pseudo-tree) to iteratively tighten the lower and upper bounds, \rev{until the lower bound of the minimum cost  solution is equal to its upper bound}. ADOPT agents store \rev{lower bounds} as thresholds, which can be used to prune partial assignments that are provably sub-optimal. 

Similar to SyncBB and AFB, ADOPT agents perform $O(d^n)$ number of operations since the lowest priority agent needs to enumerate through all possible value combinations for all variables when the pseudo-tree degenerates into a pseudo-chain. It is also not an anytime algorithm as it is a best-first search algorithm. The memory requirement per ADOPT agent is $O(n+ld)$, where $O(n)$ is used to store a \emph{context}, which is the value assignment of all higher-priority variables, and $O(ld)$ is used to store the lower and upper bounds for each domain value and variable belonging to the agent's child agents. Finally, ADOPT agents communicate exclusively with their neighboring agents.

ADOPT has been extended in several ways. In particular, \emph{BnB-ADOPT} \cite{yeoh:10,gutierrez:12} uses a branch-and-bound method to reduce the amount of computation performed during search, and \emph{ADOPT(k)} combines both ADOPT and BnB-ADOPT into an integrated algorithm \cite{gutierrez:11}. There are also extensions that trade  solution optimality for smaller runtimes \cite{yeoh:09b}, extensions that use more memory for smaller runtimes \cite{yeoh:09a}, and extensions that maintain soft arc-consistency \cite{bessiere:12,bessiere:14,gutierrez:12b,gutierrez:13}.

\rev{
Finally, the \emph{No-Commitment Branch and Bound (NCBB)} algorithm \cite{chechetka:06} can be considered as a variant of ADOPT and SyncBB. Similar to ADOPT, NCBB agents exploit the structure defined by a pseudo-tree order to decompose the global objective function. This allow the agents to search non-intersecting parts of the search space concurrently. 
Another main feature of NCBB is the eager propagation of lower bounds on solution cost: An NCBB agent propagates its lower bound every time it learns about its ancestors' assignments. This feature provides an efficient pruning of the search space. 
The runtime, memory, and communication characteristics of NCBB are the same as those of ADOPT except that NCBB is an anytime algorithm.
}

\algref{ConcFB}{netzer:12}
\rev{
\emph{Concurrent Forward Bounding (ConcFB)} is a complete, asynchronous, search-based algorithm that  runs multiple parallel versions of AFB concurrently. By running multiple concurrent search procedures, it is able to quickly find a solution, apply a forward bounding process to detect regions of the search space to prune, and to dynamically create new search processes when detecting promising sub-spaces.
Similar to AFB, it uses a complete ordering of agents and variables instead of pseudo-trees. As such, it is able to simplify the management of reordering heuristics, which can provide substantial speed up to the search process \cite{zivan:06}. 

The algorithm operates as follows: Each agent maintains a global upper bound, which is updated during the search process.
The highest-priority agent begins the process by generating a number of different search processes (SP), one for each value of its variable. It then sends an \emph{LB\_Request} message to all unassigned agents. This \emph{LB\_Request} message contains the current CPA and triggers a calculation of the lower bounds of the receiving agents, which are sent back to the sender agent via a \emph{LB\_Report} message. 
If the sum of the aggregated costs and the current CPA cost is no smaller than the current upper bound, the agent selects another value for its variable and repeats the process. If the agent has exhausted all value assignments for its variable, then it backtracks, sending the CPA to the last assigning agent. If the CPA cost is lower than the current upper bound, then it forwards the \emph{CPA} message to the next non-assigned agent. 
Upon receiving a \emph{CPA} message, the agent repeats the above process. 
When the lowest-priority agent finds a solution resulting to a new upper bound, it broadcasts the upper bound via a \emph{UB} message, which is stored by each each agent.

\citeA{netzer:12} described a series of enhancements that can be used to speed up the search process of ConcFB, including dynamic variable ordering and dynamic splitting. Despite the process within a subproblem is carried out in a synchronous fashion, different subproblems are explored independently. Thus, the agents act asynchronously and concurrently. The runtime, memory, and communication characteristics of ConcFB are identical to those of AFB since it runs multiple parallel versions of AFB concurrently. 


}

\algref{DPOP}{petcu:05}
\emph{Distributed Pseudo-tree Optimization Procedure (DPOP)} is a complete, synchronous, inference-based algorithm that makes use of  a DFS pseudo-tree ordering of the agents. It involves three phases. In the first phase, the agents order themselves into a DFS pseudo-tree. In the second phase, called the UTIL propagation phase, each agent, starting from the leaves of the pseudo-tree, aggregates the \rev{costs} in its subtree for each value combination of variables in its separator.  The aggregated \rev{costs} are encoded in a \emph{UTIL} message, which is propagated from children to their parents, up to the root. In the third phase, called the VALUE propagation phase, each agent, starting from the root of the pseudo-tree, selects the optimal value for its variable. The optimal values are calculated based on the UTIL messages received from the agent's children and the \emph{VALUE} message received from its parent. The \emph{VALUE} messages contain the optimal values of the agents and are propagated from parents to their children, down to the leaves of the pseudo-tree. 

\rev{DPOP agents perform $O(d^{w^*})$ number of operations. When an agent optimizes for each value combination of variables in its separator, it takes $O(d^{w^*})$ operations since there are $w^*$ variables in the separator set in the worst case. It is not an anytime algorithm as it terminates upon finding its first solution, which is an optimal solution. The memory requirement per DPOP agent is $O(d^{w^*})$ since it stores all value combinations of variables in its separator. In terms of communication requirement, DPOP agents send $O(n)$ messages in total; $O(n)$ UTIL messages are propagated up the pseudo-tree and $O(n)$ VALUE messages are propagated down the pseudo-tree. The largest message sent by an agent, which contains the aggregated costs in its subtree for each value combination of variables in its separator, is $O(d^{w^*})$. Finally, DPOP agents only communicate with their neighboring agents only.
}

DPOP has also been extended in several ways to enhance its performance and capabilities. O-DPOP and MB-DPOP trade runtimes for smaller memory requirements \cite{petcu:06b,petcu:07}, A-DPOP trades solution optimality for smaller runtimes \cite{petcu:05c}, SS-DPOP trades runtime for increased privacy \cite{greenstadt:07}, PC-DPOP trades privacy for smaller runtimes \cite{petcu:07b}, H-DPOP propagates hard constraints for smaller runtimes \cite{kumar:08}, BrC-DPOP enforces branch consistency for smaller runtimes \cite{fioretto:14b}, and ASP-DPOP is a declarative version of DPOP that uses Answer Set Programming \cite{le:15}.

\algref{OptAPO}{mailler:04}
\emph{Optimal Asynchronous Partial Overlay (OptAPO)} \rev{is a complete, asynchronous, search-based algorithm}. It trades  agent privacy for smaller runtimes through partial centralization. It employs  a cooperative mediation schema, where agents can act as mediators and propose value assignments to other agents. In particular, the agents check if there is a conflicting assignment with some neighboring agent. If a conflict is found, the agent with the highest priority acts as a mediator.  During mediation, OptAPO solves subproblems using a centralized branch-and-bound-based search, and when solutions of overlapping subproblems still have conflicting assignments, the solving agents increase the degree of centralization to resolve them. By sharing their knowledge with centralized entities, agents can improve their local decisions, reducing the communication costs. For instance, the algorithm has been shown to be superior to ADOPT on simple combinatorial problems \cite{mailler:04}. However, it is possible that several mediators solve overlapping problems, duplicating efforts \cite{petcu:07b}, which can be a bottleneck in dense problems. 

\rev{OptAPO agents perform $O(d^{n})$ number of operations, in the worst case, as a mediator agent may solve the entire problem. Like ADOPT and DPOP, OptAPO is not an anytime algorithm. The memory requirement per OptAPO agent is $O(nd)$ since it needs to store all value combinations of variables in its mediation group, which is of size $O(n)$. 
In terms of communication requirement, OptAPO agents send $O(d^n)$ messages in the worst case, though the number of messages decreases with increasing partial centralization. The size of the messages is bounded by $O(d + n)$, where in the initialization phase of each mediation step, each agent sends its domain to its neighbors and the list of variables that it seeks to mediate. Finally, OptAPO agents can communicate with non-neighboring agents during the mediation phase.}

\rev{The original version of OptAPO has been shown to be incomplete due to the asynchronicity of the different mediators' groups, which can lead to race conditions. \citeA{grinshpoun:08} proposed a complete variant that remedies this issue.}
\subsubsection{Incomplete Algorithms}
\label{sec:DCOP_incomplete}


\algref{Max-Sum}{farinelli:08}
\rev{\emph{Max-Sum} is an incomplete, synchronous, inference-based algorithm based on belief propagation}. It operates on factor graphs by performing a marginalization process of the \rev{cost functions}, and optimizing the costs for each given variable. This process is performed by recursively propagating messages between variable nodes and factor nodes. The value assignments take into account their impact on the marginalized \rev{cost function}. Max-Sum is guaranteed to converge to an optimal solution in acyclic graphs, but convergence is not guaranteed on  cyclic graphs. 

Max-Sum agents perform $O(d^{l})$ number of operations in each iteration, where each agent needs to optimize for all value combinations of neighboring variables. It is not an anytime algorithm. The memory requirement per Max-Sum agent is $O(d^l)$ since it needs to store all value combinations of neighboring variables. In terms of communication requirement, in the worst case, each Max-Sum agent sends $O(l)$ messages in each iteration, one to each of its neighbor. Thus, the total number of messages sent across all agents is $O(\ell nl)$. Each message is of size $O(d)$ as it needs to contain the current aggregated costs of all the agent's variable's values. Finally, the agents communicate exclusively with their neighboring agents.

Max-Sum has been extended in several ways. 
Bounded Max-Sum bounds the quality of the solutions found by removing a subset of edges from a cyclic DCOP graph to make it acyclic, and running Max-Sum to solve the acyclic problem \cite{rogers:11}; 
Improved Bounded Max-Sum improves on the error bounds \cite{rollon:12}; and 
\rev{Max-Sum\_ADVP} guarantees convergence in acyclic graphs through a two-phase value propagation phase \cite{zivan:12,chen:17}. 
Max-Sum and its extensions have been successfully used to solve a number of large scale, complex MAS applications (see Section~\ref{sec:applications}).

\algref{Region Optimal}{pearce:07}
{\emph{Region-optimal} algorithms are incomplete, synchronous, search-based algorithms that allow users to specify regions of the constraint graph and solve the subproblem within each region optimally. Regions may be defined to have a maximum size of $k$ agents \cite{pearce:07}, $t$ hops from each agent \cite{kiekintveld:10}, or a combination of both size and hops \cite{vinyals:11}. The concept of $k$-optimality is defined with respect to the number of agents whose assignments conflict, whose set is denoted by $\setf{c}(\sigma, \sigma')$, for two assignments $\sigma$ and $\sigma'$. The deviating cost of $\sigma$ with respect to $\sigma'$, denoted by $\Delta(\sigma, \sigma')$, is defined as the difference of the aggregated \rev{cost} associated to the assignment $\sigma$ ($F(\sigma)$) minus the \rev{cost} associated to $\sigma'$ ($F(\sigma')$). An  assignment $\sigma$ is $k$-optimal if $\forall \sigma' \in \Sigma$, such that $|\setf{c}(\sigma, \sigma')| \leq k$, we have that $\Delta(\sigma, \sigma') \geq 0$.
In contrast, the concept of $t$-distance emphasizes the number of hops from a central agent $a$ of the region $\Omega_t(a)$, that is the set of agents which are separated from $a$ by at most $t$ hops. An assignment $\sigma$ is $t$-distance optimal if, $\forall \sigma' \in \Sigma$, $F(\sigma) \geq F(\sigma')$ with $\setf{c}(\sigma, \sigma') \subseteq \Omega_t(a)$, for any $a \in \setf{A}$. \rev{Therefore, the solutions found have theoretical error bounds that are a function of $k$ and/or $t$.}
Region-optimal algorithms adopt a partially-centralized resolution scheme in which the subproblem within each region is solved optimally by a centralized authority \cite{tassa:16}. However, this scheme can be altered to use a distributed algorithm to solve each subproblem.

\rev{Region-optimal agents perform $O(d^{w^*})$ number of operations in each iteration, as each agent runs DPOP to solve the problem within each region optimally. It is also an anytime algorithm as solutions of improving quality are found until they are region-optimal. The memory requirement per region-optimal agent is $O(d^{w^*})$ since its region may have an induced width of $w^*$ and it uses DPOP to solve the problem within its region. In terms of communication requirement, each region-optimal agent sends $O(n)$ messages, one to each agent within its region. Thus, the total number of messages sent across all agents is $O(\ell n^2)$. Each message is of size $O(d^{w^*})$ as it uses DPOP. 
Finally, the agents communicate to all agents within their region -- either to a distance of $\lfloor\frac{k}{2}\rfloor$ or $t$ hops away. Thus, they may communicate with non-neighboring agents.
}

\rev{An asynchronous version of regional-optimal algorithms, called \emph{Distributed Asynchronous Local Optimization} (DALO), was proposed by \citeA{kiekintveld:10}. The DALO simulator provides a mechanism to coordinate the decision of local groups of agents based on the concepts of $k$-optimality and $t$-distance.} 

\algref{MGM}{maheswaran:04b}
\emph{Maximum Gain Message (MGM)} is an incomplete, synchronous, search-based algorithm that performs a distributed local search. Each agent starts by assigning a random value to each of its variables. Then, it sends this information to all its neighbors. Upon receiving the values of its neighbors, it calculates the maximum gain \rev{(i.e.,~the maximum decrease in cost)} if it changes its value and sends this information to all its neighbors. Upon receiving the gains of its neighbors, the agent changes its value if its gain is the largest among those of its neighbors. This process repeats until a termination condition is met. MGM provides no quality guarantees on the returned solution. 

\rev{MGM agents perform $O(ld)$ number of operations in each iteration, as each agent needs to compute the cost for each of its values by taking into account the values of all its neighbors. MGM is anytime  since agents only change their values when they have a non-negative gain. The memory requirement per MGM agent is $O(l)$. Each agent needs to store the values of all its neighboring agents. In terms of communication requirement, each MGM agent sends $O(l)$ messages, one to each of its neighboring agents. Thus, the total number of messages sent across all agents is $O(\ell nl)$. Each message is of constant size $O(1)$ as it contains either the agent's current value or the agent's current gain. Finally, the agents communicate exclusively with their neighboring agents.
}

\algref{DSA}{zhang:05}
\emph{Distributed Stochastic Algorithm (DSA)} is an incomplete, synchronous, search-based algorithm that is similar to MGM, except that each agent does not send its gains to its neighbors and it does not change its value to the value with the maximum gain. Instead, it decides stochastically if it takes on the value with the maximum gain or other values with smaller gains. This stochasticity allows DSA to escape from local minima. Similar to MGM, it repeats the process until a termination condition is met, and it cannot provide quality guarantees on the returned solution. 
\rev{The runtime, memory, and communication characteristics of DSA are identical to those of MGM since it is essentially a stochastic variant of MGM.}

\algref{DUCT}{ottens:17}
The \emph{Distributed Upper Confidence Tree (DUCT)} algorithm is an incomplete, synchronous, sampling-based  algorithm that is inspired by Monte-Carlo Tree Search and employs confidence bounds to solve DCOPs.
DUCT emulates a search process analogous to that of ADOPT, where agents select the values to assign to their variables according to the information encoded in their \emph{context} messages (i.e.,~the assignments to all the variables in the receiving variable's separator). 
However, rather than systematically selecting the next value to assign to their own variables, DUCT agents sample such values. To focus on promising assignments, DUCT constructs a confidence bound $B$, such that cost associated to the best value for any context is at least $B$, and hence agents sample the choice with the lowest bound. 
This process is started by the root agent of the pseudo-tree: After sampling a value for its variable, it communicates its assignment to its children in a \emph{context} message. When an agent receives this message, it repeats this process  until the leaf agents are reached. 
When the leaf agents choose a value assignment, they calculate the cost within their context and propagate this information up to the tree in a \emph{cost} message. 
This process continues for a given number of iterations or until convergence is achieved, i.e.,~until the sampled values in two successive iterations do not change. Therefore, DUCT is able to provide quality guarantees on the returned solution. 

\rev{DUCT agents perform $O(ld)$ number of operations in each iteration, as each agent needs to compute the cost for each of its values by taking into account the values of all its neighbors. It is an anytime algorithm; The quality guarantee improves with increasing number of iterations. The memory requirement per DUCT agent is $O(d^{n})$\footnote{It is actually $O(d^t)$, where $t$ is the depth of the pseudo-tree. However, in the worst case, when the pseudo-tree degenerates into a pseudo-chain, then $t = n$.} since it needs to store the best cost for all possible contexts. In terms of communication requirement, in each iteration, each DUCT agent sends one message to its parent in the pseudo-tree and one message to each of its children in the pseudo-tree. Thus, the total number of messages sent across all agents is $O(\ell n)$. Each message is of size $O(n)$; \emph{context} messages contain the value assignment for all higher priority agents. Finally, the agents communicate exclusively with their neighboring agents.}

\algref{D-Gibbs}{nguyen:13}
The \emph{Distributed Gibbs (D-Gibbs)} algorithm is an incomplete, synchronous, sampling-based algorithm that extends the Gibbs sampling process \cite{geman:84} by tailoring it to solve DCOPs in a decentralized manner. The Gibbs sampling process is a centralized Markov Chain Monte-Carlo algorithm that can be used to approximate joint probability distributions. By mapping DCOPs to maximum a-posteriori estimation problems, probabilistic inference algorithms like Gibbs sampling can be used to solve DCOPs. 

Like DUCT, it too operates on a pseudo-tree, and the agents sample sequentially from the root of the pseudo-tree down to the leaves. Like DUCT, each agent also stores a context (i.e.,~the current assignment to all the variables in its separator) and it samples based on this information. Specifically, it computes the probability for each of its values given its context and chooses its current value based on this probability distribution. After it chooses its value, it informs its lower priority neighbors of its value, and its children agents start to sample. This process continues until all the leaf agents sample. Cost information is propagated up the pseudo-tree. This process continues for a fixed number of iterations or until convergence. Like DUCT, D-Gibbs is also able to provide quality guarantees on the returned solution. 

\rev{The runtime characteristics of D-Gibbs are identical to that of DUCT and for the same reasons. However, its memory requirements are smaller: The memory requirement per D-Gibbs agent is $O(l)$ since it needs to store the current values of all its neighbors. In terms of communication requirement, in each iteration, each D-Gibbs agent sends $O(l)$ messages, one to each of its neighbors. Thus, the total number of messages sent across all agents is $O(\ell nl)$. Each message is of constant size $O(1)$ since they contain only the current value of the agent or partial cost of its solution. Finally, the agents communicate exclusively with their neighboring agents.
}

A version of the algorithm that speeds up the agents' sampling process with Graphical Processing Units (GPUs) is
described in   \cite{fioretto:CP-16}.

\subsection{Tradeoffs Between the Various DCOP Algorithms} 
\label{sec:tradeoffs}
\rev{
The various DCOP algorithms discussed above provide a good coverage across various characteristics that may be important in different applications. As such, how well suited an algorithm is for an application depends on how well the algorithm's characteristics match up to the application's characteristics. The next section discusses several suggestions for the types of algorithms that are recommended based on the characteristics of the application at hand.

\subsubsection{Complete Algorithms}

When optimality is a requirement of the application, then one is limited to complete algorithms:
\bitemize
\item If the agents in the application have large amounts of memory and it is faster to send few large messages than many small messages, then inference-based algorithms (e.g., DPOP and its extensions) are  preferred over search-based algorithms (e.g.,~SyncBB, AFB, ADOPT, ConcFB, OptAPO). 
This is because, in general, search algorithms perform some amount of redundant communication. Thus, for a given problem instance, the overall runtime of inference-based algorithms tend to be smaller than the runtime of search-based ones. 

\item If the agents in the application have limited amounts of memory, then one has to use the search-based algorithms (e.g.,~SyncBB, AFB, ADOPT, ConcFB, OptAPO), which have small memory requirements. The exception is when the problem has a small induced width (e.g.,~the constraint graph is acyclic), in which case inference-based algorithms (e.g., DPOP) are also preferred.
    \bitemize
    \item If partial centralization is allowed by the application, then OptAPO is preferred as it has been shown to outperform many of the other search algorithms \cite{mailler:04}. 
    \item Otherwise, ConcFB is recommended as it has been shown to outperform AFB due to the concurrent search \cite{netzer:12}, and AFB has been shown to outperform ADOPT and SyncBB \cite{gershman:09}. The exception is if the application does not permit agents to communicate directly to non-neighbors, in which case \text{ConcFB}, AFB, and SyncBB cannot be used and one is restricted to use ADOPT or one of its variants. Note that many of the variants (e.g.,~BnB-ADOPT, NCBB) have been shown to significantly outperform ADOPT while maintaining the same runtime, memory, and communication requirements \cite{chechetka:06,yeoh:10}. 
    \eitemize
\eitemize

\subsubsection{Incomplete Algorithms}

In terms of incomplete algorithms, the following recommendations are given:
\bitemize
\item If the solution returned must have an accompanying quality guarantee, then, one can choose to use Bounded Max-Sum, region-optimal algorithms, DUCT, or D-Gibbs. Bounded Max-Sum allows users to choose the error bound as a function of the different subsets of edges that can be removed from the graph to make it acyclic. Region-optimal algorithms allow users to parameterize the error bound according to the size of the region $k$ or the number of hops $t$ that the solution should be optimal for. Finally, DUCT and D-Gibbs allow users to parameterize the error bound based on the number of sampling iterations to conduct. The error bounds for these two algorithms are also probabilistic bounds (i.e.,~the likelihood that the quality of the solution is within an error bound is a function of the number of iterations). Therefore, the choice of algorithm will depend on the type of error bound one would like to impose on the solutions. One may also choose to use a number of extensions of complete algorithms (e.g.,~Weighted (BnB-)ADOPT and A-DPOP) that allow users to parameterize the error bound and affect the degree of speedup.

\item If the solution quality guarantee is not required, then one can also use Max-Sum, MGM, or DSA. Their performance depends on a number of factors: If the problem has large domain sizes, MGM and DSA often outperform Max-Sum, since the memory and computational complexities of Max-Sum grows exponentially with the domain size. However, if the problem has small induced widths (for instance, when its constraint graph is acyclic), then Max-Sum is very efficient. It is even guaranteed to find optimal solutions when the induced width is 1. In general, Max-Sum tends to find solutions of good quality especially when considering its recent improvements (e.g., \cite{zivan:17}).

\item If the problem has hard constraints (i.e.,~certain value combinations are prohibited), then the sampling algorithms (i.e.,~DUCT and D-Gibbs) are not recommended as they are not able to handle such problems. They require the cost functions to be smooth, and exploit that characteristic to explore the  search space. Thus, one is restricted to search- or inference-based algorithms.

\item In general, MGM and DSA are good robust benchmarks as they tend to find reasonably high quality solutions in practice. However, if specific problem characteristics are known, such as the ones discussed above, then certain algorithms may be able to exploit them to find better solutions. 

\eitemize
}
 
\subsection{Notable Variant: Asymmetric DCOPs}
\label{sec:asymmetric_dcop}

\emph{Asymmetric DCOPs} \cite{grinshpoun:13} are used to model multi-agent problems where agents controlling variables in the scope of a cost function \rev{can incur to different costs}, given a fixed join assignment. Such a problem cannot be naturally represented by classical DCOPs, which require that all agents controlling variables participating in a cost function \rev{incur to the same cost as each other}.

\subsubsection{Definition}
An \emph{Asymmetric DCOP} is a tuple $\langle \setf{A, X, D, F}, \alpha \rangle$, where $\setf{A, X, D}$, and $\alpha$ are as defined in Definition~\ref{def:dcop}, and each cost function $f_i \in \setf{F}$ is defined as: 
$f_i : \bigtimes_{x_j \in \scope{i}} D_j \times \alpha(f_i) \to \left  (\mathbb{R}^{+} \cup \{\bot\} \right) $.
In other words, an Asymmetric DCOP is a DCOP where the \rev{cost that an agent incurs from a cost function may differ from the cost that another agent incurs from the same cost function}. 

As \rev{costs} for participating agents may differ from each other, the goal in Asymmetric DCOPs is  different from the goal in classical DCOPs. Given a \rev{cost} function $f_j \in \setf{F}$ and complete assignment $\sigma$, let $f_j(\sigma, a_i)$ denote the \rev{cost incurred} by agent $a_i$ from \rev{cost} function $f_j$ with complete assignment $\sigma$. Then, the goal in Asymmetric DCOPs is to find the solution $\sigma^*$:
\begin{equation}
	\label{eq:adcop_goal}
	\displaystyle \sigma^* \coloneqq \argmin_{\sigma \in \Sigma} 
	\sum_{f_j \in \setf{F}} \sum_{a_i \in \alpha(f_j)} f_j(\sigma_{\scope{j}}, a_i )
\end{equation}

As in classical DCOPs, solving Asymmetric DCOPs is NP-hard. 
In particular, it is possible to reduce any Asymmetric DCOP to an equivalent classical DCOP by introducing a polynomial 
number of variables and constraints, as described in the next section.

\subsubsection{Relation to Classical DCOPs}
One  way to solve MAS problems with asymmetric \rev{costs} via classical DCOPs is through the \emph{Private Event As Variables (PEAV)} model \cite{maheswaran:04b}. It can capture \rev{asymmetric costs} by introducing, for each agent, as many ``mirror'' variables as the number of variables held by neighboring agents. The consistency with the neighbors' state variables is imposed by a set of equality constraints. However, this formalism suffers from scalability problems, as it may result in a significant increase in the number of variables in a DCOP. In addition, \citeA{grinshpoun:13} showed that most of the existing incomplete classical DCOP algorithms cannot be used to effectively solve Asymmetric DCOPs, even when the problems are reformulated through the PEAV model. 
They show that such algorithms are unable to distinguish between different solutions that satisfy all hard constraints, resulting in a convergence to one of those  solutions and the inability to escape that local optimum. Therefore, it is important to design \rev{specialized} algorithms to solve Asymmetric DCOPs.

\subsubsection{Algorithms}
The current research direction in the design of Asymmetric DCOP algorithms has focused on adapting existing classical DCOP algorithms to handle the \rev{asymmetric costs}. Asymmetric DCOPs require that each agent, whose variables participate in a \rev{cost} function, coordinate the aggregation of their individual \rev{costs}. To do so, two approaches have been identified \cite{brito:09}:

\bitemize 
\item A \emph{two-phase strategy}, where only one side of the constraint (i.e.,~the \rev{cost induced by} one agent) is considered in the first phase. The other side(s) (i.e.,~the \rev{cost induced by} the other agent(s)) is considered in the second phase once a complete assignment is produced. As a result, the \rev{costs} of all agents are aggregated. %
\item A \emph{single-phase strategy}, which requires a systematic check of each side of the constraint before reaching a complete assignment. Checking each side of the constraint is often referred to as \emph{back checking}, a process that can be performed either synchronously or asynchronously. %
\eitemize

\bigskip\noindent \textsc{Complete Algorithms}

\algref{SyncABB-2ph}{grinshpoun:13}
\emph{Synchronous Asymmetric Branch and Bound - 2-phase (SyncABB-2ph)} is a complete, \rev{synchronous}, search-based algorithm that extends SyncBB with the two-phase strategy. Phase~1 emulates SyncBB, where each agent considers the values of its cost functions with higher-priority agents.
Phase~2 starts once a complete assignment is found. During this phase, each agent aggregates the sides of the cost functions that were not considered during Phase~1 and verifies that the known bound is not exceeded. If the bound is exceeded, Phase~2 ends and the agents restart Phase~1 by backtracking and resuming the search from the lower priority agent that exceeded the bound. The worst case runtime, memory, and communication requirements of this algorithm are the same as those of SyncBB.

\algref{SyncABB-1ph}{grinshpoun:13,levit:13}
\emph{Synchronous Asymmetric Branch and Bound - 1-phase (SyncABB-1ph)} is a complete, \rev{synchronous}, search-based algorithm that extends SyncBB with the one-phase strategy. Each agent, after having extended the CPA, updates the bound with its local \rev{cost} associated to the cost functions involving its variables -- as done in SyncBB. In addition, the CPA is sent back to the assigned agents to update its bound via a sequence of back checking operations. The worst case runtime, memory, and communication requirements of this algorithm are the same as those of SyncBB.

\algref{ATWB}{grinshpoun:13}
The \emph{Asymmetric Two-Way Bounding (ATWB)} algorithm is a complete, asynchronous, search-based algorithm that extends AFB to accommodate both forward bounding and backward bounding. The forward bounding is performed analogously to AFB. The backward bounding, instead, is achieved by sending copies of the CPA backward to the agents whose assignments are included in the CPA. Similar to what is done in AFB, agents that receive a copy of the CPA compute their estimates and send them forward to the assigning agent. The worst case runtime, memory, and communication requirements of this algorithm are the same as those of AFB.

\bigskip\noindent \textsc{Incomplete Algorithms}

\algref{ACLS}{grinshpoun:13}
 \emph{Asymmetric Coordinated Local Search (ACLS)} is an incomplete, synchronous, search-based algorithm that extends DSA. After a random value initialization, each agent exchanges its values with all its neighboring agents. At the end of this step, each agent identifies all possible improving assignments for its own variables, given the current neighbors choices. Each  agent then selects one such assignments, according to the distribution of gains (i.e.,~reductions in costs) from each proposal assignment, and exchanges it with its neighbors. When an agent receives a proposal assignment, it responds with the evaluation of its side of the cost functions, resulting from its current assignment and the proposal assignments of the other agents participating in the cost function. After receiving the evaluations from each of its neighbors, each  agent estimates the potential gain or loss derived from its assignment, and commits to a change with a given probability, similar to agents in DSA, to escape from local minima. The worst case runtime, memory, and communication requirements of this algorithm are the same as those of DSA.

\algref{MCS-MGM}{grinshpoun:13}
\emph{Minimal Constraint Sharing MGM (MCS-MGM)} is an incomplete, synchronous, search-based algorithm that extends MGM by considering each side of the cost function. Like MGM, the agents operate in an iterative fashion, where they exchange their current values at the start of each iteration. Afterwards, each agent sends the \rev{cost} for its side of each cost function to its neighboring agents that participate in the same 
cost function.\footnote{This is a version of the algorithm with a guarantee that it will converge to a local optima. In the original version of the algorithm, which does not have such guarantee, each agent sends the \rev{cost} only if its gain with the neighbor's new values is larger than the neighbor's last known gain.} Upon receiving this information, each agent  knows the total \rev{cost} for each cost function -- by adding together the value of both sides of the cost function. Therefore, like in MGM, the agent can calculate the maximum gain (i.e.,~maximum reduction in \rev{costs}) if it changes its values, and will send this information to all its neighbors. Upon receiving the gains of its neighbors, each agent changes its value if its gain is the largest among its neighbors. The worst case runtime, memory, and communication requirements of this algorithm are the same as those of MGM.

\subsection{Notable Variant: Multi-Objective DCOPs} 

\emph{Multi-Objective Optimization (MOO)} \cite{miettinen:99,marler:04} aims at solving problems involving more than one objective function to be optimized simultaneously. In a MOO problem, optimal decisions need to accommodate potentially conflicting objectives. 
\emph{Multi-Objective DCOPs} extend MOO problems and DCOPs \cite{dellefave:11}.

\subsubsection{Definition}
%
A \emph{Multi-Objective DCOP (MO-DCOP)} is  a tuple $\langle \setf{A},  \setf{X}, \setf{D}, \vec{\setf{F}}, \alpha \rangle$, where $\setf{A, X, D}$, and $\alpha$ are as defined in Definition~\ref{def:dcop}, and $\vec{\setf{F}} = [F_1, \ldots, F_{h}]^T$ is a vector of multi-objective functions, where each $F_i$ is a set of cost functions $f_j$ as defined in Definition~\ref{def:dcop}.
For a complete assignment $\sigma$ of a MO-DCOP, let the \rev{cost} for $\sigma$ according to the \nth{i} multi-objective optimization function set $F_i$, where $1 \leq i \leq h$, be
\begin{equation}
	\label{eq:modcop_reward}
	F_i( \sigma ) \coloneqq \sum_{f_j \in F_i} f_j(\sigma_{\scope{j}})
\end{equation}
The goal of a MO-DCOP is to find a complete assignment $\sigma^*$ such that:
\begin{equation}
	\label{eq:modcop_goal}
	\sigma^* \coloneqq \argmin_{\sigma \in \Sigma} \vec{\setf{F}}(\sigma) = 
    \argmin_{\sigma \in \Sigma} \:[F_1(\sigma), \ldots, F_{h}(\sigma)]^T
\end{equation}
where $\vec{\setf{F}}(\sigma)$ is a \emph{\rev{cost vector}} for the MO-DCOP. A solution to a MO-DCOP involves the optimization of a set of partially-ordered assignments. The above definition considers point-wise comparison of vectors---\rev{i.e.,~
$\vec{\setf{F}}(\sigma) \leq \vec{\setf{F}}(\sigma')$ if $F_i(\sigma) \leq F_i(\sigma')$ for all $1\leq i \leq h$}.
Typically, there is no single global solution where all the objectives are optimized at the same time. Thus, solutions of a MO-DCOP are characterized by the concept of \emph{Pareto optimality}, which can be defined through the concept of \emph{dominance}:

\begin{definition}[Dominance]%
    A solution $\sigma \in \Sigma$ is \emph{dominated} by a solution $\sigma' \in \Sigma$ \emph{iff} 
    \rev{$\vec{\setf{F}}(\sigma') \leq \vec{\setf{F}}(\sigma)$ and $F_i(\sigma') < F_i (\sigma)$ for at least one $F_i$}. 
\end{definition}%

\begin{definition}[Pareto Optimality]%
  \label{def:pareto_optimal}
    A solution $\sigma' \in \Sigma$ is \emph{Pareto optimal} \emph{iff} it is not dominated by any other solution. 
\end{definition}%
\noindent Therefore, a solution is Pareto optimal \emph{iff} there is no other solution that improves at least one objective function without deteriorating the \rev{cost} of another function. Another important concept is the \emph{Pareto front}:

\begin{definition}[Pareto Front]%
The \emph{Pareto front} is the set of all \rev{cost vectors} of all Pareto optimal solutions.
\end{definition}%
\noindent Solving a MO-DCOP is equivalent to finding the Pareto front. However, even for tree-structured MO-DCOPs, the size of the Pareto front may be exponential in the number of variables.\footnote{In the worst case, every possible solution is a Pareto optimal solution.} Thus, multi-objective algorithms often provide solutions that may not be Pareto optimal but may satisfy other criteria that are significant for practical applications. A widely-adopted criterion is that of \emph{weak Pareto optimality}:
\begin{definition}[Weak Pareto Optimality]
  \label{def:weakly_pareto_optimal}
    A solution $\sigma' \in \Sigma$ is \emph{weakly Pareto optimal} \emph{iff} there is no other solution $\sigma \in \Sigma$ such that \rev{$\vec{\setf{F}}(\sigma) < \vec{\setf{F}}(\sigma')$}.
\end{definition}
\noindent In other words, a solution is weakly Pareto optimal if there is no other solution that improves \emph{all} of the objective functions simultaneously. An alternative approach to Pareto optimality is one that uses the concept of \emph{utopia points}:

\begin{definition}[Utopia Point]%
  \label{def:utopia_point}
A cost vector $\vec{\setf{F}}^\circ = [F^\circ_1, \ldots, F^\circ_h]^T$ is a \emph{utopia point} \emph{iff} $F^\circ_i = \min_{\sigma \in \Sigma} F_i(\sigma)$ for all $1 \leq i \leq h$.
\end{definition}%
\noindent In other words, a utopia point  is the vector of \rev{costs} obtained by independently optimizing $h$ DCOPs, each associated to one objective of the multi-objective function vector. In general, $\vec{\setf{F}}^\circ$ is unattainable. Therefore, different approaches focus on finding a \emph{compromise solution} \cite{salukvad:71}, which is a Pareto optimal solution that is \emph{close} to the utopia point. The concept of \emph{closeness} is dependent on the approach adopted.

Similar to their centralized counterpart, MO-DCOPs have been shown to be NP-hard (their decision versions), and \#P-hard (the related counting versions), and to have exponentially many non-dominated points \cite{glasser:10}.

\subsubsection{Algorithms}
This section categorizes the proposed MO-DCOP algorithms into two classes: \emph{complete} and \emph{incomplete} algorithms, according to their ability to find the complete set of Pareto optimal solutions or only a subset of it.

\medskip\noindent\textsc{Complete Algorithms}

\algref{MO-SBB}{medi:14}
\emph{Multi-Objective Synchronous Branch and Bound (MO-SBB)} is a complete, synchronous, search-based algorithm that extends SyncBB. It uses an analogous search strategy to that of the mono-objective SyncBB: After establishing a complete ordering, MO-SBB agents extend a CPA with their own value assignments and the current associated \rev{cost vectors}. Once a non-dominated solution is found, it is broadcasted to all agents, which add the solution to a list of global bounds. Thus, agents maintains an approximation of the Pareto front, which is used to bound the exploration, and extend the CPA only if the new partial assignment is not dominated by solutions in the list of global bounds. When the algorithm terminates, it returns the set of Pareto optimal solutions obtained by filtering the list of global bounds by dominance. The worst case runtime and communication requirements of this algorithm are the same as those of SyncBB. In terms of memory requirement, each MO-SBB agent needs $O(np)$ amount of memory, where $p$ is the size of the Pareto set.

\algref{Pseudo-tree Based Algorithm}{matsui:12a}
The proposed algorithm is a complete, asynchronous, search-based algorithm that extends ADOPT. It introduces the notion of boundaries on the vectors of multi-objective values, which extends the concept of lower and upper bounds to vectors of values. The proposed approach starts with 
the assumption that $|F_1| = \dots = |F_h| = k$. Furthermore, the cost functions within each $F_i$ are sorted according to a predefined ordering, and for each $1\leq j \leq k$, the scope of $f^i_j$ (i.e.,~the $j^{th}$ function in $F_i$) is the same for each $i$ (i.e.,~all functions in the same position in different $F_i$ have the same scope). Thus, without loss of generality, the notation $\scope{j}$ will be used to refer to the scope of $f^i_j$.

Given a complete assignment $\sigma$, for $1\leq j\leq k$, 
let $\vec{\sigma}_{j} = [\sigma_{f_j}^1, \ldots, \sigma_{f_j}^h] = [f_j^1(\sigma_{\scope{j}}), \ldots, f_j^h(\sigma_{\scope{j}})]$ be the vector of cost values. 
The notion of non-dominance is applied to these vectors, \rev{where a vector $\vec{\sigma}_{j}$ is non-dominated \emph{iff} there is no other vector $\vec{\sigma}_{j}'$ such that $ub(\sigma_{j}'\!^{s}) \leq lb(\sigma_{j}^{s})$ for all $1 \leq s \leq h$ and 
$ub(\sigma_{j}'\!^{s}) < lb(\sigma_{j}^{s})$ for at least one $s$.}
The algorithm uses the notion of non-dominance for bounded vectors to retain exclusively non-dominated vectors. 

\rev{
The worst case runtime and communication requirements of this algorithm are the same as those of ADOPT. In terms of memory requirement, each agent needs $O(np)$ amount of memory. However, notice that the number of combinations of cost vectors grows exponentially with the number of tuples of cost values, in the worst case. }
This algorithm has also been extended to solve Asymmetric MO-DCOPs \cite{matsui:14}, which is an extension of both Asymmetric DCOPs and MO-DCOPs.

\medskip\noindent\textsc{Incomplete Algorithms}

\algref{B-MOMS}{dellefave:11}
\emph{Bounded Multi-Objective Max-Sum (B-MOMS)} is an incomplete, asynchronous, inference-based algorithm, and was the first MO-DCOP algorithm introduced. It extends Bounded Max-Sum to compute bound approximations for MO-DCOPs. It consists of three phases. The \emph{Bounding Phase} generates an acyclic subgraph of the multi-objective factor graph, using a generalization of the maximum spanning tree problem to vector weights. During the \emph{Max-sum Phase}, the agents coordinate to find the Pareto optimal set of solutions to the acyclic factor graph generated in the bounding phase. This is achieved by extending the addition and marginal maximization operators adopted in Max-Sum to the case of  multiple objectives. Finally, the \emph{Value Propagation Phase} allows agents to select a consistent variable assignment, as there may multiple Pareto optimal solutions. The bounds provided by the algorithm are computed using the notion of utopia points. 

\rev{The worst case runtime requirement of this algorithm is the same as those of Max-Sum. In terms of communication requirement, the number of messages sent is also like Max-Sum, but the size of each message is now $O(pd^n)$. In terms of memory requirement, each B-MOMS agent needs $O(pd^n)$ amount of memory to store and process the messages received.
}

\algref{DP-AOF}{okimoto:13a}
\emph{Dynamic Programming based on Aggregate Objective Functions (DP-AOF)} is an incomplete, synchronous, inference-based algorithm. It adapts the AOF technique \cite{miettinen:99}, designed to solve centralized multi-objective optimization problems, to solve MO-DCOPs. 
Centralized AOF adopts a scalarization to convert a MOO problem into a single objective optimization. This is done by assigning weights $(\alpha_1, \ldots, \alpha_h)$ to each of the cost functions in the objective vector $[F_1 ,\ldots, F_h]^T$ such that $\sum_{i=1}^h \alpha_i \is 1$ and $\alpha_i > 0$ for all $1 \leq i \leq h$. The resulting mono-objective function $\sum_{i \is 1}^h \alpha_i\,F_i$ can be solved using any mono-objective optimization technique with guarantee to find a Pareto optimal solution \cite{miettinen:99}.

DP-AOF proceeds in two phases. First, it computes the utopia point $\vec{\setf{F}}^\circ$ by solving as many mono-objective DCOPs as the number of objective functions in the MO-DCOP. DP-AOF uses DPOP to solve these mono-objective DCOPs. It then constructs a new problem building upon the solutions obtained from the first phase. Such a problem is used to assign weights to each objective function of the MO-DCOP to construct the new mono-objective function in the same way as centralized AOF, which then can be solved optimally. The worst case runtime, memory, and communication requirements of this algorithm are the same as those of DPOP, except that the number of operations and the number of messages are larger by a factor of $h$ since it runs DPOP $h$ times to solve the $h$ mono-objective DCOPs. 

\algref{{MO-DPOP}$_{L_p}$}{okimoto:14}
\emph{Multi-Objective $L_p$-norm based Distributed Pseudo-tree Optimization Procedure ({MO-DPOP}$_L{_p}$)} is an incomplete, synchronous, inference-based algorithm. It adapts DPOP using a scalarization measure based on the $L_p$-norm to find a subset of the Pareto front of a MO-DCOP. Similar to DP-AOF, the algorithm proceeds in two phases. Its first phase is the same as the first phase of DP-AOF: It solves $h$ mono-objective DCOPs using DPOP to find the utopia point $\vec{\setf{F}}^\circ$. In the second phase, the agents coordinate to find a solution that minimizes the distance from $\vec{\setf{F}}^\circ$ according to the $L_p$-norm. The algorithm is guaranteed to find a Pareto optimal solution only when the $L_1$-norm (Manhattan norm) is adopted. In this case, {MO-DPOP}$_{L_1}$ finds a Pareto optimal solution \rev{that minimizes the average cost values of all objectives}. The worst case runtime, memory, and communication requirements of this algorithm are the same as those of DP-AOF.

\algref{DIPLS}{wack:14}
\emph{Distributed Iterated Pareto Local Search (DIPLS)} is an incomplete, synchronous, search-based algorithm. It extends the \emph{Pareto Local Search (PLS)} algorithm \cite{paquete:04}, which is a hill climbing algorithm designed to solve centralized multi-objective optimization problems, to solve MO-DCOPs. The idea behind DIPLS is to evolve an initial solution toward the Pareto front. To do so, it starts from an initial set of random assignments, and applies PLS iteratively to generate new non-dominated solutions. DIPLS requires a total ordering of agents and elects one agent as the \emph{controller}. At each iteration, the controller filters the set of solutions by dominance and broadcasts them to the agents in the MO-DCOP. Upon receiving a solution, an agent generates a list of \emph{neighboring solutions} by modifying the assignments of the variables that it controls, and sends them back to the controller. When the controller receives the messages from all agents, it proceeds to filter (by dominance) the set of solutions received, and if a new non-dominated solution is found, it repeats the process. 

\rev{The worst case runtime of this algorithm is $O(\ell k p)$ as the controller agent is required to check the dominance of the newly generated solutions at each iteration. In terms of memory requirement, DIPLS agents use $O(np)$ space to store the Pareto front. Finally, in terms of communication requirement, the controller agent broadcasts messages that contain the current Pareto front. Thus, the message size is $O(np)$.}


\section{Dynamic DCOPs}
\label{sec:dynamic_dcop}

Within a real-world MAS application, agents often act in dynamic environments that evolve over time. For instance, in a disaster management search and rescue scenario, new information (e.g.,~the number of victims in particular locations or priorities on the buildings to evacuate) typically becomes available in an incremental manner. Thus, the information flow modifies the environment over time. To cope with such a requirement, researchers have introduced the \emph{Dynamic DCOP (D-DCOP)} model, \rev{where cost functions} can change during the problem solving process, agents may fail, and new agents may be added to the DCOP being solved. With respect to the categorization described in Section~\ref{sec:dcop_classification}, in the D-DCOP model, the agents are fully cooperative and they have deterministic behavior and total knowledge. On the other hand, the environment is dynamic and deterministic.

\subsection{Definition}

The \emph{Dynamic DCOP (D-DCOP)} model is defined as a sequence of classical DCOPs: $\listf{D}_1 ,\ldots, \listf{D}_T$, where each $\listf{D}_t \is \langle \setf{A}^t, \setf{X}^t, \setf{D}^t, \setf{F}^t, \alpha^t \rangle$ is a DCOP representing the problem at time step $t$, for $1 \leq t \leq T$. The goal in a D-DCOP is to solve the DCOP at each time step optimally. By assumption, the agents have total knowledge about their current environment (i.e.,~the current DCOP) but they are unaware of changes to the problem in future time steps.

In a dynamic system, agents are required to adapt as fast as possible to environmental changes. \emph{Stability} \cite{dijkstra:74,verfaillie:05} is a core algorithmic concept in which an algorithm seeks to minimize the number of steps that it requires to converge to a solution each time the problem changes. 
In such a context, these converged solutions are also called \emph{stable solutions}. 
\emph{Self-stabilization} is a related concept derived from the area of fault-tolerance:
\begin{definition}[Self-stabilization]
A system is \emph{self-stabilizing} \emph{iff} the following two properties hold:
\begin{list}{$\bullet$}{\topsep=1pt \parsep=0pt \itemsep=1pt}
	\item[\iOne]\emph{Convergence:} The system reaches a stable solution in a finite number of steps, starting from any given state. In the DCOP context, this property expresses the ability of the agents to coordinate a joint assignment for their variables that optimizes the problem at time step $t+1$, starting from an assignment of the problem's variables at time step $t$.

	\item[\iTwo]\emph{Closure:} The system remains in a stable solution, 
provided that no changes in the environment happens. 
In the DCOP context, this means that agents do not change the assignment for their variables after converging to a solution.
\end{list}
\end{definition}


Solving D-DCOPs is NP-hard, as it requires to solve each DCOP of the D-DCOP independently. 

\subsection{Algorithms}

In principle, one could use classical DCOP algorithms to solve the DCOP $\mathcal{D}_t$ at each time step $1\leq t \leq T$. However, the dynamic environment evolution encourages firm requirements on the algorithm design in order for the agents to respond automatically and efficiently to  environmental changes over time. In particular, D-DCOP algorithms often follow the \emph{self-stabilizing} property. As in the previous sections, the algorithms are categorized as being either complete or incomplete, according to their ability to determine the optimal solution at each time step.

\subsubsection{Complete Algorithms}

\algref{S-DPOP}{petcu:05b}
\emph{Self-stabilizing DPOP (S-DPOP)} is a synchronous, inference-based algorithm that extends DPOP to handle dynamic environments. It is composed of three  self-stabilizing phases:
\iOne A self-stabilizing DFS pseudo-tree generation, whose goal is to create and maintain a DFS pseudo-tree structure;
\iTwo A self-stabilizing algorithm for the UTIL propagation phase; and 
\iThree A self-stabilizing algorithm for the VALUE propagation phase. 
These procedures work as in DPOP and they are invoked whenever any change in the DCOP problem sequence is revealed. Additionally, \citeA{petcu:05b} discuss self-stabilizing extensions 
that can be used to provide guarantees about the way the system transitions from a valid state to the next, after an environment change. 

The worst case runtime, memory, and communication requirements of this algorithm to solve the DCOP at each time step are the same as those of DPOP.  Additionally, upon changes to the problem, S-DPOP stabilizes after at most $\tau$ UTIL messages and $k$ VALUE messages, where $\tau$ is the depth of the pseudo-tree and $k$ is the number of cost functions of the problem.

\algref{I-ADOPT and I-BnB-ADOPT}{yeoh:11}
\emph{Incremental Any-space ADOPT (I-ADOPT)} and \emph{Incremental Any-space BnB-ADOPT (I-BnB-ADOPT)} are asynchronous, search-based algorithms that extend ADOPT and BnB-ADOPT, respectively.
In the incremental any-space versions of the algorithms, each agent maintains bounds for multiple contexts; in contrast, agents in ADOPT and BnB-ADOPT maintain bounds for one context only. By doing so, when solving the next DCOP in the sequence, agents may reuse the bounds information computed in the previous DCOP. In particular, the algorithms identify \emph{affected agents}, which are agents that cannot reuse the information computed in the previous iterations, and they recompute bounds exclusively for such agents.

The worst case runtime and communication requirements of this algorithm to solve the DCOP at each time step are the same as those of ADOPT. However, since these algorithms have the any-space property, their minimal memory requirements are the same as those of ADOPT but they can use more memory, if available, to speed up the algorithms.

\subsubsection{Incomplete Algorithms}

\algref{SBDO}{billiau:12a}
\emph{Support Based Distributed Optimization (SBDO)} is an asynchronous search-based algorithm that extends the \emph{Support Based Distributed Search} algorithm \cite{harvey:07} to the multi-agent case. It uses two types of messages: \emph{is-good} and \emph{no-good}. Is-good messages contain an ordered partial assignment and are exchanged among neighboring agents upon a change in their value assignments. Each agent, upon receiving a message, decides what value to assign to its own variables, attempting to minimize their local costs, and communicates such decisions to its neighboring agents via is-good messages. No-good messages are used in response to violations of hard constraints, or in response to obsolete assignments. A no-good message is augmented with a \emph{justification}, that is, the set of hard constraints that are violated, and are saved locally within each agent. This information is used to discard partial assignments that are supersets of one of the known no-goods.
The changes of the dynamic environment are communicated via messages, which are sent from the \emph{environment} to the agents. In particular, changes in hard constraints require the update of all the justifications in all no-goods.

\rev{The worst case runtime, memory, and communication requirements of this algorithm are the same as those of SyncBB each time the problem changes.}

\algref{FMS}{ramchurn:10} 
\emph{Fast Max-Sum (FMS)} is an asynchronous inference-based algorithm that extends {Max-Sum} to the Dynamic DCOP model. As in  Max-Sum, the algorithm operates on a factor graph. Solution stability is maintained by recomputing only those factors that changed between the previous DCOP $\mathcal{D}_{t-1}$ and the current DCOP $\mathcal{D}_t$.
\citeA{ramchurn:10} exploit domain-specific properties in a task allocation problem to reduce the number of states over which each factor has to compute its solution. In addition, FMS is able to efficiently manage addition or removal of tasks (e.g.,~factors), by performing message propagation exclusively on the factor graph regions that are affected by such topological changes.
The worst case runtime, memory, and communication requirements of this algorithm to solve the DCOP at each time step are the same as those of Max-Sum. 

FMS has been extended in several ways. \emph{Bounded Fast Max-Sum} provides bounds on the solution found, as well as it guarantees self-stabilization \cite{macarthur:10}. 
\emph{Branch-and-Bound Fast Max-Sum (BnB-FMS)} extends FMS providing online domain pruning using a branch-and-bound technique \cite{macarthur:11}.

\subsection{Notable Variants: D-DCOPs with Commitment Deadlines or Markovian Properties}

\rev{
We now describe several notable variants of D-DCOPs and their corresponding algorithms. 

\algref{RS-DPOP}{petcu:07a} In this proposed model, agents have \emph{commitment deadlines} and \emph{stability constraints}. In other words, some of the variables may be unassigned at a given point in time, while others must be assigned within a specific deadline. Commitment deadlines are either hard or soft. \emph{Hard commitments} model irreversible processes. When a hard committed variable is assigned, its value cannot be changed.  \emph{Soft commitments} model contracts with penalties. If a soft committed variable $x_i^t$ has been assigned at time step $t$, its value can be changed at time step $t' > t$, at the price of a cost penalty. These costs are modeled via \emph{stability constraints}, which are defined as binary relations $s_i : D_i \times D_i \to \mathbb{R}^+$, representing the cost of changing the value of
variable $x_i$ from time step $t$ to time step $t+1$. Given the set of stability constraints $\setf{S} \subseteq \setf{F}$, at each time step~$t$, the goal is to find a solution $\sigma^*_t$:
\rev{
$$
\sigma^*_t \coloneqq \argmin_{\sigma \in \Sigma} \left ( \setf{F}_g(\sigma) + \sum_{s_j \in \setf{S}}  s_j(\sigma^*_{t-1}(x_j), \sigma(x_j)) \right ).
$$ 
}
The latter term accounts for the penalties associated to the value assignment updates for the soft committed variables. RS-DPOP has the same order complexity as S-DPOP.

To solve this problem, \citeA{petcu:07a} extended S-DPOP to RS-DPOP.\footnote{The full name of the algorithm was not provided by \citeA{petcu:07a}.} Like S-DPOP, it is a synchronous, inference-based algorithm. Unlike S-DPOP, it's UTIL and VALUE propagation phases now take into account the commitment deadlines. The worst case runtime, memory, and communication requirements of this algorithm to solve the DCOP at each time step are the same as those of S-DPOP.

\algref{Distributed Q-learning and R-learning}{nguyen:14} In this proposed model, called \emph{Markovian Dynamic DCOPs (MD-DCOPs)}, the DCOP in the next time step $\listf{D}_{t+1}$ depends on the solution (i.e.,~assignment of all variables) adopted by the agents for the DCOP in the current time step $\listf{D}_t$. However, the transition function between these two DCOPs are not known to the agents and the agents must, thus, learn them. The \emph{Distributed Q-learning and R-learning} algorithms are synchronous reinforcement-learning-based algorithms that extend the centralized Q-learning \cite{abounadi:01} and centralized R-learning \cite{schwartz:93,mahadevan:96} algorithms. Each agent maintains Q-values and R-values for each $\sigma_{t-1}, d_i^t$ pair, where $\sigma_{t-1}$ is the solution for the DCOP $\listf{D}_{t-1}$ and $d_i^t$ is the value of its variables in the \rev{cost function} $f_i^t \in \setf{F^t}$. These Q- and R-values represent the predicted \rev{cost} the agent will incur if it assigns its variables values according to $d_i^t$ when $\sigma_{t-1}$ is the previous solution. The agents repeatedly refine these values at every time step and choose the values with the \rev{minimum} Q- or R-value at each time step. 

The worst case runtime, communication, and memory requirements of these two algorithms to solve the DCOP at each time step are the same as those of DPOP, as they use DPOP as a subroutine to update the Q- and R-values. The exception is that agents in the Distributed Q-learning algorithm also broadcast their value assignments at each time step to all other agents. Thus, they send $O(m^2)$ messages in each time step instead of the $O(m)$ complexity of DPOP.\footnote{A single broadcast message is counted as $m$ peer-to-peer messages, where $m$ is the number of agents in the problem.}

A related model is the \emph{Proactive Dynamic DCOPs (PD-DCOPs)} \cite{fioretto:AAMAS-16a,fioretto:AAMAS-17c}, where the transition functions between two subsequent DCOPs are known and can be exploited by the resolution process. Additionally, another key difference between these two models is that the DCOP in the next time step $\listf{D}_{t+1}$ does not depend on the solution in the current time step, but instead depends on the values of the random variables at the current time step. Researchers have introduced a number of offline proactive and online reactive algorithms to solve this problem \cite{fioretto:AAMAS-16a,fioretto:AAMAS-17c}.
}

\section{Probabilistic DCOPs}
\label{sec:probabilistic_dcop}

The DCOP models discussed so far can model MAS problems in deterministic environments. 
However, many real-world applications are characterized by environments with a stochastic behavior. In other words, there are exogenous events that can influence the outcome of an agent's action. For example, the weather conditions or the state of a malfunctioning device can affect the \rev{cost} of an agent's action. To cope with such scenarios, researchers have introduced \emph{Probabilistic DCOP (P-DCOP)} models, where the uncertainty in the state of the environment is modeled through stochasticity in the \rev{cost functions}. With respect to the DCOP categorization described in Section~\ref{sec:dcop_classification}, in the P-DCOP model, the agents are fully cooperative and have a deterministic behavior. Additionally, the environment is static and stochastic. While a large body of research has focused on problems where agents have \emph{total knowledge}, this section includes a discussion of a subclass of P-DCOPs where the agents' knowledge of the environment is limited, and the agents must balance the \emph{exploration} of the unknown environment and the \emph{exploitation} of the known \rev{costs}.

\subsection{Definition}
\label{sec:pdcop_def}

A common strategy to model uncertainty is to augment the outcome of the \rev{cost} functions with a stochastic character \cite{atlas:10,stranders:11,nguyen:12}. Another method is to introduce additional random variables as input to the \rev{cost} functions, which simulate exogenous uncontrollable traits of the environment \cite{leaute:09a,leaute:11,wang:11}. To cope with such a variety, this section introduces the \emph{Probabilistic DCOP (P-DCOP)} model, which generalizes the proposed models of uncertainty. A P-DCOP is defined by a tuple $\langle \setf{A, X, D, F},\alpha, \listf{I}, \Omega, \setf{P}, \mathcal{E}, \mathcal{U} \rangle$, where $\setf{A}$ and $\setf{D}$ are as defined in Definition~\ref{def:dcop}. In addition, 
\bitemize
	\item $\setf{X}$ is a mixed set of decision variables and random variables.

	\item $\listf{I} = \{r_1, \ldots, r_q\} \subseteq \setf{X}$ is a set of \emph{random variables} modeling uncontrollable stochastic events, such as weather or a malfunctioning device.

	\item $\setf{F} = \{f_1, \ldots, f_k\}$ is the set of \rev{cost functions}, each defined over a mixed set of decision variables and random variables, and such that each value combination of the decision variables on the \rev{cost function} results in a probability distribution. 
	\rev{As a result, $f_i$ is itself a random variable, given the local value assignment $\sigma_{\scope{i} \setminus \listf{I}}$ and a realization for the random variables involved in $f_i$.}

	\item $\alpha : \setf{X} \setminus \listf{I} \to \setf{A}$ is a mapping from decision variables to agents. Notice that random variables are not controlled by any agent, as their outcomes do not depend on the agents' actions.

	\item $\Omega = \{\Omega_1, \ldots, \Omega_q \}$ is the (possibly discrete) set of events for the random variables (e.g.,~the different weather conditions or stress levels a device is subjected to) such that each random variable $r_i \in \listf{I}$ takes values in $\Omega_i$. In other words, $\Omega_i$ is the domain of random variable $r_i$.
	        
	\item $\setf{P} = \{p_1, \ldots, p_q\}$ is a set of probability distributions for the random variables, such that $ p_i : \Omega_i \to [0,1] \subseteq \mathbb{R}$ assigns a probability value to an event for $r_i$ and $\int_{\omega \in \Omega_i} p_i(\omega) \ d\omega$ = 1 for each random variable $r_i \in \listf{I}$.
	          		
	\item $\mathcal{E}$ is an \emph{evaluator function} from random variables to real values, that, given an assignment of values to the decision variables, summarizes the  distribution of the aggregated \rev{cost functions}.

	\item $\mathcal{U}$ is a \emph{utility function} that given a random variable 
	returns an ordered set of different outcomes, and it is 
	based on the decision maker preferences. This function is needed when the \rev{cost functions} have uncertain outcomes and, thus, these distributions are not readily comparable.
\eitemize

\noindent The goal in a P-DCOP is to find a solution $\sigma^*$, that is, an assignment of values to all the decision variables, such that:
\begin{equation}
	\label{eq:pdcop_goal}
	\sigma^* \coloneqq  \displaystyle{\argminmax_{\sigma \in \Sigma}} \ \,
	\mathcal{E} \displaystyle \left[
		\sum_{f_i \in \setf{F}} 
		\mathcal{U} \big( 
		 f_i(\sigma_{\scope{i} \setminus \mathcal{I}})
		\big) \right]
\end{equation}
where argmin or argmax are selected depending on the algorithm adopted, $\sum$ is the operator that is used to aggregate the values from the functions $f_i \in \setf{F}$. Typically such an operator is a summation, however, to handle continuous distributions, other operators have been proposed.

The probability distribution over the domain of random variables $r_i \in \listf{I}$ is called a \emph{belief}. An assignments of all random variables in $\listf{I}$ describes a (possible) \emph{scenario} governed by the environment.
As the random variables are not under the control of the agents, they \emph{act} independently of the decision variables. Specifically, their beliefs are drawn from probability distributions. 
Furthermore, they are assumed to be independent of each other and, thus, they model independent sources of exogenous uncertainty. 

The utility function $\mathcal{U}$ enables us to compare the uncertain cost outcomes of the \rev{cost} functions. \rev{In general, the utility function is non-decreasing, that is, the lower the cost, the higher the utility}. However, the utility function should be defined for the specific application of interest. 
For example, in farming, the utility increases with the amount of produce harvested. 
However, farmers may prefer a smaller but highly certain amount of produce harvested over a larger but highly uncertain and, thus, risky outcome. 

The evaluation function $\mathcal{E}$ is used to summarize in one criterion the \rev{costs} of a given assignment that depends on the random variables. A possible evaluation function is the \emph{expectation} function: $\mathcal{E}[\cdot] = \mathbb{E}[\cdot]$.

Let us now introduce some concepts that are commonly adopted in the study of P-DCOPs.

\begin{definition}[Convolution]%
	The \emph{convolution} of the \emph{probability density function (PDF)} $f(x)$ and $g(x)$ of two independent random variables $X$ and $Y$ is the integral of the product of the two functions after one is reversed and shifted:
	\begin{equation}
		\label{def:convolution}
		h(z) = (f * g )(z)\ \coloneqq\ 
		\int_{-\infty}^\infty f(\tau)\, g(z - \tau)\, d\tau =
		\int_{-\infty}^\infty f(z - \tau)\, g(\tau)\, d\tau 
	\end{equation}
\end{definition}
\noindent
It produces a new PDF $h(z)$ that defines the overlapping area between $f(x)$ and $g(y)$ as a function of the quantity that one of the original functions is translated by. 
In other words, the convolution is a method of determination of the sum of two random variables. The counterpart
for the distribution of the sum $Z=X+Y$ of two independent discrete variables is:
\begin{equation}
P(Z=z)=\sum _{k=-\infty }^{\infty }P(X=k)P(Y=z-k).
\end{equation}

In a P-DCOP, the value returned by a function $f_i$, for an assignment on its scope $\scope{i}$, is a random variable $V_i$ 
($V_i \sim f_i(\scope{i})$). Thus, the global value $\sum_{f_i \in \setf{F}} V_i$ is also a random variable, whose probability density function is the convolution of the PDFs of the individual $V_i$'s. Thus, the concept of convolution of two PDFs in a P-DCOP is related to the summation  of the utilities of two \rev{cost} functions in classical DCOPs.

A common concept in optimization with uncertainty is that of \emph{ranking} a set of random variables \{$r_1, r_2, \dots$\} with \emph{Cumulative PDFs (CDFs)} \{$F_1(x), F_2(x), \dots\}$. These distributions are also commonly called \emph{lotteries}, a concept related to that of \emph{stochastic dominance}, which is a form of stochastic ordering based on preference regarding outcomes. It refers to situations where a probability distribution over possible outcomes can be ranked as superior to another. 

The first-order stochastic dominance refers to the situation when one lottery is unambiguously better than another:
\begin{definition}[First-Order Stochastic Dominance] 
	\label{def:fo_stochastic_dominance}
Given two random variables $r_i$ and $r_j$ with CDFs $F_i(x)$ and $F_j(x)$, respectively, $F_i$ first-order stochastically dominates $F_j$ \emph{iff}: 
	\begin{equation}
		\label{eq:fo_stochastic_dominance}
		F_i(x) \leq F_j(x), 
	\end{equation}
for all $x$ with a strict inequality over some interval.
\end{definition}
\noindent If $F_i$ first-order stochastically dominates $F_j$, then $F_i$ necessarily has a strictly \rev{smaller} expected value: \rev{$\mathbb{E} [F_i(x)] < \mathbb{E} [F_j(x)]$}. In other words, if $F_i$ dominates $F_j$, then the decision maker prefers $F_i$ over $F_j$ regardless of his utility function $\mathcal{U}$ is, as long as it is weakly increasing.

It is not always the case that one CDF will first-order stochastically dominate another. In such a case, one can use the second-order stochastic dominance to compare them. The latter refers to the situation when one lottery is unambiguously less \emph{risky} than another:
\begin{definition}[Second-Order Stochastic Dominance] 
	\label{def:so_stochastic_dominance}	
Given two random variables $r_i$ and $r_j$ with CDFs $F_i(x)$ and $F_j(x)$, respectively, $F_i$ second-order stochastically dominates $F_j$ \emph{iff}: 
	\begin{equation}
		\label{eq:so_stochastic_dominance}
		\int_{-\infty}^{c} F_i(x) \, dx \leq \int_{-\infty}^{c} F_j(x) \, dx, 
	\end{equation}
for all $c$ with a strict inequality for some values of $c$.
\end{definition}
\noindent If $F_i$ second-order stochastically dominates $F_j$, then \rev{$\mathbb{E} [F_i (x)] \leq \mathbb{E} [F_j(x)]$}. 
If Equation~\ref{eq:so_stochastic_dominance} holds for all $c \geq c'$, for some sufficiently large $c'$, then $\mathbb{E} [F_i (x)] = \mathbb{E} [F_j(x)]$.
In this case, as both lotteries are equal in expectation, the decision maker prefers the lottery $F_i$, which has less variance and is, thus, less risky.

Another common concept in P-DCOPs is that of \emph{regret}. In decision theory, regret expresses the negative \emph{emotion} arising from learning that a different solution than the one adopted, would have had a more favorable outcome. In P-DCOPs the regret of a given solution is typically defined as the difference between its associated \rev{cost} and that of the theoretical optimal solution. The notion of regret is especially useful in allowing agents to make robust decisions in settings where they have limited information about the \rev{cost} functions. 

An important type of regret is the \emph{minimax regret}. Minimax regret is a decision rule used to minimize the possible loss for a worst case (i.e, maximum) regret. As opposed to the (expected) regret, minimax regret is independent of the probabilities of the various outcomes. Thus, minimax regret could be used when the probabilities of the outcomes are unknown or difficult to estimate.

Solving P-DCOPs is PSPACE-hard, as in general, the process is required to remember a solution for each possible state associated to the uncertain random variables. 
The study of complexity classes for P-DCOPs is largely unexplored. Thus, we foresee this as a potential direction for future research, in which particular focus could be given in determining fragments of P-DCOPs characterized by lower complexity than the one above.

\subsection{Algorithms}

Unlike for Classical DCOPs and Dynamic DCOPs, where the algorithms solve the same problem, P-DCOP algorithms approach the problem uncertainty in different ways and, thus, solve different variants of the problem. This is due to the greater modeling flexibility offered by the P-DCOP framework. As such, the proposed algorithms are often not directly comparable to one another.
We categorize P-DCOP algorithms into \emph{complete} and \emph{incomplete} algorithms, according to their ability to guarantee to find the optimal solutions or not, for a given evaluator and utility functions. Unless otherwise specified the ordering operator in Equation~\ref{eq:pdcop_goal} refers to the $\argmax$ operator.

\subsubsection{Complete Algorithms}

\algref{$\mathbb E$[DPOP]}{leaute:11}
$\mathbb{E}$[DPOP] is a synchronous, sampling-based and inference-based algorithm. It can be either complete or incomplete based on the $\mathbb{E}$[DPOP] variant used, and described below. $\mathbb{E}$[DPOP] uses a \emph{collaborative sampling} strategy, where all agents concerned with a given random variable agree on a common sample set that will be used to estimate the PDF of that random variable. Agents performing collaborative sampling independently propose sample sets for the random variables influencing the variables they control, and elect one agent among themselves as responsible for combining the proposed sample sets into one. The algorithm is defined over P-DCOPs with $\mathcal{I} \neq \emptyset$ and deterministic \rev{cost} function outcomes, that is, for each combination of values for the variables in $\scope{i}$, $f_i(\sigma_{\scope{i} \setminus \mathcal{I}})$ is a degenerate distribution (i.e.,~a distribution that results in a single value) and the utility function $\mathcal{U}$ is the identity function. 
$\mathcal{E}$ is an arbitrary evaluator function summing over all functions in $\setf{F}$. 

$\mathbb{E}$[DPOP] builds on top of DPOP and proceeds in four phases: In Phase~1, the agents order themselves into a pseudo-tree ignoring the random variables. In Phase~2, the agents bind random variables to some decision variable. In Phases 3 and 4, the agents run the UTIL and VALUE propagation phases like in DPOP except that random variables are sampled. Based on different strategies adopted in binding the random variables in Phase~2, the algorithm has two variants \cite{leaute:09a}. In \emph{Local-$\mathbb{E}$[DPOP]}, a random variable $r_i \in \listf{I}$ is assigned to each decision variable responsible for enforcing a constraint involving $r_i$. In this approach, the agents do not collaborate by exchanging information about how their utilities depend on the random variables. In contrast, \emph{Global-$\mathbb{E}$[DPOP]} assigns $r_i$ to the \emph{lowest common ancestor} agent,\footnote{The agent that is separated by the smallest number of tree edges from all variables constrained with the given random variable.} which is responsible for combining the proposed samples. While this additional information can produce higher-quality solutions, both algorithms are generally incomplete. One exception is when the evaluation function $\mathcal{E}$ adopted is \emph{linear}, as in the case of the \emph{expectation} function, in which case the algorithms are complete.

The worst case runtime, memory, and communication requirements of this algorithm are the same as those of DPOP. The exception is the message size of \emph{Global-$\mathbb{E}$[DPOP]}, which is $O(d^{w^*} s^q)$, where $s$ is the largest sample set size; the UTIL messages have this size when the root as well as all leaves of the pseudo-tree are constrained with all $q$ random variables.

\algref{SD-DPOP}{nguyen:12}
\emph{Stochastic Dominance DPOP (SD-DPOP)} operates on a P-DCOP model where $\mathcal{I} = \emptyset$, $\mathcal{E}$ is the second order stochastic dominance criteria, $\mathcal{U}$ is the identity function, and $\Sigma$ denotes the convolution of the distributions $f_i(\sigma_{\scope{i} \setminus \mathcal{I}})$.  
It is a complete synchronous inference-based algorithm that extends DPOP to solve P-DCOPs. Similar to DPOP, it has three phases. In Phase~1, like DPOP, it constructs a pseudo-tree. In Phase~2, instead of summing up costs, the agents convolve \rev{cost} functions, and instead of propagating \rev{costs} up the pseudo-tree, they propagate convolved \rev{cost functions}. In Phase~3, like DPOP, the agents choose values for their variables. However, instead of choosing values that \rev{minimize the cost} of their subtrees, the agents choose their values according to the second-order stochastic dominance criteria.

Like DPOP, SD-DPOP requires a linear number of messages. In addition, in SD-DPOP, VALUE messages contain each Pareto optimal value of the sending agent, and UTIL messages contain a representation of the \rev{cost function} for each Pareto optimal solution and each combination of values of the parent and pseudo-parents of the sending agents. Thus, for continuous PDFs that could be represented by mean and variance, the message size is $O(p d^{w^*})$, where $p$ is the size of the Pareto set. If the \rev{cost functions} are represented by discretized bins, then the message size is $O(b p d^{w^*})$, where $b$ is the maximum number of bins used to represent a \rev{cost function}. The memory requirement of each SD-DPOP agent is  $O(p d^{w^*})$ or $O(b p d^{w^*})$, similarly as above, as they need to store and process the messages received. The worst case runtime requirement and the number of messages sent by SD-DPOP are the same as those of DPOP.

\subsubsection{Incomplete Algorithms}

\algref{DNEA}{atlas:10}
The P-DCOP model proposed by \citeA{atlas:10} is characterized by uncertainty exclusively at the level of the outcome
of the \rev{cost functions}, and not due to random variables. Thus, $\listf{I} = \emptyset$. In addition, the utility function $\mathcal{U}$ is the identity function, while $\mathcal{E}$
is a given evaluator function (e.g.,~the expectation) for the functions $f_i \in \setf{F}$. In this  settings, by employing the evaluation function, \citeauthor{atlas:10} show that one can reduce the uncertainty associated to each \rev{cost function} to the deterministic case. Thus, one can solve the proposed P-DCOP problems using classical DCOP approaches. 

In particular, they propose the \emph{Distributed Neighbor Exchange Algorithm (DNEA)}, which is an incomplete, synchronous, search-based algorithm that is similar to DSA. Each agent starts by assigning a random value to each of its variables and sends this information to all its neighbors. Upon receiving the values of its neighbors, it computes a \rev{cost vector} containing the \rev{costs} for each possible combination of values for all its variables, under the assumption that its neighbors' values are those in the messages received. It then sends this \rev{cost vector} to all its neighbors. Upon receiving the \rev{cost vector} of its neighbors, it computes the best value for each of its variables, assigns those values to its variables probabilistically, and sends the assigned values to all its neighbors. This process repeats until a termination condition is satisfied.  

\rev{The runtime requirement of this algorithm is $O(\ell (ld + d^2))$. In terms of communication requirement, the number of messages sent is $O(\ell nl)$, and the size of each message is $O(d)$. In terms of memory requirement, each DNEA agent needs $O(ld)$ amount of memory to store and process the messages received.}

\algref{U-GDL}{stranders:11}
The P-DCOP model proposed by \citeA{stranders:11} also assumes that the \rev{cost functions} are not dependent on random variables. Thus, $\mathcal{I} = \emptyset$. Additionally, they assume that  
$\mathcal{E}$ is the expectation of the convolution ($\Sigma$) of the distributions $f_i(\sigma_{\scope{i} \setminus \mathcal{I}})$, and $\mathcal{U}$ is a given risk function.
\citeauthor{stranders:11} propose the \emph{Uncertain Generalized Distributive Law (U-GDL)} algorithm, which is an incomplete asynchronous inference-based algorithm similar to Max-Sum, \rev{and operates on acyclic graphs. A cyclic constraint graph $G$ is converted into an acyclic graph $\hat{G}$ by merging variables until the new resulting graph contains no cycles. Merging two variables creates a new variable whose domain is the Cartesian product of the domain of the merged variables.} 
U-DGL extends the \emph{Generalized Distributive Law (GDL)} algorithm \cite{aji:00} by redefining \rev{the $(\min, +)$}\footnote{\rev{U-GDL was originally defined for maximization problems. Its  presentation is adapted to minimization problems for consistency of the DCOP models objective presented in this survey.}} algebra to the setting where \rev{costs} are random variables rather than scalars. The $+$ operator is extended to perform convolution of two random variables. To cope with the potential issue  that not all PDFs are closed under convolution, \citeA{stranders:11} suggest to resort to sampling methods to approximate such operations. The \rev{$\min$ operator} is defined to distribute over convolution and to select the \rev{minimal} elements from a set of random variables based on their \rev{expected cost}. 
In order to filter partial potential solutions that can never achieve global optimality, the authors introduce a first-order stochastic dominance condition, which is employed in the context of the \rev{$\min$ operator}. They also discuss \emph{necessary} and \emph{sufficient} conditions for dominance, where the former discards all dominated solutions, but it might also discard some non-dominated solution -- this is equivalent to using a classical DCOP algorithms to solve the P-DCOP model adopted in their work. The latter preserves optimal solutions, but retains, in general, sub-optimal ones as well.

\rev{The runtime requirement of this algorithm is $O(p\hat{d}^l)$, where $\hat{d}$ is the size of the largest domain of the merged variables in $\hat{G}$. In terms of memory requirement, each U-GDL agent needs $O(p\hat{d}^l)$ space to store all value combinations of neighboring variables for each solution in its current Pareto frontier. In terms of communication requirement, the number of messages sent is $O(\delta(\hat{G}))$, where $\delta(\hat{G})$ is the diameter of the resulting acyclic graph, and the size of each message is $O(p\hat{d})$ as agents need to send the current aggregated costs of all the agent's variable's values for each solution in its current Pareto frontier.}

\subsection{Notable Variant: P-DCOPs with Partial Agent Knowledge}

This section describes a class of Probabilistic DCOPs where agents have \emph{partial} knowledge about the environment. In other words, the \rev{cost functions} are only partially known and, therefore, agents may discover the unknown \rev{costs} via \emph{exploration} \cite{taylor:11}. The new model aims at capturing those domains where agents have an ``explorative nature,'' i.e.,~ one of the agents' goals is to acquire knowledge about the environment in which they act. Agents are concerned with a total, online, \rev{cost} achievable in a limited time frame. 
In this context, agents must balance the coordinated \emph{exploration} of the unknown environment and the \emph{exploitation} of the known portion of the \rev{costs}, in order to \rev{minimize the global utility}. This model was originally called \emph{Distributed Coordination of Exploration and Exploitation (DCEE)} \cite{taylor:10}.

\subsubsection{Definition}
The P-DCOP model for agents with partial knowledge is described by extending the P-DCOP model introduced in Section~\ref{sec:pdcop_def}, as follows: $\langle \setf{A, X, D, F},\alpha, \listf{I}, \Omega, \setf{P}, \mathcal{E}, \mathcal{U}, T \rangle$, where $T>0$ is a finite time horizon characterizing the time within which the agents can exploit the unknown \rev{cost functions} and explore the search space. The goal in such a P-DCOP problem is to find a set of complete assignments $\vec{\sigma}^* = [\sigma_1^*, \ldots, \sigma_T^*]$ that \rev{minimizes the utility} of the cumulative \rev{cost} within the finite time horizon $T$: 
\begin{equation}
	\label{eq:pdcop2_goal}
	\vec{\sigma}^* \coloneqq \argminmax_{\sigma^1, \ldots, \sigma^T} \ \, 
	\mathcal{E} \displaystyle \left[
		\sum_{t=0}^T \,
		\sum_{f_i \in \setf{F}} 
		\mathcal{U} \big( 
		 f_i(\sigma_{\scope{i} \setminus \listf{I}}^t)
		\big) \right]
\end{equation}
where $\sigma^t \in \Sigma$ denotes a solution at time step $t$. In other words, agents have at most $T$ time steps to modify the value of their decision variables, and solve $T$ P-DCOP problems by acquiring more and more knowledge on the environment as the time unrolls.

\subsubsection{Algorithms}
In a stochastic and unknown environment, the \rev{cost functions} need to be learned online through interactions between the agents and their environment. Thus, the algorithms presented in this section are targeted to coordinate agents to solve a sequence of optimization problems in order to simultaneously reduce uncertainty about the local \rev{cost functions} (exploration) and optimize the global objective (exploitation). In addition, the following algorithms are incomplete.

\algref{BE-Rebid}{taylor:10}
The \emph{Balanced Exploration Rebid (BE-Rebid)} is a synchronous, search-based algorithm that solves P-DCOPs with $\listf{I} = \emptyset$, and
$\mathcal{U}$ and $\mathcal{E}$ are the identity functions.   
It extends MGM as it calculates and communicates its expected gain. The algorithm is introduced in the context of a \emph{wireless sensor network} problem, where agents can perform small movements in order to enhance their communication capabilities, which are characterized by the distance between pairs of agents. 
Each agent can perform three actions: \emph{stay} in the current position, \emph{explore} another position, or \emph{backtrack} to a previously explored position and halt movement. In each time step, BE-Rebid computes the expected \rev{cost} of executing the \emph{explore} or \emph{backtrack} actions, assuming complete knowledge of the underlying distribution of the \rev{cost functions}. Exploring is evaluated by using order statistics and is based on the \rev{cost} of the best value found during exploration. Backtracking to a known position results in a \rev{cost} associated to the backtracked state for the remainder of the time steps (i.e.,~it \emph{stays} in that state).

Following the region-optimal approaches presented in the context of classical DCOPs, \citeA{taylor:11} propose a version of the algorithm, called \emph{BE-Rebid-2}, that allows pairs of agents to  explore in a coordinated fashion. Interestingly, in such settings, the authors find that increasing coordination (measured by the number of agents that can execute a joint action) can decrease solution quality. This phenomenon is referred to as  \emph{team uncertainty penalty}. 
\rev{
The worst case runtime and communication requirements of this algorithm are the same as those of MGM. 
}

\algref{Heist}{stranders:12}
\emph{Heist} is a synchronous, inference-based algorithm that solves P-DCOPs with $\listf{I} = \emptyset$, 
$\mathcal{E}$ is the expectation function, and $\mathcal{U}$ is the identity function. 
It aims at \rev{minimizing the expected utility of the cumulative cost function} $\setf{F}_g$, within the finite time horizon $T$. It does so by modifying a \emph{Multi-Armed Bandit (MAB)} approach \cite{vermorel:05} to a distributed scenario. A MAB is a slot machine with multiple arms, each of which yields a \rev{cost} drawn from an unknown but fixed probability distribution. It trades exploration and exploitation by pulling the arms in order to \rev{minimize the cumulative cost} over a finite horizon. To cope with the uncertain and stochastic nature of the \rev{cost functions}, Heist models each \rev{cost function} as a MAB, such that the joined assignment of the variables in the scope of the given \rev{cost function} becomes an arm of that bandit. It seeks to \rev{minimize the expected cumulative optimization cost} received over a finite time horizon by repeatedly pulling the MAB arms to select the joint action with the highest estimated \emph{Upper Confidence Bound (UCB)} \cite{auer:02} on the sum of the local gains received in a single time step. To do so, it employs a belief propagation algorithm, known as \emph{Generalized Distributive Law (GDL)} \cite{aji:00}, in order to \rev{minimize} the UCB in a decentralized fashion. 
\citeA{stranders:12} show that Heist enables agents to balance between exploration and exploitation, and derive optimal asymptotic bounds on the regret of the global cumulative \rev{cost} attained.

\rev{The worst case runtime requirement of this algorithm is $O(\ell T d^{l})$ as each agent computes the maximum marginal UCB for its variable assignment for each time step before the horizon. In terms of memory requirement, each Heist agent needs $O(d^l)$ space to store all value combinations of neighboring variables. In terms of communication requirement, the number of messages sent is $O(\ell Tl)$, one to each neighbor and time step in each iteration, and the size of each message is $O(Td)$ as it contains the aggregated costs of all the agent's variable's values for each time step.}

\algref{ICG-Max-Sum}{wu:13}
The \emph{Iterative Constraint Generation Max-Sum (ICG-Max-Sum)} algorithm is a synchronous, inference-based algorithm that solves P-DCOPs with 
$\listf{I} \neq \emptyset$, 
$\mathcal{E}$ is the identity function, and 
$\mathcal{U} (f_i (\cdot))$ is the maximal regret function. 
The algorithm aims at minimizing the sum of maximal regrets for all the functions in $\setf{F}$.
Furthermore, the horizon is $T=1$. Thus, unlike the previous algorithms, IGC-Max-Sum does not attempt to learn the outcome of the \rev{cost functions}. Its objective is to find \emph{robust} solutions to the uncertain problem distributions; it does so by finding the solution that minimizes the maximum regret. 
The algorithm extends the \emph{Iterative Constraint Generation (ICG)} method \cite{benders:62,regan:10} to the decentralized case by decomposing the overall problem into a \emph{master} problem and a \emph{subproblem} that are iteratively solved until convergence. At each iteration, the resolutions of these problems are attained by using Max-Sum. The \emph{master} problem solves a relaxation of the minimax regret goal, where only a subset of all possible joint beliefs is considered, attempting to minimize the loss for the worst case derived from the considered joint belief. Once it generates a solution, the \emph{subproblem} finds the maximally violated constraint associated to such a solution. This is referred to as the \emph{witness} point, indicating that the current solution is not the best one in terms of the minimax regret. This point is added to the set of joint beliefs considered by the master problem, and the process is repeated until no new witness points can be found. 

\rev{The worst case runtime requirement of this algorithm is $O(\ell |\listf{I}| d^{l})$, which is dominated by the master problem, whose computation is exponential in the number of variables in the scope of the associated \rev{cost function} for each belief and iteration of the algorithm. In terms of memory requirement, each ICG-Max-Sum agent needs $O(|\listf{I}| d^{l})$ space (dominated by the master problem again) to store all value combinations of neighboring variables for each belief. In terms of communication requirement, the number of messages sent is the same as that of Max-Sum since two parallel iterations of Max-Sum is executed in each ICG-Max-Sum iteration. However, the size of each message is $O(|\listf{I}| d)$ as it contains the aggregated cost of all the agent's variable's value for each belief.}

A variation of this algorithm that aims at minimizing the expected regret, rather than minimizing the maximum regret, was introduced by \citeA{le:16a}.


\section{Quantified DCOPs}
\label{sec:noncooperative_dcop}

The various extensions of the DCOP model discussed so far differ from each other in terms of agent behavior (deterministic vs.~stochastic), agent knowledge (total vs.~partial), environmental behavior (deterministic vs.~stochastic), and environment evolution (static vs.~dynamic). However, in terms of the agent teamwork, all of these models assume the agents are fully cooperative. Researchers have introduced the \emph{Quantified DCOP (QDCOP)} model \cite{matsui:10}, which assumes a subset of agents to be adversarial, that is, the agents are \emph{partially cooperative} or \emph{competitive}.

\subsection{Definition}
\label{sec:noncooperative_dcop:definition}

The \emph{Quantified DCOP (QDCOP)} model \cite{matsui:10} adapts the \emph{Quantified Constraint Satisfaction Problem (QCSP)} \cite{benedetti:08} and \emph{Quantified Distributed CSP (QDCSP)} \cite{baba:10,baba:11}  models to DCOPs. In QCSPs and QDCSPs, all variables are associated to quantifiers and the constraints should be satisfied independently of the value taken by universally quantified variables. Analogously, in QDCOPs, existential ($\exists$) and universal ($\forall$) quantifiers are introduced to differentiate the cooperative agents from the adversarial ones. 


A \emph{QDCOP} has the form $Q(\setf{F}) \coloneqq q_0 x_0 \ldots q_n x_n$.\footnote{In the original proposal, the set $\setf{F}$ is separated in a set of constraints $C$, representing relationships among variables, and a set of functions $F$, assigning values to each valid assignment.} $Q$ is a sequence of quantified variables, where each $q_i \in \{\exists, \forall\}$ quantifies the variable $x_i$. The goal of a QDCOP is to find a global optimal solution of the corresponding DCOP. However, a universally quantified variable is not coordinated nor assigned, as the result has to hold when it takes any value from its domain. In contrast, an existentially quantified variable takes exactly one  value from its domain, as in (\emph{cooperative}) DCOPs. Thus, the optimal solution of a QDCOP may be different from that of the corresponding DCOP. While a DCOP solution defines a single value, associated to its cost, a QDCOP defines \emph{upper} and \emph{lower bounds} to the optimal solution. In particular, the best choice in a QDCOP defines the smallest \emph{lower bound}. In the worst case, the universally quantified variables can worsen the overall objective as much as possible. Therefore, the worst case defines the smallest \emph{upper bound}. While finding an optimal solution for a DCOP is NP-hard, solving a QDCOP is, in general, P-SPACE-hard \cite{benedetti:08,Lallouet2015}.

\subsection{Algorithms}
\label{sec:noncooperative_dcop:algorithms}

QDCOPs impose a rigid order on the variables, which reflects the correct order of evaluation of the quantifiers. Therefore, classical DCOP algorithms cannot be directly applied to solve QDCOPs. 
\citeA{matsui:10} proposed several variations of ADOPT to solve QDCOPs, which are all based on a DFS pseudo-tree ordering. To keep the ordering of the quantifiers unchanged, the pseudo-tree can be reshaped by applying extra null edges for each pair of nodes, if necessary.

\rev{All the algorithms presented here are complete, and are based on the intuition that universally quantified variables can be seen as adversarial virtual agents, whose goal is to minimize the overall objective}. Following this intuition and the pseudo-tree modifications discussed above, pseudo-tree-based DCOP algorithms can be extended to solve QDCOP.

\algref{Min-max ADOPT}{matsui:10}
\emph{Min-max ADOPT} is an asynchronous, search-based algorithm that extends ADOPT to solve QDCOPs. It uses VALUE messages to communicate values of the variables, and COST messages to announce their \rev{costs}, similar to ADOPT. Each agent, starting from the root of the pseudo-tree, assigns values to its variables and propagates them to its neighboring agents with lower priority. Upon receiving VALUE messages from all higher-priority neighbors, the agent updates its context and repeats the same process  by choosing an assignment that \rev{minimizes its local cost}. In Min-max ADOPT, the existentially quantified variables are used to compute the lower bound, while the universally quantified variables are used to compute the upper bound. This process is executed until the root agent detects that the upper bound is equal to the lower bound. This algorithm has a relatively simple structure and does not adopt any major pruning strategy.

\algref{Alpha-beta ADOPT}{matsui:10}
\emph{Alpha-beta ADOPT} is an asynchronous, search-based algorithm that extends Min-max ADOPT by adapting the \emph{alpha-beta} search strategy, a common pruning strategy adopted in \emph{game-tree search}. This strategy employs two boundary parameters, \emph{alpha} and \emph{beta}, representing the lower bound and the upper bound for each possible \rev{cost} of an assignment, respectively. Alpha represents the lower bound, controlled by the universally quantified variables, while beta represents the upper bound, controlled by the existentially quantified variables. 
Lower bound and upper bound can be modified exclusively by universally quantified and existentially quantified variables, respectively.
In Alpha-beta ADOPT, when an agent reports the \rev{cost} value of the current partial assignment, its parent reduces the alpha/beta threshold accordingly. Thus, the new alpha/beta values are used to prune the search when an agent detects that the current assignment cannot be better than any other solution already evaluated. Alpha and beta values are obtained using a backtracking technique similar to  how thresholds are obtained through backtracking in the original ADOPT.

\algref{Bi-threshold ADOPT}{matsui:10}
\emph{Bi-threshold ADOPT} extends ADOPT by employing two \emph{backtracking thresholds} instead of one as in ADOPT. In ADOPT, each agent $a_i$ maintains the threshold invariant $lb_i^* \leq t_i \leq ub_i^*$, where $lb_i^*$ and $ub_i^*$ are the smallest lower and upper bounds, respectively, of the agent over all of its values, and $t_i$ is the threshold of the agent. In contrast, in Bi-threshold ADOPT, each agent maintains the threshold invariant $lb_i^* \leq t_i^\alpha \leq t_i^\beta \leq ub_i^*$, where $t_i^\alpha$ is a lower bound on the threshold, similar to alpha in Alpha-beta ADOPT, and $t_i^\beta$ is an upper bound on the threshold, similar to beta in Alpha-beta ADOPT.


\section{DCOP Applications}
\label{sec:applications}

DCOP models have been adopted to represent a wide range of MAS applications, thanks to their 
 ability  to capture essential and fundamental MAS aspects as well as the  support for the development of general domain-independent algorithms. This section describes some of the most compelling applications as well as a general overview of their corresponding DCOP models. A comprehensive list of DCOP applications, categorized according to the DCOP classification of Table~\ref{tab:dcop_classification}, is given in Table~\ref{tab:dcop_applications}. 

\begin{table}[t]
	\resizebox{0.9\linewidth}{!}
{
	\centering
	\begin{tabular}{|c | r | c c c |}
		\hline
		\multicolumn{1}{|c}{\bf Group} & 	\multicolumn{1}{c}{\bf Problem} & 	\multicolumn{3}{c|}{\bf Model} \\ 
		& & AK & EB & EE \\
		\hline\hline
		 \multirow{1}{*}{\sc Disaster Management}
			&  \multirow{1}{*}{\sl Disaster Evacuation} & T & D & S/D \\
		{\sc \& Coordination} 
			&  {\sl Coalition Formation} & P & D & D \\
		\hline
		\multirow{1}{*}{\sc Radio Frequency Allocation} 
		& \multirow{1}{*}{\sl Cooperative Channel Assignment} & T & D & S/D \\
		\hline
		\multirow{1}{*}{\sc Recommendation Systems} & {\sl Group Recommendation} & T & D & S \\
		\hline
		\multirow{4}{*}{\sc Scheduling} 
		& {\sl Distributed Meeting Scheduling} 		& T & D & S \\
		 & {\sl Water Allocation Scheduling} 			& T & D & S \\
		& {\sl Vessel Rotation Planning}  				& T & D & S \\
		& {\sl Patient Scheduling}  					& T & D & S/D \\

		\hline
		\multirow{4}{*}{\sc Sensor Network} 
		& {\sl Target Tracking} 		  	 	& T & D/S & D \\
		& {\sl Robotic Network Optimization}	& P & D & D \\
		 & {\sl Mobile Sensor Team}	 			& P & S & D \\
		& {\sl Sensor Sleep Scheduling}  		& T & D & S \\
		\hline
		\multirow{2}{*}{\sc Service-Oriented Computing}
		& {\sl Application Component Placement} 	& T & D & S \\
		& {\sl Server Allocation} 					& T & D & D \\
		\hline
		 \multirow{5}{*}{\sc Smart Grid and Smart Buildings} 
		& {\sl Economic Dispatch} 			& T & D/S & S/D \\
		& {\sl Power Supply Restoration}  	& T & D & S \\
		& {\sl Microgrid Islanding} 		& T & D & S \\
		& {\sl Prosumer Energy Trading} 	& T & D/S & S \\
		& {\sl Smart Building Devices Scheduling} & T & D/S & S/D \\
		\hline
		\multirow{1}{*}{\sc Supply Chain Management} 
		&  {\sl Supply Chain Formation} 	& T & D & S/D \\
		\hline
		\multirow{1}{*}{\sc Traffic Control}
		& {\sl Traffic Light Synchronization} 	& T/P & S & D \\
		\hline
	\end{tabular}
	}
	\caption{DCOP Applications. AK = Agent Knowledge, it can be [T]otal or [P]artial; EB = Environment Behavior, it can be [D]eterministic or [S]tochastic; EE = Environment Evolution, it can be [S]tatic or [D]ynamic \label{tab:dcop_applications}}
\end{table}

\subsection{Disaster Management and Coordination Problems}
\label{sec:app:disaster}
Disaster management and coordination problems refer to how to efficiently and effectively respond to an emergency. In these scenarios, low-powered mobile devices that require limited bandwidth are often deployed and utilized to assist the disaster management process. Due to their decentralized nature, the DCOP approach fits naturally with this application. A description of several problems within this application domain is presented next. 

\domainref{Disaster Evacuation Problems}
In a disaster scenario, moving evacuees to the closest refuge shelter can quickly overwhelm shelter capacities. A number of researchers have proposed a DCOP model for disaster evacuation, in which several groups of evacuees have to be led to available shelters \cite{carpenter:07,kopena:08,lass:08a,lass:08b,kinoshita:13}. Group leaders can communicate via mobile devices to monitor and coordinate actions. Each group is represented by a DCOP agent managing variables that represent shelter allocations. Thus, the domain of each variable corresponds to the available shelters. Group sizes and shelter capacity, as well as additional group requirements (e.g.,~medical needs) and the distance of a group to shelters, are encoded as cost functions. Solving the DCOP ensures an assignment of all groups to shelters that minimizes overflow, such that groups receive the services they need and their travel distances are minimal.

\domainref{Coalition Formation with Spatial and Temporal Constraints Problems}
In a \emph{Coalition Formation with Spatial and Temporal Constraints (CFST)} problem \cite{ramchurn:10,steinbauer:12,pujol:15}, ambulance and fire brigade agents cooperate in order to react efficiently to an emergency scenario so as to rescue victims and extinguish fires located in different locations. Agents can travel from one location to another in a given time. Each task (i.e.,~rescuing a victim, extinguishing a fire) has a deadline, representing the time until which the victim will survive, and a workload, denoting the amount of time necessary to rescue the victim or put out the fire. The locations of the victims and the fires may be unknown to the agents, and need to be discovered at runtime, which requires  agents to dynamically update the sequence of the tasks they will attempt, taking account of two main constraints: (1) Spatial constraints, which model where an agent can travel and at what time; and (2) Temporal constraints, which model task deadlines and completion times. Agents may also form coalitions to execute a given task faster or if the requirements of a given task cannot be met by a single agent. Hence, the agents' arrival times at each task need to be coordinated in order to form the desired coalition. The objective is to maximize the number of tasks to be completed. 

A DCOP formalization for the CFST is described by \citeA{ramchurn:10}, where ambulances and fire brigades are modeled as DCOP agents. Each agent controls a variable that encodes the current task that the agent will attempt, and whose domain represents task locations. Unary constraints restrict the set of reachable locations, according to distance from a destination and the victim's deadline. Agents' coalitions are defined as groups of agents traveling to the same location. Each task is associated to a utility, which encodes the success for a coalition to complete such task. The goal is to find an assignment of agents to tasks that maximizes the overall utility.

\rev{A variant of the CFST problem, called the \emph{Law Enforcement Problem (LEP)}, was introduced by \citeA{amador:14}. Similar to the CFST problem, in a LEP, police officers need to execute a number of tasks, and may form coalitions to improve their response quality. Additionally, and differently from the CFST, new tasks can be revealed to the agents dynamically over time. Agents can choose to interrupt their current task to perform a new revealed task at the cost of a penalty that is proportional to the importance of the task being interrupted.}

\subsection{Radio Frequency Allocation Problems}
The performance of a \emph{wireless local area network (WLAN)} depends on the channel assignments among neighboring \emph{access points (APs)}. Neighboring transmissions occurring in APs on the same channel or adjacent channels degrade network performance due to transmission interference. In dense urban areas, different APs may belong to different administrative domains, whose control is delegated to different entities. Thus, a distributed approach to the channel assignment is necessary.

\domainref{Cooperative Channel Assignment Problems}
In a cooperative channel assignment problem \cite{hollos:04,monteiro:12a,monteiro:12b}, APs need to be configured in order 
to reduce the overall interference between simultaneous  transmissions on neighboring channels.
\citeA{monteiro:12a,monteiro:12b} proposed a DCOP-based approach for cooperative channel assignment in WLANs where APs may belong to different administration entities. In the proposed model, each AP is represented by a DCOP agent, which controls a decision variable modeling a choice for the AP's channels. The signal-to-interference-and-noise ratio perceived by an AP or a client is modeled as a cost function, as the overall concurrent transmissions occurring in the same channel and in partially overlapped adjacent channels. The goal is to find an assignment of channels to APs that minimizes the total interference experienced in the WLAN.

\citeA{xie:07} and \citeA{mir:10} study dynamic solutions to the problem of allocating and utilizing the wireless network's available spectrum. 
In such a problem, the agents operate in a dynamic radio frequency environment that is composed of time-varying interference sources, which are periodically sampled and measured.

\subsection{Recommendation Systems}
Recommendation systems are tools that provide user-tailored information about items to users. These systems provide information that is tailored to the characteristics and preferences of the users.

\domainref{Group Recommendation Problems}
Like individual recommendations, group recommendations need to take into account the preferences of all group members and formulate a recommendation that suits the whole group.
\citeA{lorenzi:08} propose a DCOP-based travel package recommendation system for groups of users. The users in the group share a common goal (the travel package recommendation), and have individual preferences for each travel service (hotel, flight companies, tour operators, etc). In such a problem, the objective is to find a group recommendation that optimizes the users' preferences. The proposed DCOP solution is composed of two types of agents: user agents and recommender agents. Each user agent controls a decision variable that models that user's travel choices, while each recommender agent controls a decision variable that models a travel service supplier's recommendations. User travel preferences are modeled via unary constraints on the user agents' decision variables. Binary constraints between user agents in a group and their associated recommender agent ensure that each user's choice in the group is compatible to the recommendation of the recommender agent. The goal is to find the best recommendation for the entire group.

\subsection{Scheduling Problems}
\label{sec:app:scheduling}
Scheduling problems are an important class of problems that have been long studied in the area of constraint programming and operations research \cite{giffler:60,solomon:87,minton:92,hentenryck:09}. In such problems, time schedules are to be associated to resource usage. The problem is made particularly difficult when the scheduling process needs to be coordinated in a distributed manner across several entities. In such a context, many scheduling problems can be naturally mapped to DCOPs.

\domainref{Distributed Meeting Scheduling Problems}
The distributed meeting scheduling problem captures generic scheduling problems where one wishes to schedule a set of events within a time range \cite{jennings:95,garrido:96}. Each event is defined by: 
\iOne the resources required for it to be completed, 
\iTwo the time required for it to be completed, within which it holds the required resources, and 
\iThree the cost of using such resources at a given time. A scheduling conflict occurs if two events with at least one common resource are scheduled in overlapping time slots. The goal is to maximize the utilities over all the resources, defined as the net gain between the opportunity benefit and opportunity cost of scheduling various events. 

\citeA{maheswaran:04} discuss three possible DCOP formulations for this problem: \emph{Time slots as variables (TSAV)}, \emph{events as variables (EAV)}, and \emph{private events as variables (PEAV)}. We describe the EAV formulation and refer the reader to the original article for the other two formulations and additional details. In the EAV formulation, events are considered as decision variables. Each variable can take on a value from the time slot range that is sufficiently early to schedule the required resources for the required amount of time, or zero to denote that an event is not scheduled. If a variable takes on a non-zero value, then all its required resources cannot be assigned to any other overlapping event.

\domainref{Water Allocation Scheduling Problems}
The management of water resources in large-scale systems is often associated with multiple institutionally-independent decision makers, which may represent different and conflicting interests, such as flood prevention, hydropower production, and water supply \cite{giuliani:14}. The aim of such problems is to find an efficient use of water allocation and distribution according to the different users' interests.

\citeA{giuliani:14} formalize a regulatory mechanism in water management as a DCOP. The model involves several \emph{active} human agents and \emph{passive} ecological agents. 
Each agent is associated with an objective function that it seeks to maximize. Active agents make decisions about the amount of water to divert from the river or to be released from a dam in order to maximize their corresponding objective functions. Passive agents, on the other hand, represent ecological interests through their associated objective functions and do not  make decisions. The agents model different water supplies for cities and agricultural districts, hydropower productions, and ecological preservation. The goal is to optimize the agents' objective functions, satisfying hard (physical) constraints and maximizing the soft (normative) constraints, which aim at protecting the interests of the passive agents. A solution to such problem, which makes use of a multi-objective DCOP formalization, is presented in \cite{amigoni:15}.

\domainref{Vessel Rotation Planning Problem}
\rev{
In a large port, vessels are scheduled to visit different terminals for loading and unloading operations. The sequence of terminals visited by a vessel is called the \emph{vessel rotation}. Due to the different nature of the terminal visits (e.g.,~loading, unloading containers, different inland shipping activities) there are often dependencies between activities performed at the terminals. Additionally, vessel operators have their own preferences on when to visit a particular terminal. 
The \emph{vessel rotation planning problem (VRPP)} \cite{li:16} describes the problem of assigning rotations to vessels, consisting of sequences of visits to terminals and arrival and departure times of vessels to terminals in a port area, while satisfying the terminals' visits dependencies  and taking into account operators' preferences. 
Due to the geographically distributed nature of the resources (vessels, terminals, loading facilities) and the distributed coordination process undertaken by vessels operators, this problem fits naturally in the DCOP framework. 

\citeA{li:16} proposed a DCOP model for the VRPP, where vessels and terminals are modeled as DCOP agents. The problem's variables are the time slot for which a vessel $i$ is at terminal $j$ ($x_{ij}$), the arrival and departure times of vessel $i$ at terminal $j$ ($a_{ij}$ and $d_{ij}$, respectively), and the number of time steps a vessel $i$ waits at terminal $j$ ($w_{ij}$). A vessel agent $i$ governs variables $x_{ij}, a_{ij}, d_{ij}$, and $w_{ij}$, for each terminal $j$ that it can  visit, while a terminal agent controls auxiliary variables $y_{ij}$ for each vessel $i$ that can visit $j$, which has the same value of $x_{ij}$ and is used by terminal agents to represent the terminal capacities.
The problem constraints represent the preferences of vessels of being at a given terminal in a given time, the penalty occurring for a waiting vessel, and the time windows during which a vessel can visit a terminal. 
}

\domainref{Patient Scheduling Problems}
Medical appointments scheduling problems are related to meeting scheduling problems, as they need to associate patients to resources (e.g.,~doctors, medical machinery) and times, but they require  different types of constraints. Patients may require several services from different departments within the same hospital or in multiple hospitals. In general, the objective is to minimize the patient treatment waiting time under limited resource conditions, as well as to ensure efficient resource usage, taking into account patient preferences. 

\citeA{hannebauer:01} formulate the problem of scheduling patients to diagnostic units in an hospital as a DCOP, where appointments are modeled as variables, whose domains describe times, durations, and locations. The constraints of the problem model the schedule feasibility, the patient preferences over hospitalization times, the workplace constraints, which restricts the types of appointment for a given workplace, and diagnostic unit constraints, which model resource usage.

\citeA{billiau:12b,billiau:12c} propose a Dynamic DCOP model for a radiotherapy patients scheduling problem. In this problem, each agent represents a patient, and it controls variables that represent private information (e.g.,~type of tumor, number of radiation doses per day, the use of chemotherapy) and public information (e.g.,~current schedule of the radiotherapy machine). The constraints of the problem model the duration of each daily treatment, as well as tumor-specific treatment restrictions. The problem objective considers patient waiting times to receive their treatment, patient priorities (based on tumor aggressiveness), and patient preferences.

\subsection{Sensor Network Problems}
\label{sec:app:sensnet}

Sensor networks typically consist of a large number of inexpensive and autonomous sensor nodes, constrained by a limited communication range and battery life. These networks have been deployed for environmental sensing (temperature, humidity, etc.), military applications (e.g.,~battlefield surveillance), and target tracking \cite{akyildiz:02}. When deploying sensor networks, it may not be possible to pre-determine the position of each sensor node. The distributed nature of the problem and the presence of several communication and sensing constraints create a natural fit for DCOPs to solve a wide range of related applications.

\domainref{Target Tracking Problems}
In a target tracking application \cite{zhang:05,matsui:08b,jain:09,ota:09,stranders:09,hosseini:13}, a collection of small Doppler sensors are scattered in an area to detect possible moving targets in the region. Each sensor is battery-powered, can communicate with one another through radio communication, and can scan and detect an object within a fixed range. Communication incurs an energy cost. Thus, to save energy, a sensor may turn itself off. Multiple sensors may be necessary to detect a single target with high accuracy. The overall objective is to maximize the number of targets detected, as quickly as possible and, at the same time, preserve energy so as to prolong the system's lifetime

\citeA{zhang:05} models a simplified version of the above problem as a weighted graph coloring problem, where the total weight of violated constraints needs to be minimized. A node corresponds to a sensor, an edge between two nodes represents the constraint of a shared region between agents, and the weight captures the importance of the common region. The size of the common region reflects the amount of energy loss when two sensors scan the shared region at the same time. Each color corresponds to a time slot in which a sector is scanned. A node must have at least one color so that the corresponding sector is scanned at least once. This graph coloring problem is mapped to a DCOP, where agents represent nodes, agent's variables represent the agents decision on their color, and cost functions represent the graph edges.

\citeA{hosseini:13} use a hierarchical  DCOP approach to scale to larger problems. The authors partition the original problem into $n$ local regions, and use $n$ DCOPs to solve the smaller subproblems. Their solutions are then combined in a hierarchical approach, solved by a DCOP that encompasses variables and constraints shared among the connected regions of the lower hierarchy DCOPs.

\domainref{Robotic Network Optimization Problems}
The robotic network optimization problem describes a sensor network problem where sensors are placed on top of robots that have limited movement capability. In such a problem, robots can make small movements to optimize the wireless connectivity with their neighbors, without affecting the network topology \cite{choxi:07,jain:09}. 

\citeA{jain:09} proposed a DCOP formulation where each robot is represented by an agent. Each agent controls one variable describing the decision on the robots' possible movements. Thus, the variables' domains consist of the valid positions the agent can move to. The cost functions of the problem model the power loss (or gain, depending on the optimization criteria) of the wireless link from a transmitter and a receiver robot, and depend on their positions. Radio communication in wireless sensor networks have a predictable signal strength loss that is roughly inversely proportional to the square of the distance between transmitter and receiver. However, radio wave interference is very difficult to predict \cite{molisch:12}. Thus, \citeA{jain:09} use a P-DCOP-based approach with partial agent knowledge to capture the robot's partial knowledge on its cost functions, and to balance \emph{exploration} of the unknown costs and \emph{exploitation} of the known portion of the costs.

\domainref{Mobile Sensor Team Problem}
The \emph{Mobile Sensor Team (MST)} problem is similar to the target tracking problem with the difference that agents are capable of moving autonomously within the environment and that time is modeled explicitly as a discrete sequence of time steps. 
In an MST, agents are placed on a grid. For an agent $a_i$, $\varf{cur\_pos}_i$ denotes the agent's current position;  $\varf{SR}_i$ denotes the agent's perception \emph{sensing range}, which determines the coverage range within which an agent can detect targets; $\varf{MR}_i$ denotes the agent's mobility range, which defines the maximum distance that the agent can move within a single time step; and $\varf{cred}_i$ denotes the agent's credibility, which reflects the likelihood of the correctness of the detected targets. The targets are defined implicitly through an \emph{environmental requirement (ER)} function, which defines, for each point in the space, the minimum \emph{joint credibility} value (the sum of the credibility variables) required for that point to be sensed. In such a representation, targets are points $p$ with $ER(p) > 0$. Given a set of agents $\varf{SR}_p$ whose sensing range covers a target $p$, the \emph{remaining coverage requirement} of $p$ is the environmental requirement diminished by the joint credibility of the agents currently covering $p$: $\varf{Cur\_REQ}(p) = \max\{ 0, ER(p) \ominus \sum_{a_i \in \varf{SR}_p} \varf{cred}_i\}$, where $\ominus : \mathbb{R} \times \mathbb{R} \to \mathbb{R}$ is an operator that defines how the environmental requirement decreases by the joint credibility.  The goal of the agents is to find positions that minimize the values of $\varf{Cur\_REQ}$ for all targets.

MST problems are modeled through a subclass of Dynamic DCOPs, named \emph{DCOP\_MST} \cite{zivan:09,yedidsion:14,yedidsion:14b}. Each agent $a_i$ controls one variable $x_i$ representing its position, and whose domain contains all locations within $\varf{MR}_i$ of $\varf{cur\_pos}_i$. Thus, the domains are updated each time the agent moves.\footnote{An alternative representation is that of modeling the domain of each variable as the entire grid, and to constrain the agent variable, at each agent move, in order to hide those points lying outside $\varf{MR}_i$.} The constraint $C_p$ of a target $p$ involves exclusively those 	agents $a_i$ whose variable's domain includes a location within $\varf{SR}_i$ of $p$.
Thus, at each time step, both domains and constraints may change. As a consequence, the constraint graph changes as well -- the neighbors of each agent have to be updated at each time step. Finally, in a DCOP\_MST two agents are neighbors if their sensing areas overlap.

\domainref{Sensor Sleep Scheduling Problem}
Wireless sensor nodes are equipped with a radio, which can be used to communicate with neighboring nodes, and a limited power source. These sensor nodes are often deployed in inaccessible terrains, thus, replacing their power sources may not be possible. The wireless sensor sleeping scheduling problem aims at switching on/off a particular sensor node component (such as the sensor or the radio) for a certain period of time, so to ensure power conservation, maximizing the lifetime of the sensor network. 

\citeA{chachra:06} proposed a DCOP model for this problem, where each sensor is an agent whose variables denote its status (on or off) for each time step. Hard constraints are employed to enforced that if a sensor is on, then all its neighbors should be off, and that sensors cannot stay on for two consecutive time steps. The overall objective is to minimize the delay induced in the network. 

A similar problem is solved by \citeA{stranders:09}, where sensors are also able to harvest energy from the environment (e.g. using a photo-voltaic cell or vibration-harvesting microgenerators). In such a context, the goal is to find a schedule that maximizes the probability of detecting events while maintaining energy-neutral operations (that is, exhibit an indefinite lifetime for each of the agents).

\subsection{Service-Oriented Computing Problems}
The service-oriented computing paradigm is one that relies on sharing resources over a network, focusing on maximizing the effectiveness of the shared resources, which are used by multiple applications. Efficient solutions with optimal use of resources are crucial in this paradigm and have a wide industrial impact \cite{moreno:13}. The distributed nature of the resources and the privacy concerns arising when different clouds are involved in the deployment, makes DCOP appealing to solve a range of problems in this paradigm \cite{mejias:10}.

\domainref{Application Component Placement Problems}
An \emph{Application Component Placement (ACP)} problem is defined over a network of servers offering storage devices with various capabilities, and component-based application with requirements for processing, communication, and storage \cite{jin:11,li:14}. The ACP problem is a problem of deciding which server to assign to each application component. The component-based application is described by a set of characteristics that establish their requirements in terms of hardware (e.g.,~CPU speed, storage capacity) as well as constraints between components of the same application (e.g.,~minimum bandwidth, secure communication channel requirement). When the APC involves deployment on multiple clouds data privacy must be preserved. Additionally, in cloud environments computing resources are shared by many applications and the infrastructure is dynamically changing, making centralized solutions unfeasible. 

\citeA{jin:11} proposed a DCOP model for the ACP problem where servers bid for a component to host, with an emphasis that is proportional to the affinity of the server characteristics and the component hardware and software requirements. Each server is modeled by an agent, which controls a decision variable representing the server bids. Thus, the domain of each variable is the set of possible components that may be deployed on the server. Unary functions express utility for each component. Hard constraints are employed to ensure that each component is deployed exactly on a single server, and that two components are placed between servers satisfying the required communication bandwidth. The objective is to find a feasible assignment of component to servers that maximizes the utilities.

\domainref{Server Allocation Problems}
Services-oriented middleware networks are composed of entities connected within a physical network.
These entities can both provide and require multiple services. 
In turn, each service can be provided by multiple servers and can serve multiple clients. A service request from a given client, takes into account various \emph{Quality of Service (QoS)} parameters (e.g.,~service response time, service completion time). When a client generates a service request, it can be satisfied by any of the servers offering such requests. The server allocation problem is a problem of selecting servers to allocate services, ensuring maximum social welfare, while meeting the QoS requirements of all clients.

\citeA{choudhury:14} presented a DCOP model for this problem where agents correspond to network entities, variables correspond to services, and their values are either 1, if the associated agent is willing to provide/forward the service, or 0, otherwise. Clients' service requests are mapped to servers' service offers, accounting for the delays that occur when traversing between intermediate nodes using a routing multicast protocol \cite{yan:13}. Moreover, in order to provide a service, all requested QoS requirements need to be satisfied. The utility associated to each variable is the combination of the utility for such a node when acting as a service consumer, a service provider, and a service forwarder, and depends on several parameters, such as available GPU cycles, battery power, memory, and bandwidth. The problem may change dynamically when a new request is made, or when a new service is offered or released, and as such can be modeled as a Dynamic DCOP.

\subsection{Smart Grid and Smart Homes Problems}
\label{sec:app:smartgrid}
The smart grid is a vision of the future electricity grid (also called power network) that uses data analytics and decision making to improve the efficiency and reliability of energy production and distribution. The development of smart grids poses several challenges: 
\iOne How to deal with the increasing power network utilization due to growth of loads, such as \emph{electric vehicles (EVs)} and heat pumps; 
\iTwo How to efficiently integrate a diverse range of energy sources, including renewable generators, into the power network; and 
\iThree How to deal with the uncertainty in the equipment as well as in the participation of consumers through demand-side technologies. 
Due to the distributed and dynamic nature of loads and generators participating in the power network, agent-based decentralized autonomous control of smaller distributed microgrids is a very compelling solution \cite{davidson:09,ramchurn:12}. In particular, several agent-based decentralized optimization solutions have been explored to deliver this vision \cite{kumar:09,matsui:11,miller:12,jain:12}. The following is a list of the most prominent DCOP approaches for smart grid and smart homes applications.

\domainref{Economic Dispatch Problems}
\emph{Economic Dispatch (ED)} is the problem of coordinating the various settings of the power generators in order to meet the power loads with the lowest cost possible, while  satisfying the physical power network constraints \cite{wood2012power}. 
Researchers have cast this problem as a DCOP \cite{miller:12,jain:12,gupta:13a,athanasiadis:13}, considering a network of nodes (agents), each of which relays power to other nodes, but can also contain a combination of generators and loads. Generators are distributed across nodes, and are represented through variables whose domain describe a certain set of discrete power outputs. The power lines connecting nodes of the networks are also associated to DCOP variables, each of which has a thermal capacity describing the maximum power that it can safely carry. The DCOP objective function describes a particular optimization criteria, such as minimizing the carbon emissions of generators within the network, as well as the imposed load and network constraints. In particular, the constraints ensure that the overall demand and supply are in balance and that the thermal capacity constraints of the power lines are satisfied.

\rev{
A version of the ED problem with a planning horizon has been proposed by \citeA{fioretto:AAMAS-17b}. This formalization extends the ED model described above by capturing the physical restrictions of transmission lines, power loads, and power generators arising when deploying a sequence of solutions (set-points for generators and loads) over time.
In particular, it models the maximum incremental power that can be supplied or reduced in one time step to each power generator, which depends on the mechanical characteristics of the generators. This problem has been cast as a Dynamic DCOP, where, in addition to the variables and constraints of the classic ED problem, a set of constraints is introduced between each two consecutive time steps to limit the maximum variation of the generators' output.
}

\domainref{Power Supply Restoration Problems}
After (multiple) line failures, a power network must be reconfigured to ensure restoration of power supply. A power network distribution is a network of power lines connected by \emph{switching devices (SDs)} and fed by \emph{circuit breakers (CBs)}. SDs are analogous to sinks (transformer stations), while CBs are analogous to power sources. Both of these devices can operate in two states: open or closed. Closed SDs consume some power and forward the rest of it on other lines. Open SDs stop power flow. CBs feed the network when they are closed. The configuration of the devices' state is such that energy flow traversing CBs takes the form of a (\emph{feeder}) \emph{tree}, and that no SD is powered by more than a single power line. Flow conservation and transmission line capacity constraints must be enforced. The power supply restoration problem is the problem of finding a configuration that ensures power restoration for the maximum number of sinks affected by the line failures.

Researchers have proposed a DCOP formulation for this problem \cite{kumar:09,agrawal:15}. In such a framework, each node of the distribution network is controlled by an agent that owns all variables and constraints corresponding to that node. Two DCOP variables are associated to each network node: A load variable and a direction variable. Load variables model the amount of incoming flow for sink nodes, and the number of sinks fed for power source nodes. Direction variables model all the possibilities of feeding a node, as the set of possible configurations in which its neighboring nodes can forward power to it. The acyclicity of the power flow and the flow conservation are modeled as constraints. The former restricts the power path to be a tree as well as defines the optimization criterion. The latter enforces Kirchhoff's law, that the amount of incoming power flow to the node $i$ must equal the sum of power consumed at $i$ and the amount of power forwarded to other nodes.

\domainref{Microgrid Islanding Problems}
A microgrid islanding problem is the problem of creating islands (i.e.,~clusters of generator units and loads able to operate without external energy supply) in response to major power outages and blackouts. 
\citeA{gupta:13b} formalized this problem as a DCOP where agents represent nodes in the network and each agent has its own power generation and power consumption capabilities. The variables of the DCOP represent the amount of power that an agent generates and consumes, as well as transmission line flows and the switch status between network nodes. The flow variables are constrained by their maximum transmission line capacities, while switches are modeled as binary variables that can be turned on or off. Flow conservation are modeled as constraints to enforce Kirchhoff's law. The goal is to find a switching configuration that minimizes the unserved load of the system.

\domainref{Prosumer Energy Trading Problems}
In its more general form, a smart grid is populated by prosumers capable of both generating and consuming resources. The prosumer energy trading problem aims at setting market-based prices for prosumers to directly trade energy over the smart grid, while taking account of the power network  constraints. This problem has been cast as an optimization problem, called the energy allocation problem, where, given a graph with nodes representing prosumers and edges describing transmission lines connecting adjacent prosumers, the goal is to find an allocation that maximizes the benefits of all the prosumers while satisfying the capacity constraints of the power network.

\citeA{cerquides:15} proposed a DCOP formalization for this problem, where each prosumer is modeled as an agent. Variables are associated with edges of the energy trading network $(i,j)$ and describe the number of units of energy that prosumer $i$ sells to/buys from prosumer $j$. Thus, two variables $(i,j)$ and $(j,i)$ are associated to each edge of the network. For each prosumer, an energy balance constraint models the utility of a given instantiation of its offers as the sum of the offers associated to the energy traded with each of its neighbor. Line capacity and flow conservation constraints ensure that the energy traded along the transmission lines is within their maximum capacity and is consistent with Kirchhoff's law.

\domainref{Smart Building Devices Scheduling Problems}
\rev{
A \emph{smart building} is a residential or commercial building that is partially automated through the introduction of smart devices (e.g.,~smart thermostats, circulator heating, washing machines). Additionally, a range of smart plugs allow users to control remotely the activity of the devices connected to them. Therefore, smart device scheduling can be executed by householders without the control of a centralized authority. The distributed nature of smart devices within a smart building, and of smart buildings within a neighborhood, as well as data privacy concerns, make this domain suitable to DCOP solutions. 

Within a smart building, the \emph{Smart Environment Configuration Problem (SECP)} proposed by \citeA{rust:16} is the problem of coordinating several smart devices (light bulbs, roller shutters, luminosity sensors, etc.) whose actions affect the building environment, with the goal of reaching a desired goal state (e.g.,~a given luminosity level for a room). This resource allocation problem is cast to a DCOP in which devices are described via DCOP agents, each of which controls a single variable describing the devices' action. The domain of the variables model the possible device's actions. Finally, the impact of the device's action onto the building environment is captured by a set of soft constraints. The goal is that of finding an assignment of actions for the building's devices that satisfies the given goals  while minimizing some cost function (e.g.,~the energy consumption of each device).

Within a smart city, the \emph{Smart Home Device Scheduling Problem (SHDS)} proposed by \citeA{fioretto:AAMAS-17a} is the problem of coordinating the schedule of several smart homes' devices so to minimize the aggregated energy peaks as well as to minimize the users' energy bill cost, when adopting a real-time energy price schema. 
In the SHDS problem, each smart home controls a set of smart sensors and smart devices. Within each smart home, users establish scheduling rules defining goal states upon certain home's properties (e.g.,~reaching a certain room temperature by a given time of the day, or charging the battery of an electric vehicle of at least some amount during the night). 
Each device's action has an associated energy consumption and a cost (which depends on the time of the day in which the action is executed and on its required power). Each user in a smart home attempts to find a schedule for its smart devices that satisfies its scheduling rules while minimizing the energy cost. Additionally, multiple smart homes coordinate their devices' schedules so to minimize the daily energy peak consumption. 
Fioretto et.~al map the SHDS problem as a DCOP where each agent models a smart home. Agents control multiple variables, each representing a smart device (a sensor or an actuator). Similar to the SECP problem, variables' domains model the possible actions of a device, and constraints capture the actions' costs, energy consumption, and user preferences. Additionally, a set of hard constraints is used to model the user's temporal goals. 
A dataset for the SHDS problem has been recently released \cite{fioretto:JAAMAS-17}. Finally, \citeA{tabakhi:CP-17} investigate how to elicit preferences associated to the users' temporal goals in an extension of the SHDS problem.
}

\subsection{Supply Chain Management Problems}
The management of large businesses involves the management of the flow of goods from suppliers to customers. This flow of goods is called a supply chain. Supply chains have to be carefully managed to ensure that a sufficient quantity of raw material is available at factories for production and a sufficient quantity of processed goods is available at stores for consumers to buy. Additionally, since goods can be purchased from different producers and sold to different consumers, there is also the need to consider how much to buy/sell the goods and who to buy/sell the goods to. In such an environment, information, decision making and control are inherently decentralized. 

\domainref{Supply Chain Formation Problems}
A supply chain formation problem is the process of determining the participants in a supply chain, who will exchange what, with whom, and the terms of the exchange. Several DCOP-based approaches for this problem have been proposed in the literature \cite{gaudreault:09,penya:12a,penya:12b,winsper:13,penya:14}. They rely on the notion of a \emph{Task Dependency Network (TDN)}, a graph-based representation to capture dependencies among production processes, introduced by \citeA{walsh:00}. 
A TDN is a bipartite directed acyclic graph representing exchanges, where nodes in the graph correspond to producers and consumers and edges in the graph correspond to feasible exchanges of goods between the producers and consumers. A path from potential producers and consumers defines a feasible supply chain configuration and the goal is to find the feasible configuration that optimizes a particular cost function. A DCOP encoding for this problem models producers and consumers as agents, each of which controls a variable describing the agent's decision regarding whom to buy from/sell to as well as its associated quantities and costs. Cost functions are associated with each edge of the TDN encoding the willingness of the producer and consumer to trade with each other. 

A dynamic version of the above formalization has been investigated by \citeA{chli:15}. This model allows for the entry and departure of producers and consumers as well as changes in properties of the problem (e.g.,~prices of goods, production capacity of producers, and consumption requirements of consumers).

\subsection{Traffic Flow Control Problems}
A challenge for the increase of transportation demand is to enforce traffic flow control using the existing infrastructure, such as traffic lights, loop detectors, and cameras. Coordinating the actions of these individual devices aims at smoothing the traffic flow at the network level. Such coordinated actions often generates coherent traffic control plans faster and more accurately compared to those of a human traffic operator \cite{van:09}. Due to the distributed nature of such devices, multi-agent solutions are particularly suitable for this class of problems.

\domainref{Traffic Light Synchronization Problem}
This problem is the problem of finding a synchronization schema for the traffic lights in adjacent intersections that creates green waves, which are waves of vehicles that are traveling at a given speed and are able to cross multiple intersections without stopping at red lights. 

\citeA{deoliveira:05} and \citeA{junges:08} model this problem as a DCOP, where agents represent traffic lights, each controlling one variable that models the coordination direction for the associated traffic signal. Thus, the domain of the variables is given by two possible directions of coordination (north-south/south-north, east-west/west-east). Conflicts that arise when two neighboring traffic signals choose different directions are modeled as constraints. Cost functions are defined to model the number of incoming vehicles in a junction, and the costs are influenced by whether two adjacent agents agree on a direction of coordination or not. The goal is to minimize the global cost. Due to the dynamic nature of the problem, the proposed algorithms fall under the umbrella of Dynamic DCOP algorithms.

\citeA{pham:13b} and \citeA{brys:14} proposed a Probabilistic DCOP-based approach with partial agent knowledge to solve the traffic light synchronization problem in the context where agents have partial information about their cost functions. In particular, the authors argue that traffic patterns may vary during time and, thus, the agents should learn them to update their cost functions. In this context, agents are given a limited amount of time to explore different signal duration intervals, whose associated costs are learned by evaluating the average travel time of the first 100 cars traveling across the agent's traffic light.


\section{Analyses and Perspectives on DCOPs}
\label{sec:analysis}

DCOPs have emerged as a popular formalism for distributed reasoning and coordination in multi-agent systems. 
It provides an elegant modeling framework, which offers flexibility in both the agents' reasoning and coordination strategies. 
Its ability to support the notions of \emph{preferences} and \emph{constraints} makes it suitable to model a variety of multi-agent optimization problems. 
Preferences are a central concept of decision making and arise in a multitude of contexts in multi-agent systems. They are fundamental for the analysis of human choice behavior, and allow agents to express their inclinations through specific actions and behaviors. 
Constraints have been long studied in centralized systems \cite{rossi:06} and have been proved especially practical and efficient for modeling and solving resource allocation and scheduling problems. They are naturally handled within DCOPs and offer a flexible and effective mean to model a variety of complex problems. 
In addition, DCOPs support several aspects that are crucial in multi-agent systems, such as agent privacy, autonomy in reasoning, and cooperation. 

The classical DCOP notion is unable to capture important aspects of the problem related with the environment characteristics, such as partial observability, environment evolution, and uncertainty. Therefore, several DCOP model extensions have been recently presented. Each DCOP model imposes distinctive algorithmic requirements, which concern both the agent's reasoning and cooperation aspects. These requirements, in turn, are strictly related to the characteristics of the problem domain. Due to the performance variability imposed by such requirements, an appropriate selection of the DCOP model and algorithm is essential to obtain desirable performances in realistic application domains.
%

\subsection{Comparative Overview of DCOP Models}

\emph{Classical DCOPs} can be used to  represent a wide range of MAS applications where agents in a team need to work cooperatively to achieve a single goal in a static, deterministic, and fully observable environment. Exploring the domain structural properties, as well as understanding the requirements of the problem designer, is crucial to design and apply effective DCOP algorithms. A discussion on how to choose an appropriate algorithm based on the characteristic of the application at hand is provided in Section~\ref{sec:tradeoffs}.

\emph{Asymmetric DCOPs} are suitable when constrained agents \rev{incur different costs} for a joint action, which arise especially in scenarios where privacy is a particular concern, where agents cannot reveal to the other agents the \rev{costs} associated to their putative actions. 
Examples of problems that can be suitably modeled as Asymmetric DCOPs are resource allocation problems where agents incur in distinct costs from using the same resource, and where their preferences and constraints regarding usage time slots and durations are expected to be different. 
Asymmetric DCOPs are particularly attractive to model those domains that can be represented as \emph{graphical games} \cite{kearns:01}, and where constraint reasoning could be actively used to exploit the problem structure. In a graphical game, the costs of each agent are affected exclusively by its neighboring agents. 
It is important to note that, even though the Asymmetric DCOP model bears similarities with many game-theoretic approaches, these two models are fundamentally different. While game-theoretic agents are self-interested, and their non-cooperative actions lead to a desirable global target, Asymmetric DCOP agents are cooperative and seek to \rev{minimize the global cost} even at the expense of \rev{larger local costs}.

\emph{Multi-Objective DCOPs} are tailored to represent those classes of distributed problems that cannot be modeled with a single optimization function. The observations above on classical DCOPs also hold for Multi-Objective DCOPs and, in addition, agents cooperate to optimize multiple objectives. 
Numerous MAS applications fall in the category of multi-objective optimization. One example is that of \emph{disaster management and coordination problems}, where agents coordinate to effectively respond to an emergency scenario (see Section~\ref{sec:app:disaster}). 
Due to memory requirements, which are proportional to the number of objectives and the size of the Pareto set, incomplete strategies seem particularly promising for this research area. 

\emph{Dynamic DCOPs} capture the dynamic behavior of the evolving environment in which agents act.
Dynamic environments play a fundamental role in real-wold MAS applications. Virtually all complex MAS applications involve dynamic situations,  which may restructure the network topology due to agent movements, or bring additional information to the problem being solved. For example, in a search and rescue operation during \emph{disaster management}, as the environment evolves over time, new information becomes available about civilians to be rescued and new agencies may arrive at any time to help conduct rescue operations (see Section~\ref{sec:app:disaster}). In a \emph{smart grid domain}, real-time pricing is commonly enforced. Thus, agent preferences need to be adapted over time, while energy costs are updated (see Section~\ref{sec:app:smartgrid}).
Dynamic DCOPs are therefore a modern area that presents an exciting field for groundbreaking research.

\emph{Probabilistic DCOPs} extend the classical DCOP model to include the capability of handling uncertain events, allowing DCOP agents to handle a wider range of applications. In particular, Probabilistic DCOPs are suitable to capture those applications characterized by a static environment evolution with exogenous uncertain events (e.g.,~when the actions of agents on the environment can have different outcomes, based on external, uncertain factors) and, yet, agents have total knowledge of their own actions and of the observable environment and act in a fully cooperative context.
The domain of \emph{multi-agent task planning and scheduling} encompasses diverse problems that require complex models and robust solutions under uncertainty (see Section~\ref{sec:app:scheduling}). 
The Probabilistic DCOP model for agents with partial knowledge is suitable to model those applications where agents have no prior knowledge of how the environment reacts to some of the actions. In such a model, the agents are aware of their own actions, which are performed deterministically. However, there is uncertainty in the \rev{cost} associated to such actions, which is influenced by uncertain events that can be discovered over time. Thus, a common approach in such cases is to resort to \emph{sampling} strategies, in order to obtain simple approximations of the probability distributions -- in the form of sample realizations of the probabilistic \rev{costs}. 
Due to the uncertainty arising in such problems, it is especially appealing to recur to solutions that adopt approaches to decision making under uncertainty, such as minimax, maximin, and regret-based methods.

As outlined in Table \ref{tab:dcop_models}, more research effort is needed to solve DCOP models in which agents act in a combined uncertain and dynamic environment. This is by far the most realistic setting for teams of agents acting within a MAS. 
More generally, a coordination strategy that can adapt to the situation where the environment or network problem is evolving dynamically and rapidly and where several scenarios require different approaches to coordination, has not yet been studied.

Finally, \emph{Quantified DCOPs} model adversarial agents, which are common in many MAS applications. In a Quantified DCOP, universally quantified variables can be considered as the choice of the nature of an adversary. Quantified DCOPs can thus formalize problems where a team of agents seeks to limit the effect of the adversarial agents, as well as problems associated to planning under uncertainty. Examples of relevant applications include 
\emph{distributed surveillance planning problems}, where sensors in a network need to coordinate their surveillance areas to detect intruders, whose positions are unknown and who are trying to avoid detection. Thus, the sensors try to find a robust plan that can handle different intruding scenarios (see Section~\ref{sec:app:sensnet}).

\begin{table}[t]
	\renewcommand{\arraystretch}{1.00}
	{
	   \begin{tabular}{| r l | }
			\hline
			{\sc \textbf{DCOP Model}} & {\sc \textbf{Complexity}} \\
			\hline			\hline
			Classical  & NP-hard \\
			Asymmetric & NP-hard \\
			Multi-objective & NP-hard \\
			Dynamic & NP-hard \\
			Probabilistic & PSPACE-hard \\
			Quantified & PSPACE-hard\\
			\hline
	  \end{tabular}
	  }
	\caption{Complexity of the DCOP Models \label{tab:dcop_complexity}}
\end{table}

\rev{In terms of complexity, solving classical, Asymmetric, Multi-Objective, and Dynamic DCOPs optimally is NP-hard. 
In contrast, the Probabilistic DCOP extension to include the capability of handling uncertain events and the Quantified DCOP extension to model adversarial agents comes at a cost of increasing complexity. Both Probabilistic DCOPs and Quantified DCOPs are PSPACE-hard. 
Table~\ref{tab:dcop_complexity} summarizes the complexity of the DCOP models discussed in this survey.}

\subsection{Algorithms and Theoretical Analysis}
Despite the fact that classical DCOPs have reached a sufficient level of maturity from the algorithmic perspective, most of the other proposed formalisms fall short on both algorithmic and theoretical foundations. The proposed algorithms mostly extend classical DCOP algorithms and, therefore, they result in similar performance. Investigating strategies that are based on different backgrounds could help propel the evolution of this area. 

\smallskip\noindent\textbf{Inter-disciplinary Research:}
In particular, further investigations on relating DCOPs to the areas of game theory and decision theory are necessary. 
Similar to the work of \citeA{chapman:08}, where the authors study relationship between DCOPs and potential games, 
analyzing the relationship of DCOPs to auction mechanisms could shed light on how to effectively address coordination and reasoning strategies in DCOP with partially cooperative agents. 
Another direction is to relate DCOP with machine learning techniques. For instance, as outlined by \citeA{kumar:11} and \citeA{ghoshKV15}, one can use inference-based algorithms, such as expectation-maximization, and convex optimization machinery to develop efficient message-passing algorithms for solving large DCOPs.
Additionally, merging insights from decision theory, such as handling partial observability, with the inherent DCOP ability of naturally exploiting problem structure could result in improved performance and/or refined models \cite{nair:05,fioretto:AAMAS-16a,fioretto:AAMAS-17c}. 

\smallskip\noindent\textbf{Anytime Mechanisms:}
Due to the complexity of the DCOP models, the study of incomplete approaches to solve large DCOPs whose agents may act in a dynamic and/or uncertain environment seems particularly suitable. Within the current incomplete methods proposed, a considerable effort has been employed in developing \emph{anytime} algorithms \cite{zilberstein:96}. 
In addition to the anytime mechanism, which constrains the problem resolution within a particular time requirement, \emph{any-space} algorithms have been proposed to limit the amount of space needed to an agent during problem resolution \cite{petcu:07,yeoh:09a,yeoh:11,gutierrez:11}.
Similar to these mechanisms, we see an urge for investigating algorithms that allow problem resolution within any particular network load restriction. For instance, in a congested network agents might need to exchange smaller messages, or to communicate less frequently and/or with a restricted set of neighbors. Situations like these may arise in problems where a multitude of interconnected systems share the same medium, and thus limited bandwidth and interference may cause unsuitable delays. This context may be exacerbated with the advent of concepts such as the Internet of the Things, which expects to see million of connected devices, possibly sharing several mediums \cite{atzori:10,chen:12,miorandi:12}. 
Thus, orthogonal to the direction pursed by anytime and any-space algorithms, we envision the development of \emph{any-communication} procedures. 

\smallskip\noindent\textbf{Coordination and Cooperation Study:}
Another open question is related to agent coordination. It has been observed that simple coordination strategies give good results in extremes (high or low) agent workload environments. However, at intermediate workload levels, such strategies lose their effectiveness, and more complex coordination strategies are necessary \cite{lesser:04,zhang:13}. For example, \citeA{zhang:13} study agent learning, where coordination is driven by a dynamic decomposition of a DCOP, each solving smaller independent subproblems. Their proposal effectively produced near-optimal agent policies for learning and significantly reduced the amount of communication. 
These empirical observations suggest the existence of phase transition behaviors occurring at the level of agent coordination. A formal understanding of these phenomena, which looks at the environment, agents' local reasoning strategies, their assumptions on other agents states, as well as team coordination, could help researchers to understand the inherent complexity of coordination problems. Such a formal framework could be useful to build new and more efficient coordination strategies for a wide variety of multi-agent applications, perhaps resembling the way studies of phase transitions of NP-hard problems \cite{monasson:99} led to the understanding of problem complexity and creation of effective heuristics and search strategies.

\subsection{Evaluation Metrics and Benchmarks}
\smallskip\noindent\textbf{DCOP Resolution Assumptions:}
Modeling many real-world problems as DCOPs often require each agent to solve large complex subproblems, each requiring many variables and constraints. A limitation of most DCOP algorithms is the assumption that each agent controls exactly one variable, and that all constraints are binary. Such assumptions simplify the algorithm organization and presentation. To operate under this assumption, 
\rev{reformulation techniques are commonly adopted to transform a general DCOP into one where each agent controls exclusively one variable. 
There are two commonly used reformulation techniques \cite{burke:06,yokoo:01}: \emph{(i)}~\emph{Compilation} (or \emph{complex variables}) where each agent creates a new \emph{pseudo-variable} whose domain is the Cartesian product of the domains of all variables of the agent; and \emph{(ii)}~\emph{Decomposition} (or \emph{virtual agents}) where each agent creates a \emph{pseudo-agent} for each of its variables. 
While both techniques are relatively simple, they can be inefficient. In compilation, the memory requirement for each agent grows exponentially with the number of variables that it controls. In decomposition, the DCOP algorithms will treat two pseudo-agents as independent entities, resulting in unnecessary computation and communication costs.} 
A more realistic view is to allow each agent to solve its local subproblem (in a centralized fashion) since it is independent of the subproblems of other agents. The agent's subproblem resolution can then explore techniques from centralized reasoning, such as constraint optimization problems, linear programming, and graphical models. 
One can even exploit novel hardware platforms, such as \emph{Graphical Processing Units (GPUs)} to parallelize such solvers \cite{fioretto:Constraint-17,fioretto:CP-16,bistaffa:14}. 

\rev{Researchers have proposed re-designed versions of existing algorithms to be able to handle multi-variables agents problems \cite{davin:06,khanna:09,portway:10}. Additionally, recently, \citeA{grinshpoun:15} and \citeA{fioretto:AAAI-16} have proposed general DCOP problem decompositions which are able to solve problems where agents control multiple variables.}

\smallskip\noindent\textbf{DCOP Modeling Language:}
Despite the wide applicability of the DCOP model, unfortunately, there is no general language being used to formally specify a DCOP. While there are several DCOP simulators that include implementations of various  DCOP algorithms using a common language specification \cite{sultanik:07,leaute:09b,wahbi:11}, by and large, most stand-alone algorithms specify DCOPs in an ad-hoc manner. As a result, it is often inconvenient to experimentally compare different algorithms.
More importantly, the great majority of such languages requires constraints values to be specified explicitly. Such requirement makes unpractical to convert problems which are naturally defined as mathematical optimization problems (such as Mixed Integer Programs) into an explicit form which specifies the utilities for each value combination of the variables in the scope of the problem constraints.
The adoption of  a common distributed constraint modeling language, which allows one to express constraints as standard algebraic or logic expressions, may be beneficial for developing standard benchmarks. 
Additionally, it would provide a tool for researchers outside the AI communities to model and test the applicability of new problems, extending the applicability of DCOPs to new areas.
For instance, within the constraint programming community, the adoption of the MiniZinc language \cite{minizinc} to model CSPs and COPs has gained wide traction, and it is becoming a modeling choice even outside the constraint programming community. 

\smallskip\noindent\textbf{DCOP Simulators:}
\rev{
DCOP simulators are multi-agent software tools that simulate the execution of DCOP algorithms. They are a useful resource for DCOP researchers and practitioners for evaluating different models and algorithms. There are several MAS simulators which have been adopted by researchers to develop and compare DCOP algorithms:
\bitemize
\item \emph{DisChoco} \cite{wahbi:11} is a Java open-source framework for simulating DCOP algorithms where agents are 
executed asynchronously (each in a separate execution thread) and communication is implemented via messages passing.  
The simulator includes some problem generators and the possibility to take into account  message loss, message corruption, and message delay. It supports models in which agents control multiple variables and problems with n-ary constraints. Additionally, it has the ability to be executed in a distributed environment. 
 
\item \emph{Frodo} \cite{leaute:09b} is a Java open-source framework for DCOP which implements several complete and incomplete classic DCOP algorithms, including DPOP and some of it variants, Max-Sum, MGM, and DSA. It can be executed either in a synchronous or in an asynchronous mode. It provides a Graphical User Interface to visualize the problem structure and algorithm execution, and it includes some problem generators for benchmarking. It supports problems with n-ary constraints and models in which agents need to control the assignment of several variables. Finally, in addition to simulating DCOP algorithms in a centralized environment, it provides support for executing agents in a distributed environment.

\item \emph{AgentZero} \cite{lutati:14} is a Java open-source framework for various MAS applications. 
It provides distributed runtime environment, which can be used to simulate both synchronous and asynchronous algorithms.
AgentZero offers several high-level APIs which allow to easily handle complex tasks, such as the construction of a pseudo-tree and the evaluation of assignments. Additionally, it provides some built-in performance measures and visualization tools that can be especially useful to debug distributed programs. It supports the notion of message delays, message corruptions, and message losses. While it does not support models in which agents control multiple variables, it can deal with n-ary constraints. 
\eitemize

In addition to using DCOP simulators, researchers have developed and deployed DCOP algorithms using general MAS platforms. In particular, the \emph{Java Agent DEvelopment Framework (JADE)} \cite{bellifemine:07} is a multi-agent platform that is compliant with the FIPA specifications.\footnote{\url{http://www.fipa.org/}} 
It includes a runtime environment in which the agents act, a set of APIs to define the agents' behaviors, and a suite of graphical tools that can be used to monitor the agents' activity. JADE agents' messages are specified by the ACL language, defined by the FIPA international standard for agent interoperability.  JADE allows the deployment of agents on several environments, including mobile and wireless environments and fault-tolerant platforms.

Table~\ref{tab:dcop_sim} provides a summary of the simulators' features and describes whether the simulator 
\emph{execution mode} supports synchronous or asynchronous algorithms; 
whether it provides \emph{visualization} tools;  
the type of implementation of messages passing simulation (i.e.,~using pointers, deep copying the message structure into the receiving agent's  message queue, etc.);
if the framework provides support to simulate/handle message delays, message corruptions, and message losses; 
if it supports DCOP models in which agents control multiple variables and n-ary constraints; 
if the framework provides tools for debugging the algorithms;
the programming language in which the simulator is implemented;
whether it provides documentation;
and whether it provides support for deployment. 
For a comprehensive review of multi-agent platforms, the interested reader is referred to the survey by \citeA{kravari:15}.

The development of DCOP simulators have provided researchers with useful tools to develop new DCOP algorithms and to facilitate comparison of existing DCOP algorithms. However, all the DCOP simulators described above focus on the classical DCOP model, while Dynamic and Probabilistic DCOP simulators have received little attention. Thus, there is a need of developing high-fidelity simulators for Dynamic DCOPs, which simulate how the environment can vary over time, as well as for Probabilistic DCOPs, which simulate the stochastic nature of exogenous events.
}

\begin{table}[t]
\resizebox{1\textwidth}{!}
	{
	   \begin{tabular}{| r l l l l | }
			\hline
			{\sc \textbf{Property}} & {\sc \textbf{DisChoco}} & {\sc \textbf{FRODO}} & {\sc \textbf{AgentZero}} & {\sc \textbf{JADE}} \\
			\hline			\hline
			Execution modes 		& Async		& Sync/Async 	& Sync/Async 	& Synch/Async \\ 
			Visualization 			& No		& Yes			&	Yes			& Yes \\ 
			Message support			& Pointers	& Pointers		& Deep copy		& FIPA messages	\\
			Message delays/corruptions/losses	 & Yes	&	No  &	Yes			& Yes  \\
			Multiple variables per agent support & Yes	&	Yes	& 	No	& N/A   \\
			N-ary constraints support & Yes	&	Yes	& 	Yes	& N/A   \\
			Debugging Tools 		& No		& No 			& Yes 		&	No \\
			Programming Language	& Java		& Java			& Java 		& Java \\
			Documentation 			& No	&	Yes	&   Yes			& Yes \\
			Deployable				& No	&	No  &	No	& Yes \\
			\hline
	  \end{tabular}
	  }
	\caption{\rev{Comparison of DCOP Simulators} \label{tab:dcop_sim}}
\end{table}


\smallskip\noindent\textbf{Evaluation and Metrics:}
Another open question in this research area concerns the definition of a systematic process to evaluate and compare the DCOP algorithms. There are multiple metrics that can be used to measure the runtime of an algorithm, such as the number of \emph{Non-Concurrent Constraint Checks (NCCCs)} \cite{meisels:02}, the simulated runtime \cite{sultanik:07b}, and the number of cycles \cite{lynch:96}. However, there is no consensus on a standard metric. Such issues, combined with absence of a general DCOP language, make it inconvenient to experimentally compare different algorithms. In addition, new proposed algorithms are, in general, evaluated on arbitrary benchmarks, some inherited from the CSP literature, some other from approximations of real-world problems. 
To cope with these issues, it would be useful to develop a benchmark repository, perhaps by taking inspiration from the efforts made by the constraint programming community with CSPLib,\footnote{\url{http://www.csplib.org/}} a library of test problems for constraint solvers, or by the planning community with their international planning competitions.\footnote{\url{http://icaps-conference.org/index.php/Main/Competitions}}


\section{Conclusions}\label{concl}

DCOPs have emerged as a popular formalism for distributed reasoning and coordination in multi-agent systems. Due to its limitation to support complex, real-time, and uncertain environment, researchers have introduced several model extensions to handle dynamic and uncertain environments, as well as different levels of cooperation among the agents.  

While DCOPs' theoretical foundation and algorithmic frameworks have matured significantly over the past decade, their applicability to realistic domains is lagging behind. This survey aims at linking the DCOP theoretical framework and solving strategies with a set of potential applications where its applicability is having or may have a significant impact.

This survey provided an analysis of the recent advances made by the AAMAS community within the DCOP framework and propose a categorization based on agent characteristics, environment properties, and type of teamwork adopted. 
Within the proposed classification, it \iOne~presented a review of the characteristics of the different 
algorithmic solutions; \iTwo~discussed a number of application domains that can be naturally modeled within each DCOP framework; and \iThree~identified  some potential directions for future work with regards to agent coordination, algorithm scalability, modeling languages, and evaluation criteria of  DCOP models and algorithms.

\acks{The authors would like to thank the anonymous reviewers for their valuable comments and suggestions to improve the quality of the paper. They are also thankful to Tiep Le for early discussions on Probabilistic DCOPs. The research in this paper has been partially supported by NSF grants 
0947465, 1345232, 1401639, 1458595, and 1550662. The views and conclusions contained in this document are those of the authors and should not be interpreted as representing the official policies, either expressed or implied, of the sponsoring organizations, agencies, or the U.S. government.
}

\small
\vskip 0.2in
\bibliography{mas,Fioretto}

\begin{thebibliography}{}

\bibitem[\protect\BCAY{Abounadi, Bertsekas,\ \BBA\ Borkar}{Abounadi
  et~al.}{2001}]{abounadi:01}
Abounadi, J., Bertsekas, D., \BBA\ Borkar, V. \BBOP2001\BBCP.
\newblock \BBOQ Learning algorithms for {Markov} decision processes with
  average cost\BBCQ\
\newblock {\Bem SIAM Journal on Control and Optimization}, {\Bem 40\/}(3),
  681--698.

\bibitem[\protect\BCAY{Agrawal, Kumar,\ \BBA\ Varakantham}{Agrawal
  et~al.}{2015}]{agrawal:15}
Agrawal, P., Kumar, A., \BBA\ Varakantham, P. \BBOP2015\BBCP.
\newblock \BBOQ Near-optimal decentralized power supply restoration in smart
  grids\BBCQ\
\newblock In {\Bem Proceedings of the International Conference on Autonomous
  Agents and Multiagent Systems (AAMAS)}, \BPGS\ 1275--1283.

\bibitem[\protect\BCAY{Aji\ \BBA\ McEliece}{Aji\ \BBA\ McEliece}{2000}]{aji:00}
Aji, S.~M.\BBACOMMA\  \BBA\ McEliece, R.~J. \BBOP2000\BBCP.
\newblock \BBOQ The generalized distributive law\BBCQ\
\newblock {\Bem {IEEE} Transactions on Information Theory}, {\Bem 46\/}(2),
  325--343.

\bibitem[\protect\BCAY{Akyildiz, Su, Sankarasubramaniam,\ \BBA\
  Cayirci}{Akyildiz et~al.}{2002}]{akyildiz:02}
Akyildiz, I.~F., Su, W., Sankarasubramaniam, Y., \BBA\ Cayirci, E.
  \BBOP2002\BBCP.
\newblock \BBOQ Wireless sensor networks: a survey\BBCQ\
\newblock {\Bem Computer Networks}, {\Bem 38\/}(4), 393--422.

\bibitem[\protect\BCAY{Amador, Okamoto,\ \BBA\ Zivan}{Amador
  et~al.}{2014}]{amador:14}
Amador, S., Okamoto, S., \BBA\ Zivan, R. \BBOP2014\BBCP.
\newblock \BBOQ Dynamic multi-agent task allocation with spatial and temporal
  constraints\BBCQ\
\newblock In {\Bem Proceedings of the AAAI Conference on Artificial
  Intelligence (AAAI)}, \BPGS\ 1384--1390.

\bibitem[\protect\BCAY{Amato, Chowdhary, Geramifard, Ure,\ \BBA\
  Kochenderfer}{Amato et~al.}{2013}]{amato:13}
Amato, C., Chowdhary, G., Geramifard, A., Ure, N.~K., \BBA\ Kochenderfer, M.~J.
  \BBOP2013\BBCP.
\newblock \BBOQ Decentralized control of partially observable {Markov} decision
  processes\BBCQ\
\newblock In {\Bem Proceedings of the Annual Conference on Decision and Control
  (CDC)}, \BPGS\ 2398--2405.

\bibitem[\protect\BCAY{Amigoni, Castelletti,\ \BBA\ Giuliani}{Amigoni
  et~al.}{2015}]{amigoni:15}
Amigoni, F., Castelletti, A., \BBA\ Giuliani, M. \BBOP2015\BBCP.
\newblock \BBOQ Modeling the management of water resources systems using
  {Multi-Objective DCOPs}\BBCQ\
\newblock In {\Bem Proceedings of the International Conference on Autonomous
  Agents and Multiagent Systems (AAMAS)}, \BPGS\ 821--829.

\bibitem[\protect\BCAY{Apt}{Apt}{2003}]{apt:03}
Apt, K. \BBOP2003\BBCP.
\newblock {\Bem Principles of Constraint Programming}.
\newblock Cambridge University Press.

\bibitem[\protect\BCAY{Athanasiadis, Kockar,\ \BBA\ McArthur}{Athanasiadis
  et~al.}{2013}]{athanasiadis:13}
Athanasiadis, D., Kockar, I., \BBA\ McArthur, S. \BBOP2013\BBCP.
\newblock \BBOQ Distributed constraint optimisation for flexible network
  management\BBCQ\
\newblock In {\Bem Proceedings of the IEEE International Conference on
  Innovative Smart Grid Technologies Europe (ISGT EUROPE)}, \BPGS\ 1--5.

\bibitem[\protect\BCAY{Atlas\ \BBA\ Decker}{Atlas\ \BBA\
  Decker}{2010}]{atlas:10}
Atlas, J.\BBACOMMA\  \BBA\ Decker, K. \BBOP2010\BBCP.
\newblock \BBOQ Coordination for uncertain outcomes using distributed neighbor
  exchange\BBCQ\
\newblock In {\Bem Proceedings of the International Conference on Autonomous
  Agents and Multiagent Systems (AAMAS)}, \BPGS\ 1047--1054.

\bibitem[\protect\BCAY{Atzori, Iera,\ \BBA\ Morabito}{Atzori
  et~al.}{2010}]{atzori:10}
Atzori, L., Iera, A., \BBA\ Morabito, G. \BBOP2010\BBCP.
\newblock \BBOQ The internet of things: A survey\BBCQ\
\newblock {\Bem Computer Networks}, {\Bem 54\/}(15), 2787--2805.

\bibitem[\protect\BCAY{Auer, Cesa-Bianchi,\ \BBA\ Fischer}{Auer
  et~al.}{2002}]{auer:02}
Auer, P., Cesa-Bianchi, N., \BBA\ Fischer, P. \BBOP2002\BBCP.
\newblock \BBOQ Finite-time analysis of the multiarmed bandit problem\BBCQ\
\newblock {\Bem Machine learning}, {\Bem 47\/}(2-3), 235--256.

\bibitem[\protect\BCAY{Baba, Iwasaki, Yokoo, Silaghi, Hirayama,\ \BBA\
  Matsui}{Baba et~al.}{2010}]{baba:10}
Baba, S., Iwasaki, A., Yokoo, M., Silaghi, M.~C., Hirayama, K., \BBA\ Matsui,
  T. \BBOP2010\BBCP.
\newblock \BBOQ Cooperative problem solving against adversary: Quantified
  distributed constraint satisfaction problem\BBCQ\
\newblock In {\Bem Proceedings of the International Conference on Autonomous
  Agents and Multiagent Systems (AAMAS)}, \BPGS\ 781--788.

\bibitem[\protect\BCAY{Baba, Joe, Iwasaki,\ \BBA\ Yokoo}{Baba
  et~al.}{2011}]{baba:11}
Baba, S., Joe, Y., Iwasaki, A., \BBA\ Yokoo, M. \BBOP2011\BBCP.
\newblock \BBOQ Real-time solving of quantified {CSP}s based on monte-carlo
  game tree search\BBCQ\
\newblock In {\Bem Proceedings of the International Joint Conference on
  Artificial Intelligence (IJCAI)}, \BPGS\ 655--661.

\bibitem[\protect\BCAY{Bellifemine, Caire,\ \BBA\ Greenwood}{Bellifemine
  et~al.}{2007}]{bellifemine:07}
Bellifemine, F.~L., Caire, G., \BBA\ Greenwood, D. \BBOP2007\BBCP.
\newblock {\Bem Developing Multi-Agent Systems with JADE}, \lowercase{\BVOL}~7.
\newblock John Wiley \& Sons.

\bibitem[\protect\BCAY{Benders}{Benders}{1962}]{benders:62}
Benders, J.~F. \BBOP1962\BBCP.
\newblock \BBOQ Partitioning procedures for solving mixed-variables programming
  problems\BBCQ\
\newblock {\Bem Numerische Mathematik}, {\Bem 4\/}(1), 238--252.

\bibitem[\protect\BCAY{Benedetti, Lallouet,\ \BBA\ Vautard}{Benedetti
  et~al.}{2008}]{benedetti:08}
Benedetti, M., Lallouet, A., \BBA\ Vautard, J. \BBOP2008\BBCP.
\newblock \BBOQ Quantified constraint optimization\BBCQ\
\newblock In {\Bem Proceedings of the International Conference on Principles
  and Practice of Constraint Programming (CP)}, \BPGS\ 463--477.

\bibitem[\protect\BCAY{Bernstein, Givan, Immerman,\ \BBA\
  Zilberstein}{Bernstein et~al.}{2002}]{bernstein:02}
Bernstein, D.~S., Givan, R., Immerman, N., \BBA\ Zilberstein, S.
  \BBOP2002\BBCP.
\newblock \BBOQ The complexity of decentralized control of {Markov} decision
  processes\BBCQ\
\newblock {\Bem Mathematics of Operations Research}, {\Bem 27\/}(4), 819--840.

\bibitem[\protect\BCAY{Bessiere, Brito, Gutierrez,\ \BBA\ Meseguer}{Bessiere
  et~al.}{2014}]{bessiere:14}
Bessiere, C., Brito, I., Gutierrez, P., \BBA\ Meseguer, P. \BBOP2014\BBCP.
\newblock \BBOQ Global constraints in distributed constraint satisfaction and
  optimization\BBCQ\
\newblock {\Bem Computer Journal}, {\Bem 57\/}(6), 906--923.

\bibitem[\protect\BCAY{Bessiere, Gutierrez,\ \BBA\ Meseguer}{Bessiere
  et~al.}{2012}]{bessiere:12}
Bessiere, C., Gutierrez, P., \BBA\ Meseguer, P. \BBOP2012\BBCP.
\newblock \BBOQ Including soft global constraints in {DCOP}s\BBCQ\
\newblock In {\Bem Proceedings of the International Conference on Principles
  and Practice of Constraint Programming (CP)}, \BPGS\ 175--190.

\bibitem[\protect\BCAY{Billiau, Chang,\ \BBA\ Ghose}{Billiau
  et~al.}{2012a}]{billiau:12a}
Billiau, G., Chang, C.~F., \BBA\ Ghose, A. \BBOP2012a\BBCP.
\newblock \BBOQ {SBDO}: A new robust approach to dynamic distributed constraint
  optimisation\BBCQ\
\newblock In {\Bem Proceedings of the Principles and Practice of Multi-Agent
  Systems (PRIMA)}, \BPGS\ 11--26.

\bibitem[\protect\BCAY{Billiau, Chang, Ghose,\ \BBA\ Miller}{Billiau
  et~al.}{2012b}]{billiau:12b}
Billiau, G., Chang, C.~F., Ghose, A., \BBA\ Miller, A.~A. \BBOP2012b\BBCP.
\newblock \BBOQ Using distributed agents for patient scheduling\BBCQ\
\newblock In {\Bem Proceedings of the Principles and Practice of Multi-Agent
  Systems (PRIMA)}, \BPGS\ 551--560.

\bibitem[\protect\BCAY{Billiau, Chang, Ghose,\ \BBA\ Miller}{Billiau
  et~al.}{2012c}]{billiau:12c}
Billiau, G., Chang, C.~F., Ghose, A., \BBA\ Miller, A. \BBOP2012c\BBCP.
\newblock \BBOQ Support-based distributed optimisation: An approach to
  radiotherapy scheduling\BBCQ\
\newblock In {\Bem Electronic Healthcare}, \BPGS\ 327--334.

\bibitem[\protect\BCAY{Binmore}{Binmore}{1992}]{binmore:92}
Binmore, K. \BBOP1992\BBCP.
\newblock {\Bem Fun and Games, a Text on Game Theory}.
\newblock DC Heath and Company.

\bibitem[\protect\BCAY{Bistaffa, Farinelli,\ \BBA\ Bombieri}{Bistaffa
  et~al.}{2014}]{bistaffa:14}
Bistaffa, F., Farinelli, A., \BBA\ Bombieri, N. \BBOP2014\BBCP.
\newblock \BBOQ Optimising memory management for belief propagation in junction
  trees using {GPGPU}s\BBCQ\
\newblock In {\Bem Proceeding of the International Conference on Parallel and
  Distributed Systems ({ICPADS})}, \BPGS\ 526--533.

\bibitem[\protect\BCAY{Bowring, Tambe,\ \BBA\ Yokoo}{Bowring
  et~al.}{2006}]{bowring:06}
Bowring, E., Tambe, M., \BBA\ Yokoo, M. \BBOP2006\BBCP.
\newblock \BBOQ Multiply-constrained distributed constraint optimization\BBCQ\
\newblock In {\Bem Proceedings of the International Conference on Autonomous
  Agents and Multiagent Systems (AAMAS)}, \BPGS\ 1413--1420.

\bibitem[\protect\BCAY{Brito, Meisels, Meseguer,\ \BBA\ Zivan}{Brito
  et~al.}{2009}]{brito:09}
Brito, I., Meisels, A., Meseguer, P., \BBA\ Zivan, R. \BBOP2009\BBCP.
\newblock \BBOQ Distributed constraint satisfaction with partially known
  constraints\BBCQ\
\newblock {\Bem Constraints}, {\Bem 14\/}(2), 199--234.

\bibitem[\protect\BCAY{Brys, Pham,\ \BBA\ Taylor}{Brys et~al.}{2014}]{brys:14}
Brys, T., Pham, T.~T., \BBA\ Taylor, M.~E. \BBOP2014\BBCP.
\newblock \BBOQ Distributed learning and multi-objectivity in traffic light
  control\BBCQ\
\newblock {\Bem Connection Science}, {\Bem 26\/}(1), 65--83.

\bibitem[\protect\BCAY{Burke\ \BBA\ Brown}{Burke\ \BBA\ Brown}{2006}]{burke:06}
Burke, D.\BBACOMMA\  \BBA\ Brown, K. \BBOP2006\BBCP.
\newblock \BBOQ Efficiently handling complex local problems in distributed
  constraint optimisation\BBCQ\
\newblock In {\Bem Proceedings of the European Conference on Artificial
  Intelligence (ECAI)}, \BPGS\ 701--702.

\bibitem[\protect\BCAY{Carpenter, Dugan, Kopena, Lass, Naik, Nguyen, Sultanik,
  Modi,\ \BBA\ Regli}{Carpenter et~al.}{2007}]{carpenter:07}
Carpenter, C.~J., Dugan, C.~J., Kopena, J.~B., Lass, R.~N., Naik, G., Nguyen,
  D.~N., Sultanik, E., Modi, P.~J., \BBA\ Regli, W.~C. \BBOP2007\BBCP.
\newblock \BBOQ Intelligent systems demonstration: disaster evacuation
  support\BBCQ\
\newblock In {\Bem Proceedings of the AAAI Conference on Artificial
  Intelligence (AAAI)}, \BPGS\ 1964--1965.

\bibitem[\protect\BCAY{Cerquides, Picard,\ \BBA\
  Rodr{\'\i}guez-Aguilar}{Cerquides et~al.}{2015}]{cerquides:15}
Cerquides, J., Picard, G., \BBA\ Rodr{\'\i}guez-Aguilar, J.~A. \BBOP2015\BBCP.
\newblock \BBOQ Designing a marketplace for the trading and distribution of
  energy in the smart grid\BBCQ\
\newblock In {\Bem Proceedings of the International Conference on Autonomous
  Agents and Multiagent Systems (AAMAS)}, \BPGS\ 1285--1293.

\bibitem[\protect\BCAY{Chachra\ \BBA\ Marefat}{Chachra\ \BBA\
  Marefat}{2006}]{chachra:06}
Chachra, S.\BBACOMMA\  \BBA\ Marefat, M. \BBOP2006\BBCP.
\newblock \BBOQ Distributed algorithms for sleep scheduling in wireless sensor
  networks\BBCQ\
\newblock In {\Bem Proceedings of the {IEEE} International Conference on
  Robotics and Automation (ICRA)}, \BPGS\ 3101--3107.

\bibitem[\protect\BCAY{Chapman, Rogers,\ \BBA\ Jennings}{Chapman
  et~al.}{2008}]{chapman:08}
Chapman, A., Rogers, A., \BBA\ Jennings, N. \BBOP2008\BBCP.
\newblock \BBOQ A parameterisation of algorithms for distributed constraint
  optimisation via potential games\BBCQ\
\newblock In {\Bem International Workshop on Distributed Constraint Reasoning
  (DCR)}, \BPGS\ 99--113.

\bibitem[\protect\BCAY{Chechetka\ \BBA\ Sycara}{Chechetka\ \BBA\
  Sycara}{2006}]{chechetka:06}
Chechetka, A.\BBACOMMA\  \BBA\ Sycara, K. \BBOP2006\BBCP.
\newblock \BBOQ No-commitment branch and bound search for distributed
  constraint optimization\BBCQ\
\newblock In {\Bem Proceedings of the International Conference on Autonomous
  Agents and Multiagent Systems (AAMAS)}, \BPGS\ 1427--1429.

\bibitem[\protect\BCAY{Chen}{Chen}{2012}]{chen:12}
Chen, Y.-K. \BBOP2012\BBCP.
\newblock \BBOQ Challenges and opportunities of internet of things\BBCQ\
\newblock In {\Bem Proceedings of the Asia and South Pacific Design Automation
  Conference (ASP-DAC)}, \BPGS\ 383--388.

\bibitem[\protect\BCAY{Chen, Deng,\ \BBA\ Wu}{Chen et~al.}{2017}]{chen:17}
Chen, Z., Deng, Y., \BBA\ Wu, T. \BBOP2017\BBCP.
\newblock \BBOQ An iterative refined max-sum\_ad algorithm via single-side
  value propagation and local search\BBCQ\
\newblock In {\Bem Proceedings of the International Conference on Autonomous
  Agents and Multiagent Systems (AAMAS)}, \BPGS\ 195--202.

\bibitem[\protect\BCAY{Chli\ \BBA\ Winsper}{Chli\ \BBA\
  Winsper}{2015}]{chli:15}
Chli, M.\BBACOMMA\  \BBA\ Winsper, M. \BBOP2015\BBCP.
\newblock \BBOQ Using the {Max-Sum} algorithm for supply chain emergence in
  dynamic multiunit environments\BBCQ\
\newblock {\Bem {IEEE} Transactions of Systems, Man, and Cybernetics: Systems},
  {\Bem 45\/}(3), 422--435.

\bibitem[\protect\BCAY{Choudhury, Dey, Dutta,\ \BBA\ Choudhury}{Choudhury
  et~al.}{2014}]{choudhury:14}
Choudhury, B., Dey, P., Dutta, A., \BBA\ Choudhury, S. \BBOP2014\BBCP.
\newblock \BBOQ A multi-agent based optimised server selection scheme for {SOC}
  in pervasive environment\BBCQ\
\newblock In {\Bem Advances in Practical Applications of Heterogeneous
  Multi-Agent Systems. The PAAMS Collection}, \BPGS\ 50--61.

\bibitem[\protect\BCAY{Choxi\ \BBA\ Modi}{Choxi\ \BBA\ Modi}{2007}]{choxi:07}
Choxi, H.\BBACOMMA\  \BBA\ Modi, P. \BBOP2007\BBCP.
\newblock \BBOQ A distributed constraint optimization approach to wireless
  network optimization\BBCQ\
\newblock In {\Bem Proceedings of the AAAI-07 Workshop on Configuration},
  \BPGS\ 1--8.

\bibitem[\protect\BCAY{Davidson, Dolan, McArthur,\ \BBA\ Ault}{Davidson
  et~al.}{2009}]{davidson:09}
Davidson, E., Dolan, M., McArthur, S., \BBA\ Ault, G. \BBOP2009\BBCP.
\newblock \BBOQ The use of constraint programming for the autonomous management
  of power flows\BBCQ\
\newblock In {\Bem Proceedings of the International Conference on Intelligent
  System Applications to Power Systems (ISAP)}, \BPGS\ 1--7.

\bibitem[\protect\BCAY{Davin\ \BBA\ Modi}{Davin\ \BBA\ Modi}{2006}]{davin:06}
Davin, J.\BBACOMMA\  \BBA\ Modi, P. \BBOP2006\BBCP.
\newblock \BBOQ Hierarchical variable ordering for multiagent agreement
  problems\BBCQ\
\newblock In {\Bem Proceedings of the International Conference on Autonomous
  Agents and Multiagent Systems (AAMAS)}, \BPGS\ 1433--1435.

\bibitem[\protect\BCAY{de~Oliveira, Bazzan,\ \BBA\ Lesser}{de~Oliveira
  et~al.}{2005}]{deoliveira:05}
de~Oliveira, D., Bazzan, A.~L., \BBA\ Lesser, V. \BBOP2005\BBCP.
\newblock \BBOQ Using cooperative mediation to coordinate traffic lights: a
  case study\BBCQ\
\newblock In {\Bem Proceedings of the International Conference on Autonomous
  Agents and Multiagent Systems (AAMAS)}, \BPGS\ 463--470.

\bibitem[\protect\BCAY{Delle~Fave, Stranders, Rogers,\ \BBA\
  Jennings}{Delle~Fave et~al.}{2011}]{dellefave:11}
Delle~Fave, F.~M., Stranders, R., Rogers, A., \BBA\ Jennings, N.~R.
  \BBOP2011\BBCP.
\newblock \BBOQ Bounded decentralised coordination over multiple
  objectives\BBCQ\
\newblock In {\Bem Proceedings of the International Conference on Autonomous
  Agents and Multiagent Systems (AAMAS)}, \BPGS\ 371--378.

\bibitem[\protect\BCAY{Dijkstra}{Dijkstra}{1974}]{dijkstra:74}
Dijkstra, E.~W. \BBOP1974\BBCP.
\newblock \BBOQ Self-stabilization in spite of distributed control\BBCQ\
\newblock {\Bem Communication of the ACM}, {\Bem 17\/}(11), 643--644.

\bibitem[\protect\BCAY{Farinelli, Rogers, Petcu,\ \BBA\ Jennings}{Farinelli
  et~al.}{2008}]{farinelli:08}
Farinelli, A., Rogers, A., Petcu, A., \BBA\ Jennings, N. \BBOP2008\BBCP.
\newblock \BBOQ Decentralised coordination of low-power embedded devices using
  the {Max-Sum} algorithm\BBCQ\
\newblock In {\Bem Proceedings of the International Conference on Autonomous
  Agents and Multiagent Systems (AAMAS)}, \BPGS\ 639--646.

\bibitem[\protect\BCAY{Fioretto, Le, Yeoh, Pontelli,\ \BBA\ Son}{Fioretto
  et~al.}{2014}]{fioretto:14b}
Fioretto, F., Le, T., Yeoh, W., Pontelli, E., \BBA\ Son, T.~C. \BBOP2014\BBCP.
\newblock \BBOQ Improving {DPOP} with branch consistency for solving
  distributed constraint optimization problems\BBCQ\
\newblock In {\Bem Proceedings of the International Conference on Principles
  and Practice of Constraint Programming (CP)}, \BPGS\ 307--323.

\bibitem[\protect\BCAY{Fioretto, Pontelli, Yeoh,\ \BBA\ Dechter}{Fioretto
  et~al.}{2017}]{fioretto:Constraint-17}
Fioretto, F., Pontelli, E., Yeoh, W., \BBA\ Dechter, R. \BBOP2017\BBCP.
\newblock \BBOQ Accelerating exact and approximate inference for (distributed)
  discrete optimization with {GPUs}\BBCQ\
\newblock {\Bem Constraints}.

\bibitem[\protect\BCAY{Fioretto, Yeoh,\ \BBA\ Pontelli}{Fioretto
  et~al.}{2016a}]{fioretto:CP-16}
Fioretto, F., Yeoh, W., \BBA\ Pontelli, E. \BBOP2016a\BBCP.
\newblock \BBOQ A dynamic programming-based {MCMC} framework for solving
  {DCOP}s with {GPU}s\BBCQ\
\newblock In {\Bem Proceedings of Principles and Practice of Constraint
  Programming {(CP)}}, \BPGS\ 813--831.

\bibitem[\protect\BCAY{Fioretto, Yeoh,\ \BBA\ Pontelli}{Fioretto
  et~al.}{2016b}]{fioretto:AAAI-16}
Fioretto, F., Yeoh, W., \BBA\ Pontelli, E. \BBOP2016b\BBCP.
\newblock \BBOQ Multi-variable agents decomposition for {DCOP}s\BBCQ\
\newblock In {\Bem Proceedings of the {AAAI} Conference on Artificial
  Intelligence ({AAAI})}, \BPGS\ 2480--2486.

\bibitem[\protect\BCAY{Fioretto, Yeoh,\ \BBA\ Pontelli}{Fioretto
  et~al.}{2017a}]{fioretto:AAMAS-17a}
Fioretto, F., Yeoh, W., \BBA\ Pontelli, E. \BBOP2017a\BBCP.
\newblock \BBOQ A multiagent system approach to scheduling devices in smart
  homes\BBCQ\
\newblock In {\Bem Proceedings of the International Conference on Autonomous
  Agents and Multiagent Systems {(AAMAS)}}, \BPGS\ 981--989.

\bibitem[\protect\BCAY{Fioretto, Yeoh, Pontelli, Ma,\ \BBA\ Ranade}{Fioretto
  et~al.}{2017b}]{fioretto:AAMAS-17b}
Fioretto, F., Yeoh, W., Pontelli, E., Ma, Y., \BBA\ Ranade, S. \BBOP2017b\BBCP.
\newblock \BBOQ A {DCOP} approach to the economic dispatch with demand
  response\BBCQ\
\newblock In {\Bem Proceedings of the International Conference on Autonomous
  Agents and Multiagent Systems {(AAMAS)}}, \BPGS\ 999--1007.

\bibitem[\protect\BCAY{Garrido\ \BBA\ Sycara}{Garrido\ \BBA\
  Sycara}{1996}]{garrido:96}
Garrido, L.\BBACOMMA\  \BBA\ Sycara, K. \BBOP1996\BBCP.
\newblock \BBOQ Multi-agent meeting scheduling: Preliminary experimental
  results\BBCQ\
\newblock In {\Bem Proceedings of the International Conference on Multiagent
  Systems}, \BPGS\ 95--102.

\bibitem[\protect\BCAY{Gaudreault, Frayret,\ \BBA\ Pesant}{Gaudreault
  et~al.}{2009}]{gaudreault:09}
Gaudreault, J., Frayret, J.-M., \BBA\ Pesant, G. \BBOP2009\BBCP.
\newblock \BBOQ Distributed search for supply chain coordination\BBCQ\
\newblock {\Bem Computers in Industry}, {\Bem 60\/}(6), 441--451.

\bibitem[\protect\BCAY{Geman\ \BBA\ Geman}{Geman\ \BBA\ Geman}{1984}]{geman:84}
Geman, S.\BBACOMMA\  \BBA\ Geman, D. \BBOP1984\BBCP.
\newblock \BBOQ Stochastic relaxation, {Gibbs} distributions, and the
  {Bayesian} restoration of images\BBCQ\
\newblock {\Bem {IEEE} Transactions on Pattern Analysis and Machine
  Intelligence}, {\Bem 6\/}(6), 721--741.

\bibitem[\protect\BCAY{Gershman, Meisels,\ \BBA\ Zivan}{Gershman
  et~al.}{2009}]{gershman:09}
Gershman, A., Meisels, A., \BBA\ Zivan, R. \BBOP2009\BBCP.
\newblock \BBOQ {Asynchronous Forward-Bounding} for distributed {COP}s\BBCQ\
\newblock {\Bem Journal of Artificial Intelligence Research}, {\Bem 34},
  61--88.

\bibitem[\protect\BCAY{Ghosh, Kumar,\ \BBA\ Varakantham}{Ghosh
  et~al.}{2015}]{ghoshKV15}
Ghosh, S., Kumar, A., \BBA\ Varakantham, P. \BBOP2015\BBCP.
\newblock \BBOQ Probabilistic inference based message-passing for resource
  constrained {DCOP}s\BBCQ\
\newblock In {\Bem Proceedings of the International Joint Conference on
  Artificial Intelligence (IJCAI)}, \BPGS\ 411--417.

\bibitem[\protect\BCAY{Giffler\ \BBA\ Thompson}{Giffler\ \BBA\
  Thompson}{1960}]{giffler:60}
Giffler, B.\BBACOMMA\  \BBA\ Thompson, G.~L. \BBOP1960\BBCP.
\newblock \BBOQ Algorithms for solving production-scheduling problems\BBCQ\
\newblock {\Bem Operations research}, {\Bem 8\/}(4), 487--503.

\bibitem[\protect\BCAY{Giuliani, Castelletti, Amigoni,\ \BBA\ Cai}{Giuliani
  et~al.}{2014}]{giuliani:14}
Giuliani, M., Castelletti, A., Amigoni, F., \BBA\ Cai, X. \BBOP2014\BBCP.
\newblock \BBOQ Multiagent systems and distributed constraint reasoning for
  regulatory mechanism design in water management\BBCQ\
\newblock {\Bem Journal of Water Resources Planning and Management}, {\Bem
  41\/}(4).

\bibitem[\protect\BCAY{Gla{\ss}er, Reitwie{\ss}ner, Schmitz,\ \BBA\
  Witek}{Gla{\ss}er et~al.}{2010}]{glasser:10}
Gla{\ss}er, C., Reitwie{\ss}ner, C., Schmitz, H., \BBA\ Witek, M.
  \BBOP2010\BBCP.
\newblock \BBOQ Approximability and hardness in multi-objective
  optimization\BBCQ\
\newblock In {\Bem Proceedings of the Conference on Computability in Europe
  (CiE)}, \BPGS\ 180--189.

\bibitem[\protect\BCAY{Golomb\ \BBA\ Baumert}{Golomb\ \BBA\
  Baumert}{1965}]{golomb:65}
Golomb, S.~W.\BBACOMMA\  \BBA\ Baumert, L.~D. \BBOP1965\BBCP.
\newblock \BBOQ Backtrack programming\BBCQ\
\newblock {\Bem Journal of the ACM}, {\Bem 12\/}(4), 516--524.

\bibitem[\protect\BCAY{Greenstadt, Grosz,\ \BBA\ Smith}{Greenstadt
  et~al.}{2007}]{greenstadt:07}
Greenstadt, R., Grosz, B., \BBA\ Smith, M. \BBOP2007\BBCP.
\newblock \BBOQ {SSDPOP}: Improving the privacy of {DCOP} with secret
  sharing\BBCQ\
\newblock In {\Bem Proceedings of the International Conference on Autonomous
  Agents and Multiagent Systems (AAMAS)}, \BPGS\ 1098--1100.

\bibitem[\protect\BCAY{Grinshpoun}{Grinshpoun}{2015}]{grinshpoun:15}
Grinshpoun, T. \BBOP2015\BBCP.
\newblock \BBOQ Clustering variables by their agents\BBCQ\
\newblock In {\Bem Proceedings of the International Joint Conferences on Web
  Intelligence and Intelligent Agent Technologies (WI-IAT)},
  \lowercase{\BVOL}~2, \BPGS\ 250--256.

\bibitem[\protect\BCAY{Grinshpoun, Grubshtein, Zivan, Netzer,\ \BBA\
  Meisels}{Grinshpoun et~al.}{2013}]{grinshpoun:13}
Grinshpoun, T., Grubshtein, A., Zivan, R., Netzer, A., \BBA\ Meisels, A.
  \BBOP2013\BBCP.
\newblock \BBOQ Asymmetric distributed constraint optimization problems\BBCQ\
\newblock {\Bem Journal of Artificial Intelligence Research}, {\Bem 47},
  613--647.

\bibitem[\protect\BCAY{Grinshpoun\ \BBA\ Meisels}{Grinshpoun\ \BBA\
  Meisels}{2008}]{grinshpoun:08}
Grinshpoun, T.\BBACOMMA\  \BBA\ Meisels, A. \BBOP2008\BBCP.
\newblock \BBOQ Completeness and performance of the {APO} algorithm\BBCQ\
\newblock {\Bem Journal of Artificial Intelligence Research}, {\Bem 33},
  223--258.

\bibitem[\protect\BCAY{Gupta, Jain, Yeoh, Ranade,\ \BBA\ Pontelli}{Gupta
  et~al.}{2013a}]{gupta:13a}
Gupta, S., Jain, P., Yeoh, W., Ranade, S., \BBA\ Pontelli, E. \BBOP2013a\BBCP.
\newblock \BBOQ Solving customer-driven microgrid optimization problems as
  {DCOP}s\BBCQ\
\newblock In {\Bem International Workshop on Distributed Constraint Reasoning
  (DCR)}, \BPGS\ 45--59.

\bibitem[\protect\BCAY{Gupta, Yeoh, Pontelli, Jain,\ \BBA\ Ranade}{Gupta
  et~al.}{2013b}]{gupta:13b}
Gupta, S., Yeoh, W., Pontelli, E., Jain, P., \BBA\ Ranade, S. \BBOP2013b\BBCP.
\newblock \BBOQ Modeling microgrid islanding problems as {DCOP}s\BBCQ\
\newblock In {\Bem Proceedings of the North American Power Symposium (NAPS)},
  \BPGS\ 1--6.

\bibitem[\protect\BCAY{Gutierrez, Lee, Lei, Mak,\ \BBA\ Meseguer}{Gutierrez
  et~al.}{2013}]{gutierrez:13}
Gutierrez, P., Lee, J., Lei, K.~M., Mak, T., \BBA\ Meseguer, P. \BBOP2013\BBCP.
\newblock \BBOQ Maintaining soft arc consistencies in {BnB-ADOPT}$^+$ during
  search\BBCQ\
\newblock In {\Bem Proceedings of the International Conference on Principles
  and Practice of Constraint Programming (CP)}, \BPGS\ 365--380.

\bibitem[\protect\BCAY{Gutierrez\ \BBA\ Meseguer}{Gutierrez\ \BBA\
  Meseguer}{2012a}]{gutierrez:12b}
Gutierrez, P.\BBACOMMA\  \BBA\ Meseguer, P. \BBOP2012a\BBCP.
\newblock \BBOQ Improving {BnB-ADOPT}$^+$-{AC}\BBCQ\
\newblock In {\Bem Proceedings of the International Conference on Autonomous
  Agents and Multiagent Systems (AAMAS)}, \BPGS\ 273--280.

\bibitem[\protect\BCAY{Gutierrez\ \BBA\ Meseguer}{Gutierrez\ \BBA\
  Meseguer}{2012b}]{gutierrez:12}
Gutierrez, P.\BBACOMMA\  \BBA\ Meseguer, P. \BBOP2012b\BBCP.
\newblock \BBOQ Removing redundant messages in n-ary {BnB-ADOPT}\BBCQ\
\newblock {\Bem Journal of Artificial Intelligence Research}, {\Bem 45},
  287--304.

\bibitem[\protect\BCAY{Gutierrez, Meseguer,\ \BBA\ Yeoh}{Gutierrez
  et~al.}{2011}]{gutierrez:11}
Gutierrez, P., Meseguer, P., \BBA\ Yeoh, W. \BBOP2011\BBCP.
\newblock \BBOQ Generalizing {ADOPT} and {BnB-ADOPT}\BBCQ\
\newblock In {\Bem Proceedings of the International Joint Conference on
  Artificial Intelligence (IJCAI)}, \BPGS\ 554--559.

\bibitem[\protect\BCAY{Hannebauer\ \BBA\ M{\"u}ller}{Hannebauer\ \BBA\
  M{\"u}ller}{2001}]{hannebauer:01}
Hannebauer, M.\BBACOMMA\  \BBA\ M{\"u}ller, S. \BBOP2001\BBCP.
\newblock \BBOQ Distributed constraint optimization for medical appointment
  scheduling\BBCQ\
\newblock In {\Bem Proceedings of the International Conference on Autonomous
  Agents and Multiagent Systems (AAMAS)}, \BPGS\ 139--140.

\bibitem[\protect\BCAY{Harvey, Chang,\ \BBA\ Ghose}{Harvey
  et~al.}{2007}]{harvey:07}
Harvey, P., Chang, C.~F., \BBA\ Ghose, A. \BBOP2007\BBCP.
\newblock \BBOQ Support-based distributed search: a new approach for multiagent
  constraint processing\BBCQ\
\newblock In {\Bem Argumentation in Multi-Agent Systems}, \BPGS\ 91--106.
  Springer Berlin Heidelberg.

\bibitem[\protect\BCAY{Hirayama\ \BBA\ Yokoo}{Hirayama\ \BBA\
  Yokoo}{1997}]{hirayama:97}
Hirayama, K.\BBACOMMA\  \BBA\ Yokoo, M. \BBOP1997\BBCP.
\newblock \BBOQ Distributed partial constraint satisfaction problem\BBCQ\
\newblock In {\Bem Proceedings of the International Conference on Principles
  and Practice of Constraint Programming (CP)}, \BPGS\ 222--236.

\bibitem[\protect\BCAY{Hoang, Fioretto, Hou, Yokoo, Yeoh,\ \BBA\ Zivan}{Hoang
  et~al.}{2016}]{fioretto:AAMAS-16a}
Hoang, K.~D., Fioretto, F., Hou, P., Yokoo, M., Yeoh, W., \BBA\ Zivan, R.
  \BBOP2016\BBCP.
\newblock \BBOQ Proactive dynamic distributed constraint optimization\BBCQ\
\newblock In {\Bem Proceedings of the International Joint Conference on
  Autonomous Agents and Multiagent Systems {(AAMAS)}}, \BPGS\ 597--605.

\bibitem[\protect\BCAY{Hoang, Hou, Fioretto, Yeoh, Zivan,\ \BBA\ Yokoo}{Hoang
  et~al.}{2017}]{fioretto:AAMAS-17c}
Hoang, K.~D., Hou, P., Fioretto, F., Yeoh, W., Zivan, R., \BBA\ Yokoo, M.
  \BBOP2017\BBCP.
\newblock \BBOQ Infinite-horizon proactive dynamic {DCOP}s\BBCQ\
\newblock In {\Bem Proceedings of the International Joint Conference on
  Autonomous Agents and Multiagent Systems {(AAMAS)}}, \BPGS\ 212--220.

\bibitem[\protect\BCAY{Hollos, Karl,\ \BBA\ Wolisz}{Hollos
  et~al.}{2004}]{hollos:04}
Hollos, D., Karl, H., \BBA\ Wolisz, A. \BBOP2004\BBCP.
\newblock \BBOQ Regionalizing global optimization algorithms to improve the
  operation of large ad hoc networks\BBCQ\
\newblock In {\Bem Proceedings of the Wireless Communications and Networking
  Conference (WCNC)}, \BPGS\ 819--824.

\bibitem[\protect\BCAY{Jain, Taylor, Tambe,\ \BBA\ Yokoo}{Jain
  et~al.}{2009}]{jain:09}
Jain, M., Taylor, M.~E., Tambe, M., \BBA\ Yokoo, M. \BBOP2009\BBCP.
\newblock \BBOQ {DCOP}s meet the real world: Exploring unknown reward matrices
  with applications to mobile sensor networks\BBCQ\
\newblock In {\Bem Proceedings of the International Joint Conference on
  Artificial Intelligence (IJCAI)}, \BPGS\ 181--186.

\bibitem[\protect\BCAY{Jain, Ranade, Gupta,\ \BBA\ Pontelli}{Jain
  et~al.}{2012}]{jain:12}
Jain, P., Ranade, S.~J., Gupta, S., \BBA\ Pontelli, E. \BBOP2012\BBCP.
\newblock \BBOQ Optimum operation of a customer-driven microgrid: A
  comprehensive approach\BBCQ\
\newblock In {\Bem Proceedings of the IEEE International Conference on Power
  Electronics, Drives and Energy Systems (PEDES)}, \BPGS\ 1--6.

\bibitem[\protect\BCAY{Jennings\ \BBA\ Jackson}{Jennings\ \BBA\
  Jackson}{1995}]{jennings:95}
Jennings, N.~R.\BBACOMMA\  \BBA\ Jackson, A. \BBOP1995\BBCP.
\newblock \BBOQ Agent-based meeting scheduling: A design and
  implementation\BBCQ\
\newblock {\Bem Electronics letters}, {\Bem 31\/}(5), 350--352.

\bibitem[\protect\BCAY{Jin, Cao,\ \BBA\ Li}{Jin et~al.}{2011}]{jin:11}
Jin, Z., Cao, J., \BBA\ Li, M. \BBOP2011\BBCP.
\newblock \BBOQ A distributed application component placement approach for
  cloud computing environment\BBCQ\
\newblock In {\Bem Proceedings of the International Conference on Dependable,
  Autonomic and Secure Computing (DASC)}, \BPGS\ 488--495.

\bibitem[\protect\BCAY{Junges\ \BBA\ Bazzan}{Junges\ \BBA\
  Bazzan}{2008}]{junges:08}
Junges, R.\BBACOMMA\  \BBA\ Bazzan, A.~L. \BBOP2008\BBCP.
\newblock \BBOQ Evaluating the performance of {DCOP} algorithms in a real
  world, dynamic problem\BBCQ\
\newblock In {\Bem Proceedings of the International Conference on Autonomous
  Agents and Multiagent Systems (AAMAS)}, \BPGS\ 599--606.

\bibitem[\protect\BCAY{Kearns, Littman,\ \BBA\ Singh}{Kearns
  et~al.}{2001}]{kearns:01}
Kearns, M., Littman, M.~L., \BBA\ Singh, S. \BBOP2001\BBCP.
\newblock \BBOQ Graphical models for game theory\BBCQ\
\newblock In {\Bem Proceedings of the Conference on Uncertainty in Artificial
  Intelligence (UAI)}, \BPGS\ 253--260.

\bibitem[\protect\BCAY{Khanna, Sattar, Hansen,\ \BBA\ Stantic}{Khanna
  et~al.}{2009}]{khanna:09}
Khanna, S., Sattar, A., Hansen, D., \BBA\ Stantic, B. \BBOP2009\BBCP.
\newblock \BBOQ An efficient algorithm for solving dynamic complex {DCOP}
  problems\BBCQ\
\newblock In {\Bem Proceedings of the International Joint Conferences on Web
  Intelligence and Intelligent Agent Technologies (WI-IAT)}, \BPGS\ 339--346.

\bibitem[\protect\BCAY{Kiekintveld, Yin, Kumar,\ \BBA\ Tambe}{Kiekintveld
  et~al.}{2010}]{kiekintveld:10}
Kiekintveld, C., Yin, Z., Kumar, A., \BBA\ Tambe, M. \BBOP2010\BBCP.
\newblock \BBOQ Asynchronous algorithms for approximate distributed constraint
  optimization with quality bounds\BBCQ\
\newblock In {\Bem Proceedings of the International Conference on Autonomous
  Agents and Multiagent Systems (AAMAS)}, \BPGS\ 133--140.

\bibitem[\protect\BCAY{Kinoshita, Iizuka,\ \BBA\ Iizuka}{Kinoshita
  et~al.}{2013}]{kinoshita:13}
Kinoshita, K., Iizuka, K., \BBA\ Iizuka, Y. \BBOP2013\BBCP.
\newblock \BBOQ Effective disaster evacuation by solving the distributed
  constraint optimization problem\BBCQ\
\newblock In {\Bem Proceedings of the International Conference on Advanced
  Applied Informatics (IIAIAAI)}, \BPGS\ 399--400.

\bibitem[\protect\BCAY{Kluegel, Iqbal, Fioretto, Yeoh,\ \BBA\ Pontelli}{Kluegel
  et~al.}{2017}]{fioretto:JAAMAS-17}
Kluegel, W., Iqbal, M.~A., Fioretto, F., Yeoh, W., \BBA\ Pontelli, E.
  \BBOP2017\BBCP.
\newblock \BBOQ A realistic dataset for the smart home device scheduling
  problem for {DCOPs}\BBCQ\
\newblock In Sukthankar, G.\BBACOMMA\  \BBA\ Rodriguez-Aguilar, J.~A.\BEDS,
  {\Bem Autonomous Agents and Multiagent Systems: AAMAS 2017 Workshops,
  Visionary Papers, S{\~a}o Paulo, Brazil, May 8-12, 2017, Revised Selected
  Papers}, \BPGS\ 125--142. Springer International Publishing.

\bibitem[\protect\BCAY{Kopena, Sultanik, Lass, Nguyen, Dugan, Modi,\ \BBA\
  Regli}{Kopena et~al.}{2008}]{kopena:08}
Kopena, J.~B., Sultanik, E.~A., Lass, R.~N., Nguyen, D.~N., Dugan, C.~J., Modi,
  P.~J., \BBA\ Regli, W.~C. \BBOP2008\BBCP.
\newblock \BBOQ Distributed coordination of first responders\BBCQ\
\newblock {\Bem Internet Computing}, {\Bem 12\/}(1), 45--47.

\bibitem[\protect\BCAY{Kraus}{Kraus}{1997}]{kraus:97}
Kraus, S. \BBOP1997\BBCP.
\newblock \BBOQ Negotiation and cooperation in multi-agent environments\BBCQ\
\newblock {\Bem Artificial Intelligence}, {\Bem 94\/}(1), 79--97.

\bibitem[\protect\BCAY{Kravari\ \BBA\ Bassiliades}{Kravari\ \BBA\
  Bassiliades}{2015}]{kravari:15}
Kravari, K.\BBACOMMA\  \BBA\ Bassiliades, N. \BBOP2015\BBCP.
\newblock \BBOQ A survey of agent platforms\BBCQ\
\newblock {\Bem Journal of Artificial Societies and Social Simulation}, {\Bem
  18\/}(1), 11.

\bibitem[\protect\BCAY{Kschischang, Frey,\ \BBA\ Loeliger}{Kschischang
  et~al.}{2001}]{kschischang:01}
Kschischang, F.~R., Frey, B.~J., \BBA\ Loeliger, H.-A. \BBOP2001\BBCP.
\newblock \BBOQ Factor graphs and the sum-product algorithm\BBCQ\
\newblock {\Bem {IEEE} Transactions on Information Theory}, {\Bem 47\/}(2),
  498--519.

\bibitem[\protect\BCAY{Kumar, Faltings,\ \BBA\ Petcu}{Kumar
  et~al.}{2009}]{kumar:09}
Kumar, A., Faltings, B., \BBA\ Petcu, A. \BBOP2009\BBCP.
\newblock \BBOQ Distributed constraint optimization with structured resource
  constraints\BBCQ\
\newblock In {\Bem Proceedings of the International Conference on Autonomous
  Agents and Multiagent Systems (AAMAS)}, \BPGS\ 923--930.

\bibitem[\protect\BCAY{Kumar, Petcu,\ \BBA\ Faltings}{Kumar
  et~al.}{2008}]{kumar:08}
Kumar, A., Petcu, A., \BBA\ Faltings, B. \BBOP2008\BBCP.
\newblock \BBOQ {H-DPOP}: Using hard constraints for search space pruning in
  {DCOP}\BBCQ\
\newblock In {\Bem Proceedings of the AAAI Conference on Artificial
  Intelligence (AAAI)}, \BPGS\ 325--330.

\bibitem[\protect\BCAY{Kumar\ \BBA\ Zilberstein}{Kumar\ \BBA\
  Zilberstein}{2011}]{kumar:11}
Kumar, A.\BBACOMMA\  \BBA\ Zilberstein, S. \BBOP2011\BBCP.
\newblock \BBOQ Message-passing algorithms for quadratic programming
  formulations of map estimation\BBCQ\
\newblock In {\Bem Proceedings of the Conference on Uncertainty in Artificial
  Intelligence (UAI)}, \BPGS\ 428--435.

\bibitem[\protect\BCAY{Lallouet, Lee, Mak,\ \BBA\ Yip}{Lallouet
  et~al.}{2015}]{Lallouet2015}
Lallouet, A., Lee, J. H.~M., Mak, T. W.~K., \BBA\ Yip, J. \BBOP2015\BBCP.
\newblock \BBOQ Ultra-weak solutions and consistency enforcement in minimax
  weighted constraint satisfaction\BBCQ\
\newblock {\Bem Constraints}, {\Bem 20\/}(2), 109--154.

\bibitem[\protect\BCAY{Larrosa}{Larrosa}{2002}]{larrosa:02}
Larrosa, J. \BBOP2002\BBCP.
\newblock \BBOQ Node and arc consistency in weighted {CSP}\BBCQ\
\newblock In {\Bem Proceedings of the AAAI Conference on Artificial
  Intelligence (AAAI)}, \BPGS\ 48--53.

\bibitem[\protect\BCAY{Lass, Kopena, Sultanik, Nguyen, Dugan, Modi,\ \BBA\
  Regli}{Lass et~al.}{2008a}]{lass:08a}
Lass, R.~N., Kopena, J.~B., Sultanik, E.~A., Nguyen, D.~N., Dugan, C.~P., Modi,
  P.~J., \BBA\ Regli, W.~C. \BBOP2008a\BBCP.
\newblock \BBOQ Coordination of first responders under communication and
  resource constraints\BBCQ\
\newblock In {\Bem Proceedings of the International Conference on Autonomous
  Agents and Multiagent Systems (AAMAS)}, \BPGS\ 1409--1412.

\bibitem[\protect\BCAY{Lass, Regli, Kaplan, Mitkus,\ \BBA\ Sim}{Lass
  et~al.}{2008b}]{lass:08b}
Lass, R.~N., Regli, W.~C., Kaplan, A., Mitkus, M., \BBA\ Sim, J.~J.
  \BBOP2008b\BBCP.
\newblock \BBOQ Facilitating communication for first responders using dynamic
  distributed constraint optimization\BBCQ\
\newblock In {\Bem Proceedings of the Symposium on Technologies for Homeland
  Security ({IEEE} HST)}, \BPGS\ 316--320.

\bibitem[\protect\BCAY{Le, Fioretto, Yeoh, Son,\ \BBA\ Pontelli}{Le
  et~al.}{2016}]{le:16a}
Le, T., Fioretto, F., Yeoh, W., Son, T.~C., \BBA\ Pontelli, E. \BBOP2016\BBCP.
\newblock \BBOQ {ER-DCOP}s: A framework for distributed constraint optimization
  with uncertainty in constraint utilities\BBCQ\
\newblock In {\Bem Proceedings of the International Conference on Autonomous
  Agents and Multiagent Systems (AAMAS)}, \BPGS\ 606--614.

\bibitem[\protect\BCAY{Le, Son, Pontelli,\ \BBA\ Yeoh}{Le et~al.}{2015}]{le:15}
Le, T., Son, T.~C., Pontelli, E., \BBA\ Yeoh, W. \BBOP2015\BBCP.
\newblock \BBOQ Solving distributed constraint optimization problems with logic
  programming\BBCQ\
\newblock In {\Bem Proceedings of the AAAI Conference on Artificial
  Intelligence (AAAI)}, \BPGS\ 1174--1181.

\bibitem[\protect\BCAY{L{\'e}aut{\'e}\ \BBA\ Faltings}{L{\'e}aut{\'e}\ \BBA\
  Faltings}{2009}]{leaute:09a}
L{\'e}aut{\'e}, T.\BBACOMMA\  \BBA\ Faltings, B. \BBOP2009\BBCP.
\newblock \BBOQ {E [DPOP]}: Distributed constraint optimization under
  stochastic uncertainty using collaborative sampling\BBCQ\
\newblock In {\Bem International Workshop on Distributed Constraint Reasoning
  (DCR)}, \BPGS\ 87--101.

\bibitem[\protect\BCAY{L{\'e}aut{\'e}\ \BBA\ Faltings}{L{\'e}aut{\'e}\ \BBA\
  Faltings}{2011}]{leaute:11}
L{\'e}aut{\'e}, T.\BBACOMMA\  \BBA\ Faltings, B. \BBOP2011\BBCP.
\newblock \BBOQ Distributed constraint optimization under stochastic
  uncertainty\BBCQ\
\newblock In {\Bem Proceedings of the AAAI Conference on Artificial
  Intelligence (AAAI)}, \BPGS\ 68--73.

\bibitem[\protect\BCAY{L{\'e}aut{\'e}, Ottens,\ \BBA\ Szymanek}{L{\'e}aut{\'e}
  et~al.}{2009}]{leaute:09b}
L{\'e}aut{\'e}, T., Ottens, B., \BBA\ Szymanek, R. \BBOP2009\BBCP.
\newblock \BBOQ {FRODO} 2.0: An open-source framework for distributed
  constraint optimization\BBCQ\
\newblock In {\Bem International Workshop on Distributed Constraint Reasoning
  (DCR)}, \BPGS\ 160--164.

\bibitem[\protect\BCAY{Lesser, Decker, Wagner, Carver, Garvey, Horling, Neiman,
  Podorozhny, Prasad, Raja, et~al.}{Lesser et~al.}{2004}]{lesser:04}
Lesser, V., Decker, K., Wagner, T., Carver, N., Garvey, A., Horling, B.,
  Neiman, D., Podorozhny, R., Prasad, M.~N., Raja, A., et~al. \BBOP2004\BBCP.
\newblock \BBOQ Evolution of the {GPGP/TAEMS} domain-independent coordination
  framework\BBCQ\
\newblock {\Bem Journal of Autonomous Agents and Multi-Agent Systems}, {\Bem
  9\/}(1-2), 87--143.

\bibitem[\protect\BCAY{Levit, Grinshpoun, Meisels,\ \BBA\ Bazzan}{Levit
  et~al.}{2013}]{levit:13}
Levit, V., Grinshpoun, T., Meisels, A., \BBA\ Bazzan, A.~L. \BBOP2013\BBCP.
\newblock \BBOQ Taxation search in boolean games\BBCQ\
\newblock In {\Bem Proceedings of the International Conference on Autonomous
  Agents and Multiagent Systems (AAMAS)}, \BPGS\ 183--190.

\bibitem[\protect\BCAY{Li, Negenborn,\ \BBA\ Lodewijks}{Li
  et~al.}{2016}]{li:16}
Li, S., Negenborn, R.~R., \BBA\ Lodewijks, G. \BBOP2016\BBCP.
\newblock \BBOQ Distributed constraint optimization for addressing vessel
  rotation planning problems\BBCQ\
\newblock {\Bem Engineering Applications of Artificial Intelligence}, {\Bem
  48}, 159--172.

\bibitem[\protect\BCAY{Li, Wang, Ding,\ \BBA\ Li}{Li et~al.}{2014}]{li:14}
Li, X., Wang, H., Ding, B., \BBA\ Li, X. \BBOP2014\BBCP.
\newblock \BBOQ {MABP}: an optimal resource allocation approach in data center
  networks\BBCQ\
\newblock {\Bem Science China Information Sciences}, {\Bem 57\/}(10), 1--16.

\bibitem[\protect\BCAY{Lorenzi, dos Santos, Ferreira~Jr,\ \BBA\ Bazzan}{Lorenzi
  et~al.}{2008}]{lorenzi:08}
Lorenzi, F., dos Santos, F., Ferreira~Jr, P.~R., \BBA\ Bazzan, A.~L.
  \BBOP2008\BBCP.
\newblock \BBOQ Optimizing preferences within groups: A case study on travel
  recommendation\BBCQ\
\newblock In {\Bem Proceedings of the Brazilian Symposium on Artificial
  Intelligence (SBIA)}, \BPGS\ 103--112.

\bibitem[\protect\BCAY{Lutati, Gontmakher, Lando, Netzer, Meisels,\ \BBA\
  Grubshtein}{Lutati et~al.}{2014}]{lutati:14}
Lutati, B., Gontmakher, I., Lando, M., Netzer, A., Meisels, A., \BBA\
  Grubshtein, A. \BBOP2014\BBCP.
\newblock \BBOQ {AgentZero}: A framework for simulating and evaluating
  multi-agent algorithms\BBCQ\
\newblock In {\Bem Agent-Oriented Software Engineering}, \BPGS\ 309--327.
  Springer Berlin Heidelberg.

\bibitem[\protect\BCAY{Lynch}{Lynch}{1996}]{lynch:96}
Lynch, N.~A. \BBOP1996\BBCP.
\newblock {\Bem Distributed Algorithms}.
\newblock Morgan Kaufmann.

\bibitem[\protect\BCAY{Macarthur, Farinelli, Ramchurn,\ \BBA\
  Jennings}{Macarthur et~al.}{2010}]{macarthur:10}
Macarthur, K., Farinelli, A., Ramchurn, S., \BBA\ Jennings, N. \BBOP2010\BBCP.
\newblock \BBOQ Efficient, superstabilizing decentralised optimisation for
  dynamic task allocation environments\BBCQ\
\newblock In {\Bem International Workshop on Optimization In Multi-Agent
  Systems (OPTMAS)}, \BPGS\ 25--32.

\bibitem[\protect\BCAY{Macarthur, Farinelli, Ramchurn,\ \BBA\
  Jennings}{Macarthur et~al.}{2011}]{macarthur:11}
Macarthur, K., Farinelli, A., Ramchurn, S., \BBA\ Jennings, N. \BBOP2011\BBCP.
\newblock \BBOQ A distributed anytime algorithm for dynamic task allocation in
  multi-agent systems\BBCQ\
\newblock In {\Bem Proceedings of the AAAI Conference on Artificial
  Intelligence (AAAI)}, \BPGS\ 701--706.

\bibitem[\protect\BCAY{Mackworth\ \BBA\ Freuder}{Mackworth\ \BBA\
  Freuder}{1985}]{mackworth:85}
Mackworth, A.~K.\BBACOMMA\  \BBA\ Freuder, E.~C. \BBOP1985\BBCP.
\newblock \BBOQ The complexity of some polynomial network consistency
  algorithms for constraint satisfaction problems\BBCQ\
\newblock {\Bem Artificial intelligence}, {\Bem 25\/}(1), 65--74.

\bibitem[\protect\BCAY{Mahadevan}{Mahadevan}{1996}]{mahadevan:96}
Mahadevan, S. \BBOP1996\BBCP.
\newblock \BBOQ Average reward reinforcement learning: Foundations, algorithms,
  and empirical results\BBCQ\
\newblock {\Bem Machine Learning}, {\Bem 22\/}(1--3), 159--195.

\bibitem[\protect\BCAY{Maheswaran, Pearce,\ \BBA\ Tambe}{Maheswaran
  et~al.}{2004a}]{maheswaran:04b}
Maheswaran, R., Pearce, J., \BBA\ Tambe, M. \BBOP2004a\BBCP.
\newblock \BBOQ Distributed algorithms for {DCOP}: A graphical game-based
  approach\BBCQ\
\newblock In {\Bem Proceedings of the International Conference on Parallel and
  Distributed Computing Systems (PDCS)}, \BPGS\ 432--439.

\bibitem[\protect\BCAY{Maheswaran, Tambe, Bowring, Pearce,\ \BBA\
  Varakantham}{Maheswaran et~al.}{2004b}]{maheswaran:04}
Maheswaran, R.~T., Tambe, M., Bowring, E., Pearce, J.~P., \BBA\ Varakantham, P.
  \BBOP2004b\BBCP.
\newblock \BBOQ {Taking {DCOP} to the real world: Efficient complete solutions
  for distributed multi-event scheduling}\BBCQ\
\newblock In {\Bem Proceedings of the International Conference on Autonomous
  Agents and Multiagent Systems (AAMAS)}, \BPGS\ 310--317.

\bibitem[\protect\BCAY{Mailler\ \BBA\ Lesser}{Mailler\ \BBA\
  Lesser}{2004}]{mailler:04}
Mailler, R.\BBACOMMA\  \BBA\ Lesser, V. \BBOP2004\BBCP.
\newblock \BBOQ Solving distributed constraint optimization problems using
  cooperative mediation\BBCQ\
\newblock In {\Bem Proceedings of the International Conference on Autonomous
  Agents and Multiagent Systems (AAMAS)}, \BPGS\ 438--445.

\bibitem[\protect\BCAY{Marler\ \BBA\ Arora}{Marler\ \BBA\
  Arora}{2004}]{marler:04}
Marler, R.~T.\BBACOMMA\  \BBA\ Arora, J.~S. \BBOP2004\BBCP.
\newblock \BBOQ Survey of multi-objective optimization methods for
  engineering\BBCQ\
\newblock {\Bem Structural and Multidisciplinary Optimization}, {\Bem 26\/}(6),
  369--395.

\bibitem[\protect\BCAY{Matsui, Matsuo, Silaghi, Hirayama, Yokoo,\ \BBA\
  Baba}{Matsui et~al.}{2010}]{matsui:10}
Matsui, T., Matsuo, H., Silaghi, M.~C., Hirayama, K., Yokoo, M., \BBA\ Baba, S.
  \BBOP2010\BBCP.
\newblock \BBOQ A quantified distributed constraint optimization problem\BBCQ\
\newblock In {\Bem Proceedings of the International Conference on Autonomous
  Agents and Multiagent Systems (AAMAS)}, \BPGS\ 1023--1030.

\bibitem[\protect\BCAY{Matsui\ \BBA\ Matsuo}{Matsui\ \BBA\
  Matsuo}{2008}]{matsui:08b}
Matsui, T.\BBACOMMA\  \BBA\ Matsuo, H. \BBOP2008\BBCP.
\newblock \BBOQ A formalization for distributed cooperative sensor resource
  allocation\BBCQ\
\newblock In {\Bem Agent and Multi-Agent Systems: Technologies and
  Applications}, \BPGS\ 292--301.

\bibitem[\protect\BCAY{Matsui\ \BBA\ Matsuo}{Matsui\ \BBA\
  Matsuo}{2011}]{matsui:11}
Matsui, T.\BBACOMMA\  \BBA\ Matsuo, H. \BBOP2011\BBCP.
\newblock \BBOQ A distributed cooperative model for resource supply
  networks\BBCQ\
\newblock In {\Bem Proceedings of the International MultiConference of
  Engineers and Computer Scientists}, \BPGS\ 7--14.

\bibitem[\protect\BCAY{Matsui, Matsuo, Silaghi, Hirayama,\ \BBA\ Yokoo}{Matsui
  et~al.}{2008}]{matsui:08a}
Matsui, T., Matsuo, H., Silaghi, M., Hirayama, K., \BBA\ Yokoo, M.
  \BBOP2008\BBCP.
\newblock \BBOQ Resource constrained distributed constraint optimization with
  virtual variables\BBCQ\
\newblock In {\Bem Proceedings of the AAAI Conference on Artificial
  Intelligence (AAAI)}, \BPGS\ 120--125.

\bibitem[\protect\BCAY{Matsui, Silaghi, Hirayama, Yokoo,\ \BBA\ Matsuo}{Matsui
  et~al.}{2012}]{matsui:12a}
Matsui, T., Silaghi, M., Hirayama, K., Yokoo, M., \BBA\ Matsuo, H.
  \BBOP2012\BBCP.
\newblock \BBOQ Distributed search method with bounded cost vectors on multiple
  objective {DCOP}s\BBCQ\
\newblock In {\Bem Proceedings of the Principles and Practice of Multi-Agent
  Systems (PRIMA)}, \BPGS\ 137--152.

\bibitem[\protect\BCAY{Matsui, Silaghi, Hirayama, Yokoo,\ \BBA\ Matsuo}{Matsui
  et~al.}{2014}]{matsui:14}
Matsui, T., Silaghi, M., Hirayama, K., Yokoo, M., \BBA\ Matsuo, H.
  \BBOP2014\BBCP.
\newblock \BBOQ Leximin multiple objective optimization for preferences of
  agents\BBCQ\
\newblock In {\Bem Proceedings of the Principles and Practice of Multi-Agent
  Systems (PRIMA)}, \BPGS\ 423--438.

\bibitem[\protect\BCAY{Medi, Okimoto,\ \BBA\ Inoue}{Medi
  et~al.}{2014}]{medi:14}
Medi, A., Okimoto, T., \BBA\ Inoue, K. \BBOP2014\BBCP.
\newblock \BBOQ A two-phase complete algorithm for multi-objective distributed
  constraint optimization\BBCQ\
\newblock {\Bem Journal of Advanced Computational Intelligence and Intelligent
  Informatics}, {\Bem 18\/}(4), 573--580.

\bibitem[\protect\BCAY{Meisels}{Meisels}{2008}]{meisels:08}
Meisels, A. \BBOP2008\BBCP.
\newblock {\Bem Distributed Search by Constrained Agents: Algorithms,
  Performance, Communication}.
\newblock Springer Science \& Business Media.

\bibitem[\protect\BCAY{Meisels, Kaplansky, Razgon,\ \BBA\ Zivan}{Meisels
  et~al.}{2002}]{meisels:02}
Meisels, A., Kaplansky, E., Razgon, I., \BBA\ Zivan, R. \BBOP2002\BBCP.
\newblock \BBOQ Comparing performance of distributed constraints processing
  algorithms\BBCQ\
\newblock In {\Bem International Workshop on Distributed Constraint Reasoning
  (DCR)}, \BPGS\ 86--93.

\bibitem[\protect\BCAY{Mejias\ \BBA\ Roy}{Mejias\ \BBA\ Roy}{2010}]{mejias:10}
Mejias, B.\BBACOMMA\  \BBA\ Roy, P.~V. \BBOP2010\BBCP.
\newblock \BBOQ From mini-clouds to cloud computing\BBCQ\
\newblock In {\Bem International Conference on Self-Adaptive and
  Self-Organizing Systems Workshop (SASOW)}, \BPGS\ 234--238.

\bibitem[\protect\BCAY{Miettinen}{Miettinen}{1999}]{miettinen:99}
Miettinen, K. \BBOP1999\BBCP.
\newblock {\Bem Nonlinear Multiobjective Optimization}, \lowercase{\BVOL}~12.
\newblock Springer Berlin Heidelberg.

\bibitem[\protect\BCAY{Miller, Ramchurn,\ \BBA\ Rogers}{Miller
  et~al.}{2012}]{miller:12}
Miller, S., Ramchurn, S.~D., \BBA\ Rogers, A. \BBOP2012\BBCP.
\newblock \BBOQ Optimal decentralised dispatch of embedded generation in the
  smart grid\BBCQ\
\newblock In {\Bem Proceedings of the International Conference on Autonomous
  Agents and Multiagent Systems (AAMAS)}, \BPGS\ 281--288.

\bibitem[\protect\BCAY{Minton, Johnston, Philips,\ \BBA\ Laird}{Minton
  et~al.}{1992}]{minton:92}
Minton, S., Johnston, M.~D., Philips, A.~B., \BBA\ Laird, P. \BBOP1992\BBCP.
\newblock \BBOQ Minimizing conflicts: a heuristic repair method for constraint
  satisfaction and scheduling problems\BBCQ\
\newblock {\Bem Artificial Intelligence}, {\Bem 58\/}(1), 161--205.

\bibitem[\protect\BCAY{Miorandi, Sicari, De~Pellegrini,\ \BBA\
  Chlamtac}{Miorandi et~al.}{2012}]{miorandi:12}
Miorandi, D., Sicari, S., De~Pellegrini, F., \BBA\ Chlamtac, I. \BBOP2012\BBCP.
\newblock \BBOQ Internet of things: Vision, applications and research
  challenges\BBCQ\
\newblock {\Bem Ad Hoc Networks}, {\Bem 10\/}(7), 1497--1516.

\bibitem[\protect\BCAY{Mir, Merghem-Boulahia,\ \BBA\ Ga{\"\i}ti}{Mir
  et~al.}{2010}]{mir:10}
Mir, U., Merghem-Boulahia, L., \BBA\ Ga{\"\i}ti, D. \BBOP2010\BBCP.
\newblock \BBOQ A cooperative multiagent based spectrum sharing\BBCQ\
\newblock In {\Bem Advanced International Conference on Telecommunications
  (AICT)}, \BPGS\ 124--130.

\bibitem[\protect\BCAY{Modi, Shen, Tambe,\ \BBA\ Yokoo}{Modi
  et~al.}{2005}]{modi:05}
Modi, P., Shen, W.-M., Tambe, M., \BBA\ Yokoo, M. \BBOP2005\BBCP.
\newblock \BBOQ {ADOPT}: Asynchronous distributed constraint optimization with
  quality guarantees\BBCQ\
\newblock {\Bem Artificial Intelligence}, {\Bem 161\/}(1--2), 149--180.

\bibitem[\protect\BCAY{Molisch}{Molisch}{2012}]{molisch:12}
Molisch, A.~F. \BBOP2012\BBCP.
\newblock {\Bem Wireless Communications}, \lowercase{\BVOL}~34.
\newblock John Wiley \& Sons.

\bibitem[\protect\BCAY{Monasson, Zecchina, Kirkpatrick, Selman,\ \BBA\
  Troyansky}{Monasson et~al.}{1999}]{monasson:99}
Monasson, R., Zecchina, R., Kirkpatrick, S., Selman, B., \BBA\ Troyansky, L.
  \BBOP1999\BBCP.
\newblock \BBOQ Determining computational complexity from characteristic
  ‘phase transitions’\BBCQ\
\newblock {\Bem Nature}, {\Bem 400\/}(6740), 133--137.

\bibitem[\protect\BCAY{Monteiro, Pellenz, Penna, Enembreck, Souza,\ \BBA\
  Pujolle}{Monteiro et~al.}{2012a}]{monteiro:12a}
Monteiro, T.~L., Pellenz, M.~E., Penna, M.~C., Enembreck, F., Souza, R.~D.,
  \BBA\ Pujolle, G. \BBOP2012a\BBCP.
\newblock \BBOQ Channel allocation algorithms for {WLAN}s using distributed
  optimization\BBCQ\
\newblock {\Bem AEU-International Journal of Electronics and Communications},
  {\Bem 66\/}(6), 480--490.

\bibitem[\protect\BCAY{Monteiro, Pujolle, Pellenz, Penna, Enembreck,\ \BBA\
  Demo~Souza}{Monteiro et~al.}{2012b}]{monteiro:12b}
Monteiro, T.~L., Pujolle, G., Pellenz, M.~E., Penna, M.~C., Enembreck, F.,
  \BBA\ Demo~Souza, R. \BBOP2012b\BBCP.
\newblock \BBOQ A multi-agent approach to optimal channel assignment in
  {WLAN}s\BBCQ\
\newblock In {\Bem Proceedings of the Wireless Communications and Networking
  Conference (WCNC)}, \BPGS\ 2637--2642.

\bibitem[\protect\BCAY{Moreno-Vozmediano, Montero,\ \BBA\
  Llorente}{Moreno-Vozmediano et~al.}{2013}]{moreno:13}
Moreno-Vozmediano, R., Montero, R.~S., \BBA\ Llorente, I.~M. \BBOP2013\BBCP.
\newblock \BBOQ Key challenges in cloud computing: Enabling the future internet
  of services\BBCQ\
\newblock {\Bem Internet Computing}, {\Bem 17\/}(4), 18--25.

\bibitem[\protect\BCAY{Nair, Varakantham, Tambe,\ \BBA\ Yokoo}{Nair
  et~al.}{2005}]{nair:05}
Nair, R., Varakantham, P., Tambe, M., \BBA\ Yokoo, M. \BBOP2005\BBCP.
\newblock \BBOQ Networked distributed {POMDPs}: A synergy of distributed
  constraint optimization and {POMDPs}\BBCQ\
\newblock In {\Bem Proceedings of the International Joint Conference on
  Artificial Intelligence (IJCAI)}, \BPGS\ 1758--1760.

\bibitem[\protect\BCAY{Nethercote, Stuckey, Becket, Brand, Duck,\ \BBA\
  Tack}{Nethercote et~al.}{2007}]{minizinc}
Nethercote, N., Stuckey, P.~J., Becket, R., Brand, S., Duck, G.~J., \BBA\ Tack,
  G. \BBOP2007\BBCP.
\newblock \BBOQ Minizinc: Towards a standard {CP} modelling language\BBCQ\
\newblock In {\Bem Proceedings of the International Conference on Principles
  and Practice of Constraint Programming (CP)}, \BPGS\ 529--543.

\bibitem[\protect\BCAY{Netzer, Grubshtein,\ \BBA\ Meisels}{Netzer
  et~al.}{2012}]{netzer:12}
Netzer, A., Grubshtein, A., \BBA\ Meisels, A. \BBOP2012\BBCP.
\newblock \BBOQ Concurrent forward bounding for distributed constraint
  optimization problems\BBCQ\
\newblock {\Bem Artificial Intelligence}, {\Bem 193}, 186--216.

\bibitem[\protect\BCAY{Nguyen, Yeoh,\ \BBA\ Lau}{Nguyen
  et~al.}{2012}]{nguyen:12}
Nguyen, D.~T., Yeoh, W., \BBA\ Lau, H.~C. \BBOP2012\BBCP.
\newblock \BBOQ Stochastic dominance in stochastic {DCOP}s for risk-sensitive
  applications\BBCQ\
\newblock In {\Bem Proceedings of the International Conference on Autonomous
  Agents and Multiagent Systems (AAMAS)}, \BPGS\ 257--264.

\bibitem[\protect\BCAY{Nguyen, Yeoh,\ \BBA\ Lau}{Nguyen
  et~al.}{2013}]{nguyen:13}
Nguyen, D.~T., Yeoh, W., \BBA\ Lau, H.~C. \BBOP2013\BBCP.
\newblock \BBOQ Distributed {G}ibbs: A memory-bounded sampling-based {DCOP}
  algorithm\BBCQ\
\newblock In {\Bem Proceedings of the International Conference on Autonomous
  Agents and Multiagent Systems (AAMAS)}, \BPGS\ 167--174.

\bibitem[\protect\BCAY{Nguyen, Yeoh, Lau, Zilberstein,\ \BBA\ Zhang}{Nguyen
  et~al.}{2014}]{nguyen:14}
Nguyen, D.~T., Yeoh, W., Lau, H.~C., Zilberstein, S., \BBA\ Zhang, C.
  \BBOP2014\BBCP.
\newblock \BBOQ Decentralized multi-agent reinforcement learning in
  average-reward dynamic {DCOP}s\BBCQ\
\newblock In {\Bem Proceedings of the International Conference on Autonomous
  Agents and Multiagent Systems (AAMAS)}, \BPGS\ 1341--1342.

\bibitem[\protect\BCAY{Noriega\ \BBA\ Sierra}{Noriega\ \BBA\
  Sierra}{1999}]{noriega:99}
Noriega, P.\BBACOMMA\  \BBA\ Sierra, C. \BBOP1999\BBCP.
\newblock \BBOQ Auctions and multi-agent systems\BBCQ\
\newblock In {\Bem Intelligent Information Agents}, \BPGS\ 153--175. Springer.

\bibitem[\protect\BCAY{Okimoto, Clement,\ \BBA\ Inoue}{Okimoto
  et~al.}{2013}]{okimoto:13a}
Okimoto, T., Clement, M., \BBA\ Inoue, K. \BBOP2013\BBCP.
\newblock \BBOQ {AOF}-based algorithm for dynamic {Multi-Objective} distributed
  constraint optimization\BBCQ\
\newblock In {\Bem Multi-disciplinary Trends in Artificial Intelligence
  (MIWAI)}, \BPGS\ 175--186.

\bibitem[\protect\BCAY{Okimoto, Schwind, Clement,\ \BBA\ Inoue}{Okimoto
  et~al.}{2014}]{okimoto:14}
Okimoto, T., Schwind, N., Clement, M., \BBA\ Inoue, K. \BBOP2014\BBCP.
\newblock \BBOQ {Lp-Norm} based algorithm for multi-objective distributed
  constraint optimization ({E}xtended {A}bstract)\BBCQ\
\newblock In {\Bem Proceedings of the International Conference on Autonomous
  Agents and Multiagent Systems (AAMAS)}, \BPGS\ 1427--1428.

\bibitem[\protect\BCAY{Ota, Matsui,\ \BBA\ Matsuo}{Ota et~al.}{2009}]{ota:09}
Ota, K., Matsui, T., \BBA\ Matsuo, H. \BBOP2009\BBCP.
\newblock \BBOQ Layered distributed constraint optimization problem for
  resource allocation problem in distributed sensor networks\BBCQ\
\newblock In {\Bem Proceedings of the Principles and Practice of Multi-Agent
  Systems (PRIMA)}, \BPGS\ 245--260.

\bibitem[\protect\BCAY{Ottens, Dimitrakakis,\ \BBA\ Faltings}{Ottens
  et~al.}{2017}]{ottens:17}
Ottens, B., Dimitrakakis, C., \BBA\ Faltings, B. \BBOP2017\BBCP.
\newblock \BBOQ {DUCT:} an upper confidence bound approach to distributed
  constraint optimization problems\BBCQ\
\newblock {\Bem {ACM} Transactions on Intelligent Systems and Technology},
  {\Bem 8\/}(5), 69:1--69:27.

\bibitem[\protect\BCAY{Paquete, Chiarandini,\ \BBA\ Stützle}{Paquete
  et~al.}{2004}]{paquete:04}
Paquete, L., Chiarandini, M., \BBA\ Stützle, T. \BBOP2004\BBCP.
\newblock \BBOQ Pareto local optimum sets in the biobjective traveling salesman
  problem: An experimental study\BBCQ\
\newblock In {\Bem Metaheuristics for Multiobjective Optimisation},
  \lowercase{\BVOL}\ 535, \BPGS\ 177--199.

\bibitem[\protect\BCAY{Parsons\ \BBA\ Wooldridge}{Parsons\ \BBA\
  Wooldridge}{2002}]{parsons:02}
Parsons, S.\BBACOMMA\  \BBA\ Wooldridge, M. \BBOP2002\BBCP.
\newblock \BBOQ Game theory and decision theory in multi-agent systems\BBCQ\
\newblock {\Bem Autonomous Agents and Multi-Agent Systems}, {\Bem 5\/}(3),
  243--254.

\bibitem[\protect\BCAY{Pearce\ \BBA\ Tambe}{Pearce\ \BBA\
  Tambe}{2007}]{pearce:07}
Pearce, J.\BBACOMMA\  \BBA\ Tambe, M. \BBOP2007\BBCP.
\newblock \BBOQ Quality guarantees on k-optimal solutions for distributed
  constraint optimization problems\BBCQ\
\newblock In {\Bem Proceedings of the International Joint Conference on
  Artificial Intelligence (IJCAI)}, \BPGS\ 1446--1451.

\bibitem[\protect\BCAY{Pecora, Modi,\ \BBA\ Scerri}{Pecora
  et~al.}{2006}]{pecora:06}
Pecora, F., Modi, P., \BBA\ Scerri, P. \BBOP2006\BBCP.
\newblock \BBOQ Reasoning about and dynamically posting n-ary constraints in
  adopt\BBCQ\
\newblock In {\Bem International Workshop on Distributed Constraint Reasoning
  (DCR)}.

\bibitem[\protect\BCAY{Penya-Alba, Cerquides, Rodriguez-Aguilar,\ \BBA\
  Vinyals}{Penya-Alba et~al.}{2012}]{penya:12a}
Penya-Alba, T., Cerquides, J., Rodriguez-Aguilar, J.~A., \BBA\ Vinyals, M.
  \BBOP2012\BBCP.
\newblock \BBOQ Scalable decentralized supply chain formation through binarized
  belief propagation\BBCQ\
\newblock In {\Bem Proceedings of the International Conference on Autonomous
  Agents and Multiagent Systems (AAMAS)}, \BPGS\ 1275--1276.

\bibitem[\protect\BCAY{Penya-Alba, Vinyals, Cerquides,\ \BBA\
  Rodriguez-Aguilar}{Penya-Alba et~al.}{2014}]{penya:14}
Penya-Alba, T., Vinyals, M., Cerquides, J., \BBA\ Rodriguez-Aguilar, J.~A.
  \BBOP2014\BBCP.
\newblock \BBOQ Exploiting {Max-Sum} for the decentralized assembly of
  high-valued supply chains\BBCQ\
\newblock In {\Bem Proceedings of the International Conference on Autonomous
  Agents and Multiagent Systems (AAMAS)}, \BPGS\ 373--380.

\bibitem[\protect\BCAY{Penya-Alba, Vinyals, Cerquides,\ \BBA\
  Rodriguez-Aguilar}{Penya-Alba et~al.}{2012}]{penya:12b}
Penya-Alba, T., Vinyals, M., Cerquides, J., \BBA\ Rodriguez-Aguilar, J.~A.
  \BBOP2012\BBCP.
\newblock \BBOQ A scalable message-passing algorithm for supply chain
  formation\BBCQ\
\newblock In {\Bem Proceedings of the AAAI Conference on Artificial
  Intelligence (AAAI)}, \BPGS\ 1436--1442.

\bibitem[\protect\BCAY{Peri\ \BBA\ Meisels}{Peri\ \BBA\
  Meisels}{2013}]{peri:13}
Peri, O.\BBACOMMA\  \BBA\ Meisels, A. \BBOP2013\BBCP.
\newblock \BBOQ Synchronizing for performance-{DCOP} algorithms\BBCQ\
\newblock In {\Bem Proceedings of the International Conference on Agents and
  Artificial Intelligence (ICAART)}, \BPGS\ 5--14.

\bibitem[\protect\BCAY{Petcu\ \BBA\ Faltings}{Petcu\ \BBA\
  Faltings}{2005a}]{petcu:05c}
Petcu, A.\BBACOMMA\  \BBA\ Faltings, B. \BBOP2005a\BBCP.
\newblock \BBOQ Approximations in distributed optimization\BBCQ\
\newblock In {\Bem Proceedings of the International Conference on Principles
  and Practice of Constraint Programming (CP)}, \BPGS\ 802--806.

\bibitem[\protect\BCAY{Petcu\ \BBA\ Faltings}{Petcu\ \BBA\
  Faltings}{2005b}]{petcu:05}
Petcu, A.\BBACOMMA\  \BBA\ Faltings, B. \BBOP2005b\BBCP.
\newblock \BBOQ A scalable method for multiagent constraint optimization\BBCQ\
\newblock In {\Bem Proceedings of the International Joint Conference on
  Artificial Intelligence (IJCAI)}, \BPGS\ 1413--1420.

\bibitem[\protect\BCAY{Petcu\ \BBA\ Faltings}{Petcu\ \BBA\
  Faltings}{2005c}]{petcu:05b}
Petcu, A.\BBACOMMA\  \BBA\ Faltings, B. \BBOP2005c\BBCP.
\newblock \BBOQ Superstabilizing, fault-containing distributed combinatorial
  optimization\BBCQ\
\newblock In {\Bem Proceedings of the AAAI Conference on Artificial
  Intelligence (AAAI)}, \BPGS\ 449--454.

\bibitem[\protect\BCAY{Petcu\ \BBA\ Faltings}{Petcu\ \BBA\
  Faltings}{2006}]{petcu:06b}
Petcu, A.\BBACOMMA\  \BBA\ Faltings, B. \BBOP2006\BBCP.
\newblock \BBOQ {ODPOP}: An algorithm for open/distributed constraint
  optimization\BBCQ\
\newblock In {\Bem Proceedings of the AAAI Conference on Artificial
  Intelligence (AAAI)}, \BPGS\ 703--708.

\bibitem[\protect\BCAY{Petcu\ \BBA\ Faltings}{Petcu\ \BBA\
  Faltings}{2007a}]{petcu:07}
Petcu, A.\BBACOMMA\  \BBA\ Faltings, B. \BBOP2007a\BBCP.
\newblock \BBOQ {MB-DPOP}: A new memory-bounded algorithm for distributed
  optimization\BBCQ\
\newblock In {\Bem Proceedings of the International Joint Conference on
  Artificial Intelligence (IJCAI)}, \BPGS\ 1452--1457.

\bibitem[\protect\BCAY{Petcu\ \BBA\ Faltings}{Petcu\ \BBA\
  Faltings}{2007b}]{petcu:07a}
Petcu, A.\BBACOMMA\  \BBA\ Faltings, B. \BBOP2007b\BBCP.
\newblock \BBOQ Optimal solution stability in dynamic, distributed constraint
  optimization\BBCQ\
\newblock In {\Bem Proceedings of the International Conference on Intelligent
  Agent Technology (IAT)}, \BPGS\ 321--327.

\bibitem[\protect\BCAY{Petcu, Faltings,\ \BBA\ Mailler}{Petcu
  et~al.}{2007}]{petcu:07b}
Petcu, A., Faltings, B., \BBA\ Mailler, R. \BBOP2007\BBCP.
\newblock \BBOQ {PC-DPOP}: A new partial centralization algorithm for
  distributed optimization\BBCQ\
\newblock In {\Bem Proceedings of the International Joint Conference on
  Artificial Intelligence (IJCAI)}, \BPGS\ 167--172.

\bibitem[\protect\BCAY{Pham, Tawfik,\ \BBA\ Taylor}{Pham
  et~al.}{2013}]{pham:13b}
Pham, T., Tawfik, A., \BBA\ Taylor, M. \BBOP2013\BBCP.
\newblock \BBOQ A simple, naive agent-based model for the optimization of a
  system of traffic lights: Insights from an exploratory experiment\BBCQ\
\newblock In {\Bem Proceedings of the Conference on Agent-based Modeling in
  Transportation Planning and Operations}, \BPGS\ 1--21.

\bibitem[\protect\BCAY{Portway\ \BBA\ Durfee}{Portway\ \BBA\
  Durfee}{2010}]{portway:10}
Portway, C.\BBACOMMA\  \BBA\ Durfee, E.~H. \BBOP2010\BBCP.
\newblock \BBOQ The multi variable multi constrained distributed constraint
  optimization framework\BBCQ\
\newblock In {\Bem Proceedings of the International Conference on Autonomous
  Agents and Multiagent Systems (AAMAS)}, \BPGS\ 1385--1386.

\bibitem[\protect\BCAY{Pujol-Gonzalez, Cerquides, Farinelli, Meseguer,\ \BBA\
  Rodriguez-Aguilar}{Pujol-Gonzalez et~al.}{2015}]{pujol:15}
Pujol-Gonzalez, M., Cerquides, J., Farinelli, A., Meseguer, P., \BBA\
  Rodriguez-Aguilar, J.~A. \BBOP2015\BBCP.
\newblock \BBOQ Efficient inter-team task allocation in {RoboCup} rescue\BBCQ\
\newblock In {\Bem Proceedings of the International Conference on Autonomous
  Agents and Multiagent Systems (AAMAS)}, \BPGS\ 413--421.

\bibitem[\protect\BCAY{Raiffa}{Raiffa}{1968}]{raiffa:1968}
Raiffa, H. \BBOP1968\BBCP.
\newblock {\Bem Decision Analysis: Introductory Lectures on Choices under
  Uncertainty}.
\newblock Addison-Wesley.

\bibitem[\protect\BCAY{Ramchurn, Farinelli, Macarthur,\ \BBA\
  Jennings}{Ramchurn et~al.}{2010}]{ramchurn:10}
Ramchurn, S.~D., Farinelli, A., Macarthur, K.~S., \BBA\ Jennings, N.~R.
  \BBOP2010\BBCP.
\newblock \BBOQ Decentralized coordination in {RoboCup} rescue\BBCQ\
\newblock {\Bem Computer Journal}, {\Bem 53\/}(9), 1447--1461.

\bibitem[\protect\BCAY{Ramchurn, Vytelingum, Rogers,\ \BBA\ Jennings}{Ramchurn
  et~al.}{2012}]{ramchurn:12}
Ramchurn, S.~D., Vytelingum, P., Rogers, A., \BBA\ Jennings, N.~R.
  \BBOP2012\BBCP.
\newblock \BBOQ Putting the `smarts' into the smart grid: A grand challenge for
  artificial intelligence\BBCQ\
\newblock {\Bem Communications of the ACM}, {\Bem 55\/}(4), 86--97.

\bibitem[\protect\BCAY{Regan\ \BBA\ Boutilier}{Regan\ \BBA\
  Boutilier}{2010}]{regan:10}
Regan, K.\BBACOMMA\  \BBA\ Boutilier, C. \BBOP2010\BBCP.
\newblock \BBOQ Robust policy computation in reward-uncertain {MDP}s using
  nondominated policies\BBCQ\
\newblock In {\Bem Proceedings of the AAAI Conference on Artificial
  Intelligence (AAAI)}, \BPGS\ 1127--1133.

\bibitem[\protect\BCAY{Rogers, Farinelli, Stranders,\ \BBA\ Jennings}{Rogers
  et~al.}{2011}]{rogers:11}
Rogers, A., Farinelli, A., Stranders, R., \BBA\ Jennings, N. \BBOP2011\BBCP.
\newblock \BBOQ Bounded approximate decentralised coordination via the
  {Max-Sum} algorithm\BBCQ\
\newblock {\Bem Artificial Intelligence}, {\Bem 175\/}(2), 730--759.

\bibitem[\protect\BCAY{Rollon\ \BBA\ Larrosa}{Rollon\ \BBA\
  Larrosa}{2012}]{rollon:12}
Rollon, E.\BBACOMMA\  \BBA\ Larrosa, J. \BBOP2012\BBCP.
\newblock \BBOQ Improved {Bounded Max-Sum} for distributed constraint
  optimization\BBCQ\
\newblock In {\Bem Proceedings of the International Conference on Principles
  and Practice of Constraint Programming (CP)}, \BPGS\ 624--632.

\bibitem[\protect\BCAY{Rossi, Beek,\ \BBA\ Walsh}{Rossi
  et~al.}{2006}]{rossi:06}
Rossi, F., Beek, P.~v., \BBA\ Walsh, T. \BBOP2006\BBCP.
\newblock {\Bem Handbook of Constraint Programming (Foundations of Artificial
  Intelligence)}.
\newblock Elsevier Science Inc.

\bibitem[\protect\BCAY{Rust, Picard,\ \BBA\ Ramparany}{Rust
  et~al.}{2016}]{rust:16}
Rust, P., Picard, G., \BBA\ Ramparany, F. \BBOP2016\BBCP.
\newblock \BBOQ Using message-passing {DCOP} algorithms to solve
  energy-efficient smart environment configuration problems\BBCQ\
\newblock In {\Bem Proceedings of the International Joint Conference on
  Artificial Intelligence ({IJCAI})}, \BPGS\ 468--474.

\bibitem[\protect\BCAY{Salukvad}{Salukvad}{1971}]{salukvad:71}
Salukvad, M.~E. \BBOP1971\BBCP.
\newblock \BBOQ Optimization of vector functionals. {I}. programming of optimal
  trajectories\BBCQ\
\newblock {\Bem Automation and remote Control}, {\Bem 32\/}(8), 5--15.

\bibitem[\protect\BCAY{Schwartz}{Schwartz}{1993}]{schwartz:93}
Schwartz, A. \BBOP1993\BBCP.
\newblock \BBOQ A reinforcement learning method for maximizing undiscounted
  rewards\BBCQ\
\newblock In {\Bem Proceedings of the International Conference on Machine
  Learning (ICML)}, \BPGS\ 298--305.

\bibitem[\protect\BCAY{Semnani\ \BBA\ Basir}{Semnani\ \BBA\
  Basir}{2013}]{hosseini:13}
Semnani, S.~H.\BBACOMMA\  \BBA\ Basir, O.~A. \BBOP2013\BBCP.
\newblock \BBOQ Target to sensor allocation: A hierarchical dynamic distributed
  constraint optimization approach\BBCQ\
\newblock {\Bem Computer Communications}, {\Bem 36\/}(9), 1024--1038.

\bibitem[\protect\BCAY{Shapiro\ \BBA\ Haralick}{Shapiro\ \BBA\
  Haralick}{1981}]{shapiro:81}
Shapiro, L.~G.\BBACOMMA\  \BBA\ Haralick, R.~M. \BBOP1981\BBCP.
\newblock \BBOQ Structural descriptions and inexact matching\BBCQ\
\newblock {\Bem IEEE Transactions on Pattern Analysis and Machine
  Intelligence}, {\Bem 3\/}(5), 504--519.

\bibitem[\protect\BCAY{Solomon}{Solomon}{1987}]{solomon:87}
Solomon, M.~M. \BBOP1987\BBCP.
\newblock \BBOQ Algorithms for the vehicle routing and scheduling problems with
  time window constraints\BBCQ\
\newblock {\Bem Operations Research}, {\Bem 35\/}(2), 254--265.

\bibitem[\protect\BCAY{Steinbauer\ \BBA\ Kleiner}{Steinbauer\ \BBA\
  Kleiner}{2012}]{steinbauer:12}
Steinbauer, G.\BBACOMMA\  \BBA\ Kleiner, A. \BBOP2012\BBCP.
\newblock \BBOQ Towards {CSP}-based mission dispatching in {C2/C4I}
  systems\BBCQ\
\newblock In {\Bem IEEE International Symposium on Safety, Security, and Rescue
  Robotics (SSRR)}, \BPGS\ 1--6.

\bibitem[\protect\BCAY{Stranders, Delle~Fave, Rogers,\ \BBA\
  Jennings}{Stranders et~al.}{2011}]{stranders:11}
Stranders, R., Delle~Fave, F.~M., Rogers, A., \BBA\ Jennings, N.
  \BBOP2011\BBCP.
\newblock \BBOQ {U-GDL}: A decentralised algorithm for {DCOP}s with
  uncertainty\BBCQ\
\newblock \BTR, University of Southampton, Department of Electronics and
  Computer Science.

\bibitem[\protect\BCAY{Stranders, Farinelli, Rogers,\ \BBA\ Jennings}{Stranders
  et~al.}{2009}]{stranders:09}
Stranders, R., Farinelli, A., Rogers, A., \BBA\ Jennings, N.~R. \BBOP2009\BBCP.
\newblock \BBOQ Decentralised coordination of continuously valued control
  parameters using the {Max-Sum} algorithm\BBCQ\
\newblock In {\Bem Proceedings of the International Conference on Autonomous
  Agents and Multiagent Systems (AAMAS)}, \BPGS\ 601--608.

\bibitem[\protect\BCAY{Stranders, Tran-Thanh, Fave, Rogers,\ \BBA\
  Jennings}{Stranders et~al.}{2012}]{stranders:12}
Stranders, R., Tran-Thanh, L., Fave, F. M.~D., Rogers, A., \BBA\ Jennings,
  N.~R. \BBOP2012\BBCP.
\newblock \BBOQ {DCOPs} and bandits: Exploration and exploitation in
  decentralised coordination\BBCQ\
\newblock In {\Bem Proceedings of the International Conference on Autonomous
  Agents and Multiagent Systems (AAMAS)}, \BPGS\ 289--296.

\bibitem[\protect\BCAY{Sultanik, Modi,\ \BBA\ Regli}{Sultanik
  et~al.}{2007}]{sultanik:07}
Sultanik, E., Modi, P.~J., \BBA\ Regli, W.~C. \BBOP2007\BBCP.
\newblock \BBOQ On modeling multiagent task scheduling as a distributed
  constraint optimization problem\BBCQ\
\newblock In {\Bem Proceedings of the International Joint Conference on
  Artificial Intelligence (IJCAI)}, \BPGS\ 1531--1536.

\bibitem[\protect\BCAY{Sultanik, Lass,\ \BBA\ Regli}{Sultanik
  et~al.}{2008}]{sultanik:07b}
Sultanik, E.~A., Lass, R.~N., \BBA\ Regli, W.~C. \BBOP2008\BBCP.
\newblock \BBOQ {DCOPolis}: A framework for simulating and deploying
  distributed constraint reasoning algorithms\BBCQ\
\newblock In {\Bem Proceedings of the International Conference on Autonomous
  Agents and Multiagent Systems (AAMAS)}, \BPGS\ 1667--1668. International
  Foundation for Autonomous Agents and Multiagent Systems.

\bibitem[\protect\BCAY{Tabakhi, Le, Fioretto,\ \BBA\ Yeoh}{Tabakhi
  et~al.}{2017}]{tabakhi:CP-17}
Tabakhi, A.~M., Le, T., Fioretto, F., \BBA\ Yeoh, W. \BBOP2017\BBCP.
\newblock \BBOQ Preference elicitation for {DCOPs}\BBCQ\
\newblock In {\Bem Proceedings of Principles and Practice of Constraint
  Programming {(CP)}}, \BPGS\ 278--296.

\bibitem[\protect\BCAY{Tassa, Zivan,\ \BBA\ Grinshpoun}{Tassa
  et~al.}{2016}]{tassa:16}
Tassa, T., Zivan, R., \BBA\ Grinshpoun, T. \BBOP2016\BBCP.
\newblock \BBOQ Preserving privacy in region optimal {DCOP} algorithms\BBCQ\
\newblock In {\Bem IJCAI}, \BPGS\ 496--502.

\bibitem[\protect\BCAY{Taylor, Jain, Jin, Yokoo,\ \BBA\ Tambe}{Taylor
  et~al.}{2010}]{taylor:10}
Taylor, M.~E., Jain, M., Jin, Y., Yokoo, M., \BBA\ Tambe, M. \BBOP2010\BBCP.
\newblock \BBOQ When should there be a me in team?: {D}istributed multi-agent
  optimization under uncertainty\BBCQ\
\newblock In {\Bem Proceedings of the International Conference on Autonomous
  Agents and Multiagent Systems (AAMAS)}, \BPGS\ 109--116.

\bibitem[\protect\BCAY{Taylor, Jain, Tandon, Yokoo,\ \BBA\ Tambe}{Taylor
  et~al.}{2011}]{taylor:11}
Taylor, M.~E., Jain, M., Tandon, P., Yokoo, M., \BBA\ Tambe, M. \BBOP2011\BBCP.
\newblock \BBOQ Distributed on-line multi-agent optimization under uncertainty:
  Balancing exploration and exploitation\BBCQ\
\newblock {\Bem Advances in Complex Systems}, {\Bem 14\/}(03), 471--528.

\bibitem[\protect\BCAY{Van~Hentenryck\ \BBA\ Michel}{Van~Hentenryck\ \BBA\
  Michel}{2009}]{hentenryck:09}
Van~Hentenryck, P.\BBACOMMA\  \BBA\ Michel, L. \BBOP2009\BBCP.
\newblock {\Bem Constraint-based local search}.
\newblock The MIT Press.

\bibitem[\protect\BCAY{Van~Katwijk, De~Schutter,\ \BBA\
  Hellendoorn}{Van~Katwijk et~al.}{2009}]{van:09}
Van~Katwijk, R., De~Schutter, B., \BBA\ Hellendoorn, J. \BBOP2009\BBCP.
\newblock \BBOQ Multi-agent control of traffic networks: Algorithm and case
  study\BBCQ\
\newblock In {\Bem IEEE International Conference on Intelligent Transportation
  Systems (ITSC)}, \BPGS\ 1--6.

\bibitem[\protect\BCAY{Verfaillie\ \BBA\ Jussien}{Verfaillie\ \BBA\
  Jussien}{2005}]{verfaillie:05}
Verfaillie, G.\BBACOMMA\  \BBA\ Jussien, N. \BBOP2005\BBCP.
\newblock \BBOQ Constraint solving in uncertain and dynamic environments: A
  survey\BBCQ\
\newblock {\Bem Constraints}, {\Bem 10\/}(3), 253--281.

\bibitem[\protect\BCAY{Vermorel\ \BBA\ Mohri}{Vermorel\ \BBA\
  Mohri}{2005}]{vermorel:05}
Vermorel, J.\BBACOMMA\  \BBA\ Mohri, M. \BBOP2005\BBCP.
\newblock \BBOQ Multi-armed bandit algorithms and empirical evaluation\BBCQ\
\newblock In {\Bem Proceedings of the European Conference on Machine Learning
  (ECML)}, \BPGS\ 437--448. Springer.

\bibitem[\protect\BCAY{Vinyals, Shieh, Cerquides, Rodriguez-Aguilar, Yin,
  Tambe,\ \BBA\ Bowring}{Vinyals et~al.}{2011}]{vinyals:11}
Vinyals, M., Shieh, E., Cerquides, J., Rodriguez-Aguilar, J., Yin, Z., Tambe,
  M., \BBA\ Bowring, E. \BBOP2011\BBCP.
\newblock \BBOQ Quality guarantees for region optimal {DCOP} algorithms\BBCQ\
\newblock In {\Bem Proceedings of the International Conference on Autonomous
  Agents and Multiagent Systems (AAMAS)}, \BPGS\ 133--140.

\bibitem[\protect\BCAY{Wack, Okimoto, Clement,\ \BBA\ Inoue}{Wack
  et~al.}{2014}]{wack:14}
Wack, M., Okimoto, T., Clement, M., \BBA\ Inoue, K. \BBOP2014\BBCP.
\newblock \BBOQ Local search based approximate algorithm for {Multi-Objective}
  {DCOPs}\BBCQ\
\newblock In {\Bem Proceedings of the Principles and Practice of Multi-Agent
  Systems (PRIMA)}, \BPGS\ 390--406.

\bibitem[\protect\BCAY{Wahbi, Ezzahir, Bessiere,\ \BBA\ Bouyakhf}{Wahbi
  et~al.}{2011}]{wahbi:11}
Wahbi, M., Ezzahir, R., Bessiere, C., \BBA\ Bouyakhf, E.-H. \BBOP2011\BBCP.
\newblock \BBOQ Dischoco 2: A platform for distributed constraint
  reasoning\BBCQ\
\newblock {\Bem International Workshop on Distributed Constraint Reasoning
  (DCR)}, {\Bem 11}, 112--121.

\bibitem[\protect\BCAY{Walsh\ \BBA\ Wellman}{Walsh\ \BBA\
  Wellman}{2000}]{walsh:00}
Walsh, W.~E.\BBACOMMA\  \BBA\ Wellman, M.~P. \BBOP2000\BBCP.
\newblock \BBOQ Modeling supply chain formation in multiagent systems\BBCQ\
\newblock In {\Bem Agent Mediated Electronic Commerce II}, \BPGS\ 94--101.
  Springer.

\bibitem[\protect\BCAY{Wang, Sycara,\ \BBA\ Scerri}{Wang
  et~al.}{2011}]{wang:11}
Wang, Y., Sycara, K., \BBA\ Scerri, P. \BBOP2011\BBCP.
\newblock \BBOQ Towards an understanding of the value of cooperation in
  uncertain world\BBCQ\
\newblock In {\Bem Proceedings of the International Joint Conferences on Web
  Intelligence and Intelligent Agent Technologies (WI-IAT)},
  \lowercase{\BVOL}~2, \BPGS\ 212--215.

\bibitem[\protect\BCAY{Winsper\ \BBA\ Chli}{Winsper\ \BBA\
  Chli}{2013}]{winsper:13}
Winsper, M.\BBACOMMA\  \BBA\ Chli, M. \BBOP2013\BBCP.
\newblock \BBOQ Decentralized supply chain formation using {Max-Sum} loopy
  belief propagation\BBCQ\
\newblock {\Bem Computational Intelligence}, {\Bem 29\/}(2), 281--309.

\bibitem[\protect\BCAY{Wood\ \BBA\ Wollenberg}{Wood\ \BBA\
  Wollenberg}{2012}]{wood2012power}
Wood, A.~J.\BBACOMMA\  \BBA\ Wollenberg, B.~F. \BBOP2012\BBCP.
\newblock {\Bem Power Generation, Operation, and Control}.
\newblock John Wiley \& Sons.

\bibitem[\protect\BCAY{Wooldridge}{Wooldridge}{2009}]{wooldridge:09}
Wooldridge, M. \BBOP2009\BBCP.
\newblock {\Bem An introduction to multiagent systems}.
\newblock John Wiley \& Sons.

\bibitem[\protect\BCAY{Wu\ \BBA\ Jennings}{Wu\ \BBA\ Jennings}{2014}]{wu:13}
Wu, F.\BBACOMMA\  \BBA\ Jennings, N.~R. \BBOP2014\BBCP.
\newblock \BBOQ Regret-based multi-agent coordination with uncertain task
  rewards\BBCQ\
\newblock In {\Bem Proceedings of the AAAI Conference on Artificial
  Intelligence (AAAI)}, \BPGS\ 1492--1499.

\bibitem[\protect\BCAY{Xie, Howitt,\ \BBA\ Raja}{Xie et~al.}{2007}]{xie:07}
Xie, J., Howitt, I., \BBA\ Raja, A. \BBOP2007\BBCP.
\newblock \BBOQ Cognitive radio resource management using multi-agent
  systems\BBCQ\
\newblock In {\Bem Proceedings of the Consumer Communications and Networking
  Conference (CCNC)}, \BPGS\ 1123--1127.

\bibitem[\protect\BCAY{Yan, Lee, Shen,\ \BBA\ Qiao}{Yan et~al.}{2013}]{yan:13}
Yan, Z., Lee, J.-H., Shen, S., \BBA\ Qiao, C. \BBOP2013\BBCP.
\newblock \BBOQ Novel branching-router-based multicast routing protocol with
  mobility support\BBCQ\
\newblock {\Bem IEEE Transactions on Parallel and Distributed Systems}, {\Bem
  24\/}(10), 2060--2068.

\bibitem[\protect\BCAY{Yedidsion\ \BBA\ Zivan}{Yedidsion\ \BBA\
  Zivan}{2014}]{yedidsion:14b}
Yedidsion, H.\BBACOMMA\  \BBA\ Zivan, R. \BBOP2014\BBCP.
\newblock \BBOQ Applying {DCOP\_MST} to a team of mobile robots with
  directional sensing abilities\BBCQ\
\newblock In {\Bem International Joint Workshop on Optimization In Multi-Agent
  Systems and Distributed Constraint Reasoning (OPTMAS-DCR)}.

\bibitem[\protect\BCAY{Yedidsion, Zivan,\ \BBA\ Farinelli}{Yedidsion
  et~al.}{2014}]{yedidsion:14}
Yedidsion, H., Zivan, R., \BBA\ Farinelli, A. \BBOP2014\BBCP.
\newblock \BBOQ Explorative {M}ax-{S}um for teams of mobile sensing
  agents\BBCQ\
\newblock In {\Bem Proceedings of the International Conference on Autonomous
  Agents and Multiagent Systems (AAMAS)}, \BPGS\ 549--556.

\bibitem[\protect\BCAY{Yeoh}{Yeoh}{2010}]{yeoh:10b}
Yeoh, W. \BBOP2010\BBCP.
\newblock {\Bem Speeding Up Distributed Constraint Optimization Search
  Algorithms}.
\newblock Ph.D.\ thesis, University of Southern California, Los Angeles (United
  States).

\bibitem[\protect\BCAY{Yeoh, Felner,\ \BBA\ Koenig}{Yeoh
  et~al.}{2010}]{yeoh:10}
Yeoh, W., Felner, A., \BBA\ Koenig, S. \BBOP2010\BBCP.
\newblock \BBOQ {BnB-ADOPT}: An asynchronous branch-and-bound {DCOP}
  algorithm\BBCQ\
\newblock {\Bem Journal of Artificial Intelligence Research}, {\Bem 38},
  85--133.

\bibitem[\protect\BCAY{Yeoh, Sun,\ \BBA\ Koenig}{Yeoh et~al.}{2009a}]{yeoh:09b}
Yeoh, W., Sun, X., \BBA\ Koenig, S. \BBOP2009a\BBCP.
\newblock \BBOQ Trading off solution quality for faster computation in {DCOP}
  search algorithms\BBCQ\
\newblock In {\Bem Proceedings of the International Joint Conference on
  Artificial Intelligence (IJCAI)}, \BPGS\ 354--360.

\bibitem[\protect\BCAY{Yeoh, Varakantham,\ \BBA\ Koenig}{Yeoh
  et~al.}{2009b}]{yeoh:09a}
Yeoh, W., Varakantham, P., \BBA\ Koenig, S. \BBOP2009b\BBCP.
\newblock \BBOQ Caching schemes for {DCOP} search algorithms\BBCQ\
\newblock In {\Bem Proceedings of the International Conference on Autonomous
  Agents and Multiagent Systems (AAMAS)}, \BPGS\ 609--616.

\bibitem[\protect\BCAY{Yeoh, Varakantham, Sun,\ \BBA\ Koenig}{Yeoh
  et~al.}{2011}]{yeoh:11}
Yeoh, W., Varakantham, P., Sun, X., \BBA\ Koenig, S. \BBOP2011\BBCP.
\newblock \BBOQ Incremental {DCOP} search algorithms for solving dynamic
  {DCOPs} ({E}xtended {A}bstract)\BBCQ\
\newblock In {\Bem Proceedings of the International Conference on Autonomous
  Agents and Multiagent Systems (AAMAS)}, \BPGS\ 1069--1070.

\bibitem[\protect\BCAY{Yeoh\ \BBA\ Yokoo}{Yeoh\ \BBA\ Yokoo}{2012}]{yeoh:12}
Yeoh, W.\BBACOMMA\  \BBA\ Yokoo, M. \BBOP2012\BBCP.
\newblock \BBOQ Distributed problem solving\BBCQ\
\newblock {\Bem AI Magazine}, {\Bem 33\/}(3), 53--65.

\bibitem[\protect\BCAY{Yokoo}{Yokoo}{2001}]{yokoo:01}
Yokoo, M.\BED. \BBOP2001\BBCP.
\newblock {\Bem Distributed constraint satisfaction: Foundation of cooperation
  in Multi-agent Systems}.
\newblock Springer Berlin Heidelberg.

\bibitem[\protect\BCAY{Yokoo, Durfee, Ishida,\ \BBA\ Kuwabara}{Yokoo
  et~al.}{1998}]{yokoo:98}
Yokoo, M., Durfee, E.~H., Ishida, T., \BBA\ Kuwabara, K. \BBOP1998\BBCP.
\newblock \BBOQ The distributed constraint satisfaction problem: Formalization
  and algorithms\BBCQ\
\newblock {\Bem Knowledge and Data Engineering, IEEE Transactions on}, {\Bem
  10\/}(5), 673--685.

\bibitem[\protect\BCAY{Yokoo\ \BBA\ Hirayama}{Yokoo\ \BBA\
  Hirayama}{2000}]{yokoo:00}
Yokoo, M.\BBACOMMA\  \BBA\ Hirayama, K. \BBOP2000\BBCP.
\newblock \BBOQ Algorithms for distributed constraint satisfaction: {A}
  review\BBCQ\
\newblock {\Bem Autonomous Agents and Multi-Agent Systems}, {\Bem 3\/}(2),
  185--207.

\bibitem[\protect\BCAY{Zhang\ \BBA\ Lesser}{Zhang\ \BBA\
  Lesser}{2013}]{zhang:13}
Zhang, C.\BBACOMMA\  \BBA\ Lesser, V. \BBOP2013\BBCP.
\newblock \BBOQ Coordinating multi-agent reinforcement learning with limited
  communication\BBCQ\
\newblock In {\Bem Proceedings of the International Conference on Autonomous
  Agents and Multiagent Systems (AAMAS)}, \BPGS\ 1101--1108.

\bibitem[\protect\BCAY{Zhang, Wang, Xing,\ \BBA\ Wittenberg}{Zhang
  et~al.}{2005}]{zhang:05}
Zhang, W., Wang, G., Xing, Z., \BBA\ Wittenberg, L. \BBOP2005\BBCP.
\newblock \BBOQ Distributed stochastic search and distributed breakout:
  Properties, comparison and applications to constraint optimization problems
  in sensor networks\BBCQ\
\newblock {\Bem Artificial Intelligence}, {\Bem 161\/}(1--2), 55--87.

\bibitem[\protect\BCAY{Zilberstein}{Zilberstein}{1996}]{zilberstein:96}
Zilberstein, S. \BBOP1996\BBCP.
\newblock \BBOQ Using anytime algorithms in intelligent systems\BBCQ\
\newblock {\Bem AI Magazine}, {\Bem 17\/}(3), 73.

\bibitem[\protect\BCAY{Zivan, Glinton,\ \BBA\ Sycara}{Zivan
  et~al.}{2009}]{zivan:09}
Zivan, R., Glinton, R., \BBA\ Sycara, K. \BBOP2009\BBCP.
\newblock \BBOQ Distributed constraint optimization for large teams of mobile
  sensing agents\BBCQ\
\newblock In {\Bem Proceedings of the International Joint Conferences on Web
  Intelligence and Intelligent Agent Technologies (WI-IAT)}, \BPGS\ 347--354.

\bibitem[\protect\BCAY{Zivan\ \BBA\ Meisels}{Zivan\ \BBA\
  Meisels}{2006}]{zivan:06}
Zivan, R.\BBACOMMA\  \BBA\ Meisels, A. \BBOP2006\BBCP.
\newblock \BBOQ Dynamic ordering for asynchronous backtracking on
  {DisCSPs}\BBCQ\
\newblock {\Bem Constraints}, {\Bem 11\/}(2-3), 179--197.

\bibitem[\protect\BCAY{Zivan, Okamoto,\ \BBA\ Peled}{Zivan
  et~al.}{2014}]{zivan:14}
Zivan, R., Okamoto, S., \BBA\ Peled, H. \BBOP2014\BBCP.
\newblock \BBOQ Explorative anytime local search for distributed constraint
  optimization\BBCQ\
\newblock {\Bem Artificial Intelligence}, {\Bem 212}, 1--26.

\bibitem[\protect\BCAY{Zivan, Parash, Cohen, Peled,\ \BBA\ Okamoto}{Zivan
  et~al.}{2017}]{zivan:17}
Zivan, R., Parash, T., Cohen, L., Peled, H., \BBA\ Okamoto, S. \BBOP2017\BBCP.
\newblock \BBOQ Balancing exploration and exploitation in incomplete
  {Min/Max-Sum} inference for distributed constraint optimization\BBCQ\
\newblock {\Bem Journal of Autonomous Agents and Multi-Agent Systems}, 1--43.

\bibitem[\protect\BCAY{Zivan\ \BBA\ Peled}{Zivan\ \BBA\ Peled}{2012}]{zivan:12}
Zivan, R.\BBACOMMA\  \BBA\ Peled, H. \BBOP2012\BBCP.
\newblock \BBOQ {Max/Min-Sum} distributed constraint optimization through value
  propagation on an alternating {DAG}\BBCQ\
\newblock In {\Bem Proceedings of the International Conference on Autonomous
  Agents and Multiagent Systems (AAMAS)}, \BPGS\ 265--272.

\end{thebibliography}
\bibliographystyle{theapa}

\end{document}